\documentclass[
]{ceurart}

\usepackage{listings}
\usepackage{booktabs}
\usepackage{multirow}
\usepackage{graphicx}
\usepackage{subfig}
\begin{document}

\copyrightyear{2022}
\copyrightclause{Copyright for this paper by its authors.
  Use permitted under Creative Commons License Attribution 4.0
  International (CC BY 4.0).}

\conference{In A. Martin, K. Hinkelmann, H.-G. Fill, A. Gerber, D. Lenat, R. Stolle, F. van Harmelen (Eds.), 
Proceedings of the AAAI 2022 Spring Symposium on Machine Learning and Knowledge Engineering for Hybrid Intelligence (AAAI-MAKE 2022), 
Stanford University, Palo Alto, California, USA, March 21–23, 2022.}

\title{Combining Learning from Human Feedback and Knowledge Engineering to Solve Hierarchical Tasks in Minecraft}


\author[1]{Vinicius G. Goecks}[%
email=vinicius.goecks@gmail.com,
url=https://vggoecks.com/,
]
\author[1,2]{Nicholas Waytowich}[%
email=nicholas.r.waytowich.civ@mail.mil,
]
\author[2]{David Watkins-Valls}[%
email=davidwatkins@cs.columbia.edu,
url=https://davidwatkinsvalls.com/,
]
\author[3]{Bharat Prakash}[%
email=bhp1@umbc.edu,
]
\address[1]{DEVCOM Army Research Laboratory,
  Aberdeen Proving Ground, Maryland, USA}
\address[2]{Columbia University,
  New York City, New York, USA}
\address[3]{University of Maryland,
Baltimore, Maryland, USA}

\begin{abstract}
  Real-world tasks of interest are generally poorly defined by human-readable descriptions and have no pre-defined reward signals unless it is defined by a human designer.
  Conversely, data-driven algorithms are often designed to solve a specific, narrowly defined, task with performance metrics that drives the agent's learning.
  In this work, we present the solution that won first place and was awarded the most human-like agent in the 2021 NeurIPS Competition MineRL BASALT Challenge: Learning from Human Feedback in Minecraft, which challenged participants to use human data to solve four tasks defined only by a natural language description and no reward function.
  Our approach uses the available human demonstration data to train an imitation learning policy for navigation and additional human feedback to train an image classifier. These modules, combined with an estimated odometry map, become a powerful state-machine designed to utilize human knowledge in a natural hierarchical paradigm. We compare this hybrid intelligence approach to both end-to-end machine learning and pure engineered solutions, which are then judged by human evaluators.
  Codebase is available at \url{https://github.com/viniciusguigo/kairos_minerl_basalt}.
\end{abstract}


\maketitle

\section{Introduction}
In this paper, we present the solution that won first place and was awarded the most human-like agent in the 2021 Neural Information Processing Systems (NeurIPS) MineRL Benchmark for Agents that Solve Almost-Lifelike Tasks (BASALT) competition\footnote{Official competition webpage: \url{https://www.aicrowd.com/challenges/neurips-2021-minerl-basalt-competition}.}.
Most artificial intelligence (AI) and reinforcement learning (RL) challenges involve solving tasks that have a reward function to optimize over.
Real-world tasks, however, do not automatically come with a reward function, and defining one from scratch can be quite challenging. 
Therefore, teaching AI agents to solve complex tasks and learn difficult behaviors without any reward function remains a major challenge for modern AI research.
The MineRL BASALT competition is aimed to address this challenge by developing AI agents that can solve complex, almost-lifelike, tasks in the challenging Minecraft environment \cite{johnson2016malmo} using only human feedback data and no reward function. 

The MineRL BASALT competition tasks do not contain any reward functions for the four tasks. We propose a human-centered machine learning approach instead of using traditional RL algorithms \cite{guss2019minerlcomp}. 
However, learning complex tasks with high-dimensional state-spaces (\emph{i.e.}  from images) using only end-to-end machine learning algorithms requires large amounts of high-quality data \cite{Bojarski2016,Nair2017}.
When learning from human feedback, this translates to large amounts of either human-collected or human-labeled data.
To circumvent this data requirement, we opted to combine machine learning with knowledge engineering, also known as hybrid intelligence or informed AI \cite{kamar2016directions,dellermann2019hybrid}.
Our approach uses human knowledge of the task to break it down into a natural hierarchy of subtasks.
Subtask selection is controlled by an engineered state-machine, which relies on estimated agent odometry and the outputs of a learned state classifier.
We also use the competition-provided human demonstration dataset to train a navigation policy subtask via imitation learning to replicate how humans traverse the environment.

In this paper, we give a detailed overview of our approach and perform an ablation study to investigate how well our hybrid intelligence approach works compared to using either learning from human demonstrations or engineered solutions alone. 
Our two main contributions are:
\begin{itemize}
    \item An architecture that combines knowledge engineering modules with machine learning modules to solve complex hierarchical tasks in Minecraft.
    \item Empirical results on how hybrid intelligence compares to both end-to-end machine learning and pure engineered approaches when solving complex, real-world-like tasks, as judged by real human evaluators.
\end{itemize}

\section{Background and Related Work}

\textbf{End-to-end machine learning: }in this work we use the term ``end-to-end machine learning'' for algorithms that learn purely from data with minimal bias or constraints added by human designers, besides the ones that are already inherently built-in the learning algorithm.
For example, deep reinforcement learning algorithms learning directly from raw pixels \cite{Mnih2013,Mnih2016} and algorithms that automatically decompose tasks in hierarchies with different time scales \cite{NIPS1992_d14220ee,vezhnevets2017feudal}.
This often includes massive parallelization and distributed computation \cite{espeholt2018impala,rudin2021learning} to fulfill the data requirement of these algorithms.
Other works train robotic agents to navigate through their environment using RGB-D (visible spectrum imagery, plus depth) information to determine optimal discrete steps to navigate to a visual goal~\cite{watkinsvalls2020}.
Despite all advances made in this field, these techniques have not been demonstrated to scale to tasks with the complexity presented in the MineRL BASALT competition.

\textbf{Human-in-the-loop machine learning: }Learning from human feedback can take different forms depending on how human interaction is used in the learning-loop \cite{Waytowich2018,goecks2020human}.
A learning agent can be trained based on human demonstrations of a task \cite{pomerleau1989alvinn,Argall2009,Bojarski2016}. Agents can learn from suboptimal demonstrations \cite{brown2019extrapolating}, end goals \cite{Rahmatizadeh2016}, or directly from successful examples instead of a reward function \cite{eysenbach2021replacing}. Human operators can augment the human demonstrations with online interventions \cite{Akgun2012,Akgun2012a,Saunders2017,goecks2018efficiently} or offline labeling \cite{Ross2011,Ross2013} while still maintaining success. Agents can learn the reward or cost function used by the demonstrator \cite{Ng2000,finn2016guided} through sparse interactions in the form of evaluative feedback \cite{knox2009interactively,MacGlashan2017,Warnell2018} or human preferences given a pair of trajectories \cite{Christiano2017}. Additionally, agents can learn from natural language-defined goals \cite{zhou2008inverse}. Finally, agents can learn from combining human data with reinforcement learning \cite{rajeswaran2017learning,goecks2019integrating,reddy2019sqil}.
However, these techniques have not scaled to tasks with the complexity presented in the MineRL BASALT competition.


\textbf{The Minecraft learning environment: }Minecraft is a procedurally-generated, 3D open-world game, where the agent observes the environment from a first-person perspective, collects resources, modifies the environment's terrain, crafts tools that augment the agent's capabilities, and possibly interacts with other agents in the same environment.
Given the open-world nature of this environment, there are no predefined tasks or built-in reward signals, giving the task designer flexibility to define tasks with virtually any level of complexity.
The release of Malmo \cite{johnson2016malmo}, a platform that enabled AI experimentation in the game of Minecraft, gave researchers the capability to develop learning agents to solve tasks similar or analogous to the ones seen in the real world.

The Minecraft environment also served as a platform to collect large human demonstration datasets such as the \textit{MineRL-v0} dataset \cite{guss2019minerldata} and experiment with large scale imitation learning algorithms \cite{amiranashvili2020scaling} as a world generator for realistic terrain rendering \cite{hao2021gancraft}, a sample-efficient reinforcement learning competition environment using human priors (MineRL DIAMOND challenge) \cite{guss2019minerlcomp}; and now as a platform for a competition on solving human-judged tasks defined by a human-readable description and no pre-defined reward function, the MineRL BASALT competition \cite{shah2021basalt}.

\section{Problem Formulation}

The 2021 NeurIPS MineRL BASALT competition, ``Learning from Human Feedback in Minecraft'', challenged participants to come up with creative solutions to solve four different tasks in Minecraft \cite{shah2021basalt} using the ``MineRL: Towards AI in Minecraft''\footnote{MineRL webpage: \url{https://minerl.io/}.} simulator \cite{guss2019minerldata}.
These tasks aimed to mimic real-world tasks, being defined only by a human-readable description and no reward signal returned by the environment.
The official task descriptions for the MineRL BASALT competition\footnote{MineRL BASALT documentation: \url{https://minerl.io/basalt/}.} were the following:
\begin{itemize}
    \item \textit{FindCave}: The agent should search for a cave and terminate the episode when it is inside one.
    \item \textit{MakeWaterfall}: After spawning in a mountainous area, the agent should build a beautiful waterfall and then reposition itself to take a scenic picture of the same waterfall.
    \item \textit{CreateVillageAnimalPen}: After spawning in a village, the agent should build an animal pen containing two of the same kind of animal next to one of the houses in a village.
    \item \textit{BuildVillageHouse}: Using items in its starting inventory, the agent should build a new house in the style of the village, in an appropriate location (e.g\. next to the path through the village) without harming the village in the process.
\end{itemize}

The competition organizers also provided each participant team with a dataset of 40 to 80 human demonstrations for each task, not all completing the task, and the starter codebase to train a behavior cloning baseline.
Additionally, the training time for all four tasks together was limited to four days and participants were allowed to collect up to 10 hours of additional human-in-the-loop feedback.

\section{Methods}

\begin{figure}[!ht]
  \centering
  \includegraphics[width=0.9\linewidth]{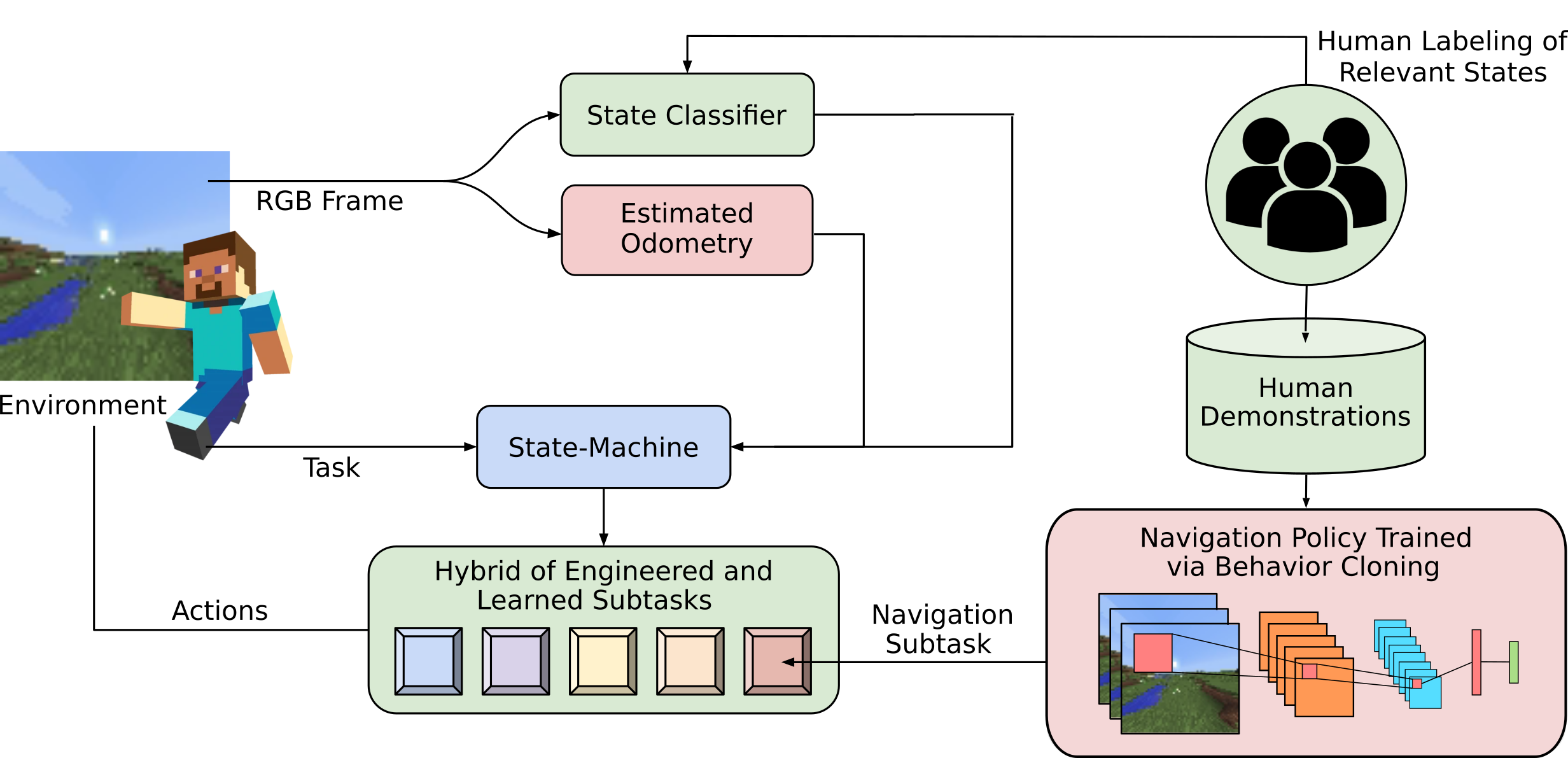}
  \caption{Diagram illustrating our approach. Using data from the available human demonstration dataset, humans provide additional binary labels to image frames to be used to train a state classifier that can detect relevant features in the environment such as caves and mountains. The available human demonstration dataset is also used to train a navigation policy via imitation learning to replicate how humans traverse the environment. A separate odometry module estimates the current agent's position and heading solely based on the action taken by the end. During test time, the agent uses the learned state classifier to provide useful information to an engineered state-machine that controls which subtask the agent should execute at every time-step.}
  \label{fig:diagram}
\end{figure}

Since no reward signal was given by the competition organizers and compute time was limited, direct deep reinforcement learning approaches were not feasible \cite{Mnih2013, lillicrap2015continuous,Mnih2016}.
With the limited human demonstration dataset, end-to-end behavior cloning also did not result in high-performing policies, because imitation learning requires large amounts of high-quality data \cite{Bojarski2016,Nair2017}.
We also attempted to solve the tasks using adversarial imitation learning approaches such as Generative Adversarial Imitation Learning (GAIL) \cite{ho2016generative}, however, the large-observation space and limited compute time also made this approach infeasible.

Hence, to solve the four tasks of the MineRL BASALT competition, we opted to combine machine learning with knowledge engineering, also known as \emph{hybrid intelligence} \cite{kamar2016directions,dellermann2019hybrid}.
As seen in the main diagram of our approach shown in Figure \ref{fig:diagram}, the machine learning part of our method is seen in two different modules: first, we learn a state classifier using additional human feedback to identify relevant states in the environment; second, we learn a navigation subtask separately for each task via imitation learning using the human demonstration dataset provided by the competition.
The knowledge engineering part is seen in three different modules: first, given the relevant states classified by the machine learning model and knowledge of the tasks, we designed a state-machine that defines a hierarchy of subtasks and controls which one should be executed at every time-step; second, we engineered solutions for the more challenging subtasks that we were not able to learn directly from data; and third, we engineered an estimated odometry module that provides additional information to the state-machine and enables the execution of the more complex engineered subtasks.

\subsection{State Classification}

\begin{figure}[!ht]
    \centering
    \begin{tabular}{cccc}
        \subfloat[\textit{has\_cave}]{\includegraphics[width=.18\linewidth]{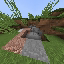}} &
        \subfloat[\textit{inside\_cave}]{\includegraphics[width=.18\linewidth]{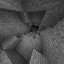}} &
        \subfloat[\textit{danger\_ahead}]{\includegraphics[width=.18\linewidth]{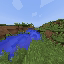}} &
        \subfloat[\textit{has\_mountain}]{\includegraphics[width=.18\linewidth]{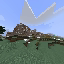}}\\
        \subfloat[\textit{facing\_wall}]{\includegraphics[width=.18\linewidth]{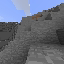}} &
        \subfloat[\textit{at\_the\_top}]{\includegraphics[width=.18\linewidth]{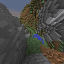}} &
        \subfloat[\textit{good\_waterfall\_view}]{\includegraphics[width=.18\linewidth]{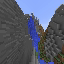}} &
        \subfloat[\textit{good\_pen\_view}]{\includegraphics[width=.18\linewidth]{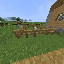}}\\
        \subfloat[\textit{good\_house\_view}]{\includegraphics[width=.18\linewidth]{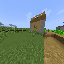}} &
        \subfloat[\textit{has\_animals}]{\includegraphics[width=.18\linewidth]{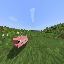}} &
        \subfloat[\textit{has\_open\_space}]{\includegraphics[width=.18\linewidth]{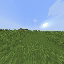}} &
        \subfloat[\textit{animals\_inside\_pen}]{\includegraphics[width=.18\linewidth]{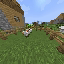}}
    \end{tabular}
    \caption{Illustration of positive classes of states classified using additional human feedback. Humans were given image frames from previously collected human demonstration data and were assigned to give binary labels for each of the illustrated 12 states, plus a null case when no relevant states were identified.}
    \label{fig:state_classifier}
\end{figure}

Our approach relies on a state machine that changes the high-level goals depending on the task to be solved. Without having information about the environment's voxel data, we opted to use the visual RGB information from the simulator to determine the agent's current state. Due to the low resolution of the simulator of $64\times64\times3$, we decided to use a classifier that labels the whole image rather than parts of the image, such as \emph{You Only Look Once} (YOLO)~\cite{yolo}. Multiple labels can be present on the same image as there were cases with multiple objects or scenes of interest at the same time in the field of view of the agent.
These labels are used by the state-machine to decide which subtask should be followed at any time-step.

There are 13 possible labels for an RGB frame, as illustrated in Figure \ref{fig:state_classifier} and described below: 
\begin{itemize}
    \item \textit{none}: frame contains no relevant states (54.47 \% of the labels).
    \item \textit{has\_cave}: agent is looking at a cave (1.39 \% of the labels).
    \item \textit{inside\_cave}: agent is inside a cave (1.29 \% of the labels).
    \item \textit{danger\_ahead}: agent is looking at a large body of water (3.83 \% of the labels).
    \item \textit{has\_mountain}: agent has a complete view of a mountain (usually, from far away) (4.38 \% of the labels).
    \item \textit{facing\_wall}: agent is facing a wall that cannot be traversed by jumping only (4.55 \% of the labels).
    \item \textit{at\_the\_top}: agent is at the top of a mountain and looking at a cliff (3.97 \% of the labels).
    \item \textit{good\_waterfall\_view}: agent see water in view (3.16 \% of the labels).
    \item \textit{good\_pen\_view}: agent has framed a pen with animals in view (4.12 \% of the labels).
    \item \textit{good\_house\_view}: agent has framed a house in view (2.58 \% of the labels).
    \item \textit{has\_animals}: frame contains animals (pig, horse, cow, sheep, or chicken) (9.38 \% of the labels).
    \item \textit{has\_open\_space}: agent is looking at an open-space of about 6x6 blocks with no small cliffs or obstacles (flat area to build a small house or pen) (7.33 \% of the labels).
    \item \textit{animals\_inside\_pen}: agent is inside the pen after luring all animals and has them in view (0.81 \% of the labels).
\end{itemize}


The possible labels were defined by a human designer with knowledge of the relevant states to be identified and given to the state-machine to solve all tasks.
These labels were also designed to be relevant to all tasks to ease data collection and labelling efforts.
For example, the ``has\_open\_space'' label identifies flat areas that are ideal to build pens or houses for both \emph{CreateVillageAnimalPen} and \emph{BuildVillageHouse} tasks.
Unknown and other non-relevant states were attached the label ``\emph{none}'' to indicate that no important states were in view.

To train this system, we labeled $81,888$ images using a custom graphical user interface (GUI), as showed in Appendix \ref{appendix:label_state_gui}.
Once the data was labeled, $80\%$ of images were used for training, $10\%$ were used for validation, and $10\%$ for testing.
The model is a convolutional neural network (CNN) classifier with a $64\times64\times3$ input and $13\times1$ output. The architecutre of our CNN is modeled after the Deep TAMER (Training Agents Manually via Evaluative Reinforcement)~\cite{Warnell2018} model.
The problem of training with an uneven number of labels for each class was mitigated by implementing a weighted sampling scheme that sampled more often classes with lower representation with probability:
\begin{equation}
    P(x_i) = 1 - \frac{N_i}{M},
\end{equation}
where $P(x_i)$ is the probability of sampling class $i$ that contains $N_i$ number of labels out of the total $M$ labels for all classes.

\subsection{Estimated Odometry}

\begin{figure}[!ht]
  \centering
  \includegraphics[width=0.65\linewidth]{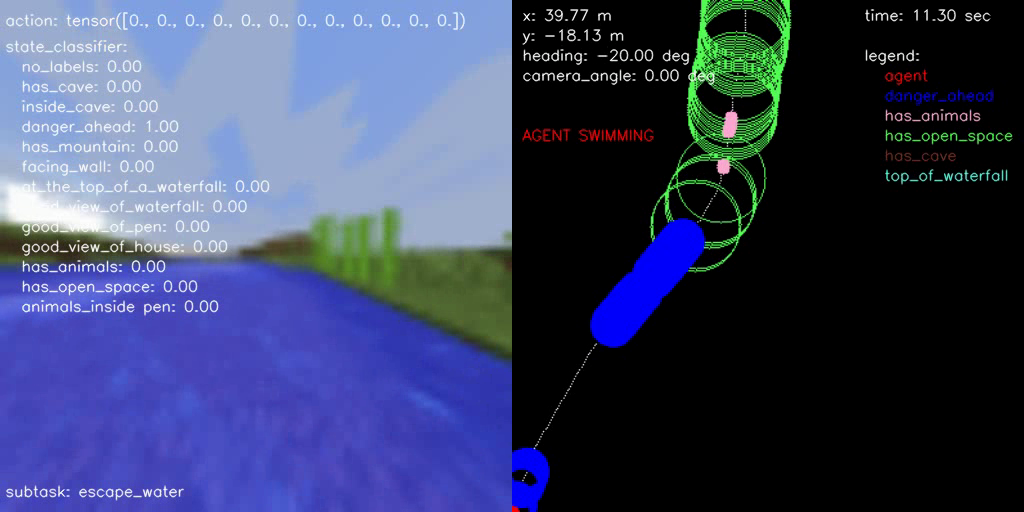}
  \caption{Example of odometry map (right frame) generated in real time from the agent's actions as it traverses the environment. Besides the agent's pose, the classified states from image data (left frame) also have their locations tagged in the map to be used for specific subtasks. For example, when the agent finishes building the pen, it uses the location of previously seen animals to attempt to navigate and lure them to the pen.}
  \label{fig:odometry}
\end{figure}

Some of the engineered subtasks required basic localization of the agent and relevant states of the environment.
For example, the agent needs to know the location of previously seen animals to guide them to a pen in the \textit{CreateVillageAnimalPen} task. However, under the rules of the competition, we were not allowed to use any additional information from the simulator besides the current view of the agent and the player's inventory. Which means there is no information about the ground truth location of the agent, camera pose, or any explicit terrain information available.

Given these constraints, we opted to implement a custom odometry method that took into consideration only the actions from the agent and basic characteristics of the Minecraft simulator.
It is known that the simulator runs at $20$ frames per second, which means there is a $0.05$ second interval between each frame.
According to the Minecraft Wiki\footnote{Minecraft Wiki - Walking: \url{https://minecraft.fandom.com/wiki/Walking}.}, walking speed is approximately $4.317$ m/s, $5.612$ m/s while sprinting, or $7.127$ m/s when sprinting and jumping at the same time, which translates to approximately $0.216$, $0.281$, or $0.356$ meters per frame when walking, sprinting, or sprinting and jumping, respectively.
If we assume the agent is operating in a flat world, starting at position $(0, 0)$ in map coordinates facing north, when the agent executes a move forward action its position is moved $0.216$ meters north to position $(0, 0.216)$. The agent does not have acceleration in MineRL and their velocity is immediately updated upon keypress. 
Another limitation is that we were not able to reliably detect when the agent is stuck behind an obstacle, which causes the estimated location to drift even though the agent is not moving in the simulator.

Since the agent already commands camera angles in degrees, the heading angle $\theta$ is simply updated by accumulating the horizontal camera angles commanded by the agent.
More generally, this odometric estimation assumes the agent follows point-mass kinematics:
\begin{align*}
    \dot{x} &= V cos(\theta) \\
    \dot{y} &= V sin(\theta),
\end{align*}
where $V$ is the velocity of the agent, which takes into consideration if the agent is walking, sprinting, or sprinting and jumping.

Using this estimated odometry and the learned state classifier, it is possible to attach a coordinate to each classified state and map important features of the environment so they can be used later by different subtasks and the state-machine.
For example, it is possible to keep track of where the agent found water, caves, animals, and areas of open space that can be used to build a pen or a house.
Figure \ref{fig:odometry} shows a sample of the resulting map overlaid with the classified states' location and current odometry readings.

\subsection{Learning and Engineering Subtasks and the State-Machine}

One of the main complexities in solving the proposed four tasks is that most required the agent to have certain levels of perception capabilities, memory, and reasoning over long-term dependencies in a hierarchical manner.
For example, the \textit{CreateVillageAnimalPen} task required the agent to first build a pen nearby an existing village, which requires identifying what a village is, then indicating a good location to build a pen such as a flat terrain.
Once the pen was built, the agent had to search for at least two of the same animal type in the nearby vicinity using $64 \times 64$ resolution images as input. Return animals to the pen required coordination to combine different blocks and place them adjacently to each other in a closed shape.
After the animals were found, the agent had to lure them with the specific food type they eat, walk them back to the pen the agent initially built, leave the pen, lock the animals inside, then take a picture of the pen with the animals inside.

Reasoning over these long-term dependencies in hierarchical tasks is one of the main challenges of end-to-end learning-based approaches \cite{vezhnevets2017feudal}.
Conversely, reactive policies such as the one required to navigate with certain boundaries and avoid obstacles have been learned directly from demonstration data or agent-generated trajectories \cite{giusti2015machine,Bojarski2016}.
In this work, we use human knowledge of the tasks to decompose these complex tasks in multiple subtasks, which are either reactive policies learned from data or directly engineered, and a state-machine that selects the most appropriate one to be followed at every time-step.
The subtask that performs task-specific navigation is learned from the provided human demonstration dataset. For example, searching for the best spot to place a waterfall in the \textit{MakeWaterfall} task requires navigation.
Subtasks with little demonstration data available are engineered in combination with the learned state classifier. Throwing a snowball while inside the cave to signal the end of the episode can be engineered using human demonstration data. 

Once the complex tasks are decomposed into multiple small subtasks, we engineered a state-machine in combination with the learned state classifier to select the best subtask to be followed at every time-step.
Each of these engineered subtasks was implemented by a human designer who hard-coded a sequence of actions to be taken using the same interface available to the agent.
In addition to these subtasks, the human designer also implemented a safety-critical subtask allowing the agent to escape a body of water whenever the state classifier detects that the agent is swimming.
Appendix \ref{appendix:state_machine} describes in detail the sequence of substasks followed by the state-machine for each task.

\subsection{Evaluation Methods}

In this work, we compared four different approaches to solve the four tasks proposed in the Minecraft competition:
\begin{itemize}
    \item \textbf{Hybrid: }the main proposed agent in this work, that combines both learned and engineered modules. The learned modules are the navigation subtask policy (learns how to navigate using the human demonstration data provided by the competition) and the state classifier (learns how to identify relevant states using additional human-labeled data). 
    The engineered modules are the multiple subtasks, hand-designed to solve subtasks that were not able to be learned from data. These engineered modules are the estimated odometry and the state-machine, which uses the output of the state classifier and engineered task structure to select which subtask should be followed at each time-step.
    \item \textbf{Engineered: }almost identical to the Hybrid agent described above, however, the navigation subtask policy that was learned from human demonstrations is now replaced by a hand-designed module that randomly selects movement and camera commands to explore the environment.
    \item \textbf{Behavior Cloning: }end-to-end imitation learning agent that learns solely from the human demonstration data provided during the competition. This agent does not use any other learned or engineered module, which includes the state classifier, the estimated odometry, and the state-machine.
    \item \textbf{Human: }human-generated trajectories provided by the competition. They are neither guaranteed to solve the task nor solve it optimally because they depend on the level of expertise of each human controlling the agent.
\end{itemize}

To collect human evaluations for each of the four baselines in a head-to-head comparison we set up a web application\footnote{Custom MineRL BASALT evaluation webpage: \url{https://kairosminerl.herokuapp.com/}.} similar to how the teams were evaluated during the official MineRL BASALT competition, as seen in Appendix \ref{appendix:evaluation_interface}.
In our case, each participant was asked to see two videos of different agents performing the same task then to answer three questions:
\begin{enumerate}
    \item \textit{Which agent best completed the task?}
    \item \textit{Which agent was the fastest completing the task?}
    \item \textit{Which agent had a more human-like behavior?}
\end{enumerate}
For each question, the participants were given three possible answers: ``Agent 1'', ``Agent 2'', or ``None''.

Our database had videos of all four types of agents (Behavior Cloning, Engineered, Hybrid, and Human) performing all four tasks (FindCave, MakeWaterfall, CreateVillageAnimalPen, and BuildVillageHouse).
There were $10$ videos of each agent type solving all four tasks, for a total of $160$ videos in the database.
Task, agent type, and videos were uniformly sampled from the database at each time a new evaluation form was generated and presented to the human evaluator.

We collected a total of 268 evaluation forms (pairwise comparison where a human evaluator judged which agent was the best, fastest, and more human-like performing the tasks) from 7 different human evaluators.

All agent-generated videos were scaled from the original $64 \times 64$ image resolution returned by the environment to $512 \times 512$ image resolution in an attempt to make the videos clearer for the human evaluators.
The videos of the ''Human'' agent type were randomly selected from the video demonstrations provided by the MineRL BASALT competition and scaled to $512 \times 512$ image resolution to match the agent-generated videos.
All videos were generated and saved at $20$ frames per second to match the sampling rate of the Minecraft simulator used by both agents and humans.

\section{Results and Discussion}

\begin{table}[]
\caption{Summary of the \textit{TrueSkill}\textsuperscript{TM}\cite{herbrich2006trueskill} scores with mean and standard deviation computed from human evaluations separately for each performance metric and agent type averaged out over all tasks. Scores were computed after collecting 268 evaluations from 7 different human evaluators.}
\label{tab:trueskill_summary_average}
\resizebox{\textwidth}{!}{%
\begin{tabular}{@{}cccccc@{}}
\toprule
\multirow{2}{*}{\textbf{Task}} & \multirow{2}{*}{\textbf{\begin{tabular}[c]{@{}c@{}}Performance\ Metric\end{tabular}}} & \multicolumn{4}{c}{\textbf{TrueSkill Rating}} \\ \cmidrule(l){3-6} 
&  & \textbf{Behavior Cloning} & \textbf{Engineered} & \textbf{Hybrid} & \textbf{Human} \\ \midrule

\multirow{3}{*}{\begin{tabular}[c]{@{}c@{}}All Tasks\\ Combined\end{tabular}} & \begin{tabular}[c]{@{}c@{}}Best\\ Performer\end{tabular} & $20.30 \pm 1.81$ & $24.21 \pm 1.46$ & $25.49 \pm 1.40$ & $32.56 \pm 1.85$ \\ \cmidrule(l){2-6} 
& \begin{tabular}[c]{@{}c@{}}Fastest\\ Performer\end{tabular} & $19.42 \pm 1.94$ & $26.92 \pm 1.45$ & $27.59 \pm 1.38$ & $28.36 \pm 1.69$ \\ \cmidrule(l){2-6} 
& \begin{tabular}[c]{@{}c@{}}More Human-like\\ Behavior\end{tabular} & $20.09 \pm 2.04$ & $26.02 \pm 1.56$ & $26.94 \pm 1.57$ & $36.41 \pm 2.12$ \\ \bottomrule
\end{tabular}%
}
\end{table}

Each combination of condition (behavior cloning, engineered, hybrid, human) and performance metric (best performer, fastest performer, most human-like performer) is treated as a separate participant of a one-versus-one competition where skill rating is computed using the \textit{TrueSkill}\textsuperscript{TM}\footnote{Microsoft's \textit{TrueSkill}\textsuperscript{TM} Ranking System: \url{https://www.microsoft.com/en-us/research/project/trueskill-ranking-system}} Bayesian ranking system \cite{herbrich2006trueskill}.
In this Bayesian ranking system, the skill of each participant is characterized by a Gaussian distribution with a mean value $\mu$, representing the average skill of a participant and standard deviation $\sigma$ representing the degree of uncertainty in the participant's skill.
There are three possible outcomes after each comparison: the first agent wins the comparison, the second agent the comparison, or there is a draw (human evaluator selects ''None'' when asked which participant performed better in a given metric).
Given this outcome, the \textit{TrueSkill}\textsuperscript{TM} ranking system updates the belief distribution of each participant using Bayes' Theorem \cite{herbrich2006trueskill}, similar to how scores were computed in the official 2021 NeurIPS MineRL BASALT competition.
We used the open-source \emph{TrueSkill} Python package\footnote{\emph{TrueSkill} Python package: \url{https://github.com/sublee/trueskill} and \url{https://trueskill.org/}.}.

The final mean and standard deviation of the \textit{TrueSkill}\textsuperscript{TM} scores computed for each performance metric and agent type are shown in Table \ref{tab:trueskill_summary_average}.
The scores were computed after collecting 268 evaluations from 7 different human evaluators.
Our main proposed ``Hybrid'' agent, which combines engineered and learned modules, outperforms both pure hand-designed (``Engineered'') and pure learned (``Behavior Cloning'') agents in the ``Best Performer'' category, achieving $5.3 \%$ and $25.6 \%$ higher mean skill rating when compared to the ``Engineered'' and ``Behavior Cloning'' baselines, respectively.
However, when compared to the ``Human'' scores, our main proposed agent achieves $21.7 \%$ lower mean skill rating, illustrating that even our best approach is still not able to outperform a human player with respect to best performing the task.

When looking at the ``Fastest Performer'' metric, our ``Hybrid'' agent outperforms both ``Engineered'' and ``Behavior Cloning'' baselines, respectively, scoring only $2.7 \%$ lower than the human players.
As expected, in the ``More Human-like Behavior'' performance metric the ``Human'' baseline wins by a large margin, however, the ``Hybrid'' still outperforms all other baselines, including the ``Behavior Cloning'' agent, which is purely learned from human data. We attribute this to the fact that the pure learned agent did not make use of the safety-critical engineered subtask, which allowed the agent to escape bodies of water and other obstacles around the environment.
Plots showing how the \textit{TrueSkill}\textsuperscript{TM} scores evolved after each match (one-to-one comparison between different agent types) are shown in Appendix \ref{appendix:trueskill}.

\begin{table}[]
\caption{Summary of the \textit{TrueSkill}\textsuperscript{TM}\cite{herbrich2006trueskill} scores with mean and standard deviation computed from human evaluations separately for each performance metric, agent type, and task. Scores were computed after collecting 268 evaluations from 7 different human evaluators.}
\label{tab:trueskill_summary}
\resizebox{\textwidth}{!}{%
\begin{tabular}{@{}cccccc@{}}
\toprule
\multirow{2}{*}{\textbf{Task}} & \multirow{2}{*}{\textbf{\begin{tabular}[c]{@{}c@{}}Performance\ Metric\end{tabular}}} & \multicolumn{4}{c}{\textbf{TrueSkill Rating}} \\ \cmidrule(l){3-6} 
&  & \textbf{Behavior Cloning} & \textbf{Engineered} & \textbf{Hybrid} & \textbf{Human} \\ \midrule

\multirow{3}{*}{FindCave} & \begin{tabular}[c]{@{}c@{}}Best\\ Performer\end{tabular} & $24.32 \pm 1.27$ & $24.29 \pm 1.21$ & $25.14 \pm 1.19$ & $32.90 \pm 1.52$ \\ \cmidrule(l){2-6} 
& \begin{tabular}[c]{@{}c@{}}Fastest\\ Performer\end{tabular} & $24.65 \pm 1.27$ & $24.16 \pm 1.21$ & $24.79 \pm 1.19$ & $32.75 \pm 1.54$ \\ \cmidrule(l){2-6} 
& \begin{tabular}[c]{@{}c@{}}More Human-like\\ Behavior\end{tabular} & $21.53 \pm 1.70$ & $26.61 \pm 1.43$ & $28.25 \pm 1.51$ & $38.95 \pm 1.96$ \\ \midrule

\multirow{3}{*}{MakeWaterfall} & \begin{tabular}[c]{@{}c@{}}Best\\ Performer\end{tabular} & $15.16 \pm 2.10$ & $23.16 \pm 1.60$ & $26.53 \pm 1.39$ & $24.39 \pm 1.62$ \\ \cmidrule(l){2-6} 
& \begin{tabular}[c]{@{}c@{}}Fastest\\ Performer\end{tabular} & $14.67 \pm 2.26$ & $28.95 \pm 1.74$ & $28.88 \pm 1.46$ & $18.85 \pm 2.02$ \\ \cmidrule(l){2-6} 
& \begin{tabular}[c]{@{}c@{}}More Human-like\\ Behavior\end{tabular} & $21.27 \pm 1.98$ & $24.51 \pm 1.52$ & $26.91 \pm 1.35$ & $26.48 \pm 1.61$ \\ \midrule

\multirow{3}{*}{\begin{tabular}[c]{@{}c@{}}CreateVillage\\ AnimalPen\end{tabular}} & \begin{tabular}[c]{@{}c@{}}Best\\ Performer\end{tabular} & $21.87 \pm 1.94$ & $23.56 \pm 1.38$ & $26.49 \pm 1.48$ & $33.89 \pm 1.73$ \\ \cmidrule(l){2-6} 
& \begin{tabular}[c]{@{}c@{}}Fastest\\ Performer\end{tabular} & $18.62 \pm 2.27$ & $27.00 \pm 1.32$ & $29.93 \pm 1.50$ & $28.59 \pm 1.53$ \\ \cmidrule(l){2-6} 
& \begin{tabular}[c]{@{}c@{}}More Human-like\\ Behavior\end{tabular} & $21.54 \pm 2.29$ & $25.53 \pm 1.57$ & $27.99 \pm 1.68$ & $40.60 \pm 2.44$ \\ \midrule

\multirow{3}{*}{\begin{tabular}[c]{@{}c@{}}BuildVillage\\ House\end{tabular}} & \begin{tabular}[c]{@{}c@{}}Best\\ Performer\end{tabular} & $19.83 \pm 1.92$ & $25.81 \pm 1.66$ & $23.81 \pm 1.55$ & $39.05 \pm 2.53$ \\ \cmidrule(l){2-6} 
& \begin{tabular}[c]{@{}c@{}}Fastest\\ Performer\end{tabular} & $19.75 \pm 1.97$ & $27.58 \pm 1.54$ & $26.76 \pm 1.35$ & $33.24 \pm 1.67$ \\ \cmidrule(l){2-6} 
& \begin{tabular}[c]{@{}c@{}}More Human-like\\ Behavior\end{tabular} & $16.04 \pm 2.19$ & $27.42 \pm 1.72$ & $24.61 \pm 1.72$ & $39.61 \pm 2.46$ \\ \bottomrule
\end{tabular}%
}
\end{table}

Table \ref{tab:trueskill_summary} breaks down the results presented in Table \ref{tab:trueskill_summary_average} for each separate task.
Similar to what was discussed for Table \ref{tab:trueskill_summary_average}, excluding the ``Human'' baseline, the ``Hybrid'' approach outperforms both ``Behavior Cloning'' and ``Engineered'' baselines in terms of mean skill rating in 8 out of the 12 performance metrics, or in $66.6 \%$ of the comparisons.
Similarly, hybrid intelligence approaches, which include both ``Hybrid'' and ``Engineered'' baselines, outperform the pure learning ``Behavior Cloning'' approach in all 12 performance metrics, not taking into account the ``Human'' baseline.
The ``Hybrid'' approach only outperforms the ``Human'' baseline in 4 out of the 12 performance metrics, or in $33.3 \%$ of the comparisons.

Particularly for the \textit{MakeWaterfall} task, the proposed hybrid approach outperforms human players for all performance metrics.
The largest margin observed is for the ``Fastest Performer'' metric; the hybrid approach scores $53.2 \%$ higher than the human players.
This large margin comes from human players taking more time to find the best spot to place the waterfall and signal the end of the episode when compared to the engineered subtasks.
Plots showing all results for each individual pairwise comparison are shown in Appendix \ref{appendix:barplot}.


We now consider qualtitative evaluation of our agents. When solving the \textit{FindCave} task\footnote{Sample trajectory of hybrid agent solving the \textit{FindCave} task: \url{https://youtu.be/MR8q3Xre_XY}.}, the agent uses the learned navigation policy to search for caves while avoiding water while simultaneously building the map of its environment. Once the agent finds the cave, it throws the snowball to signal the end of the episode. 
In the \textit{MakeWaterfall} task\footnote{Sample trajectory of hybrid agent solving the \textit{MakeWaterfall} task: \url{https://youtu.be/eXp1urKXIPQ}.}, the hybrid agent uses the learned navigation policy to climb the mountains, detects a good location to build the waterfall, builds it, then moves to the picture location using engineered subtasks, and throws the snowball to signal the end of the episode.
For the \textit{CreateVillageAnimalPen} task\footnote{Sample trajectory of hybrid agent solving the \textit{CreateVillageAnimalPen} task: \url{https://youtu.be/b8xDMxEZmAE}.}, the agent uses the learned navigation policy and the state classifier to search for an open location to build a pen, builds the pen using an engineered building subtask that repeats the actions taken by the human demonstrators, uses the state classifier and odometry map to go to previously seen animal locations, and then attempts to lure them back to the pen and throws the snowball to signal the end of the episode. 
Finally, when solving the \textit{BuildVillageHouse} task\footnote{Sample trajectory of hybrid agent solving the \textit{BuildVillageHouse} task: \url{https://youtu.be/_uKO-ZqBMWQ}.}, our hybrid agent spawns nearby a village and uses the learned navigation policy and the state classifier to search for an open location to build a house, builds a house using an engineered building subtask that repeats the actions taken by the human demonstrators, tours the house, and throws the snowball to signal the end of the episode.
Each of the described subtasks are shown in Appendix \ref{appendix:frames} as a sequence of frames.

\section{Conclusions}

In this paper, we presented the solution that won first place and was awarded the most human-like agent in the 2021 NeurIPS MineRL BASALT competition, ``Learning from Human Feedback in Minecraft.''
Our approach used the available human demonstration data and additional human feedback to train machine learning modules that were combined with engineered ones to solve hierarchical tasks in Minecraft.

The proposed method was compared to both end-to-end machine learning and pure engineered solutions by collecting human evaluations that judged agents in head-to-head matches to answer which agent best solved the task, which agent was the fastest, and which one had the most human-like behavior.
These human evaluations were converted to a skill rating score for each question, similar to how players are ranked in multiplayer online games.

After collecting 268 human evaluations, we showed that hybrid intelligence approaches outperformed end-to-end machine learning approaches in all 12 performance metrics computed, even outperforming human players in 4 of them.
Our results also showed that incorporating machine learning modules for navigation as opposed to engineering navigation policies led to higher scores in 8 out of 12 performance metrics.

Overall, we demonstrated that our hybrid intelligence approach proves advantageous to solve hierarchical tasks, compared to end-to-end machine learning approaches when the subcomponents of the task are understood by human experts and limited human feedback data is available.

\begin{acknowledgments}
  Research was sponsored by the Army Research Laboratory and was accomplished partly under Cooperative Agreement Number W911NF-20-2-0114. The views and conclusions contained in this document are those of the authors and should not be interpreted as representing the official policies, either expressed or implied, of the Army Research Laboratory or the U.S. Government. The U.S. Government is authorized to reproduce and distribute reprints for Government purposes notwithstanding any copyright notation herein. 
\end{acknowledgments}

\bibliography{sample-ceur}

%
\newpage
\section*{Appendix}
\appendix

\section{State Classifier Labeling GUI}\label{appendix:label_state_gui}

\begin{figure}[!ht]
  \centering
  \includegraphics[width=0.75\linewidth]{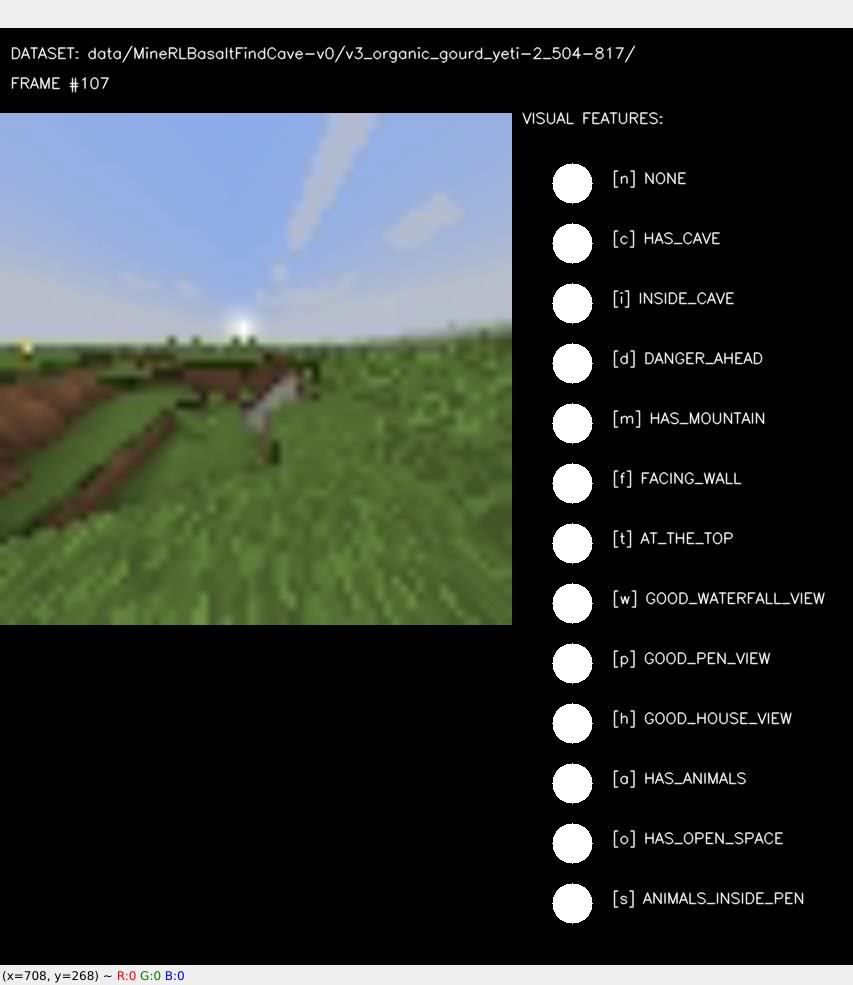}
  \caption{Custom GUI to relabel human dataset provided by the competition to train a classifier to identify relevant states for the state-machine.}
  \label{fig:label_state_gui}
\end{figure}

The labeling process of relevant states to the state-machine uses both mouse clicks and keyboard presses and takes place in a custom GUI, as seen in Figure \ref{fig:label_state_gui}.
On the top left of the GUI, users can double-check which dataset and frame number they are labeling. Below that, the GUI displays the RGB frame to be labeled (center left) and the options for labels (center right, same labels for all tasks). To label a frame, the user can simply press the keyboard key corresponding to the desired label (shown in brackets, for example, $[c]$ for $has\_cave$), or click in the white circles in front of the label, which will then turn green, indicating that the label was selected. Frames will automatically advance when a key is pressed. If the users only use the mouse to select the labels, they will istill need to press the keyboard key to advance to the next frame (any key for a label that was already selected clicking).

\section{State-Machine Definition for each Task}\label{appendix:state_machine}

The sequence of subtasks used by the state-machine for each task is defined as follows:
\begin{itemize}
    \item \textit{FindCave}:
        \begin{enumerate}
            \item Use navigation policy to traverse the environment and search for caves;
            \item If the state classifier detects the agent is inside a cave, throw a snowball to signal that the task was completed (end of episode).
        \end{enumerate}
    \item \textit{MakeWaterfall}:
        \begin{enumerate}
            \item Use navigation policy to traverse the environment and search for a place in the mountains to make a waterfall;
            \item If the state classifier detects the agent is at the top of a mountain, build additional blocks to give additional height to the waterfall;
            \item Once additional blocks are built, look down and place the waterfall by equipping and using the bucket item filled with water;
            \item After the waterfall is built, keep moving forward to move away from it;
            \item Once the agent has moved away from the waterfall, turn around and throw a snowball to signal that a picture was taken, and the task was completed (end of episode).
        \end{enumerate}
    \item \textit{CreateVillageAnimalPen}:
        \begin{enumerate}
            \item Use navigation policy to traverse the environment and search for a place to build a pen;
            \item If the state classifier detects an open-space, build the pen. The subtask to build the pen directly repeats the actions taken by a human while building the pen, as observed in provided demonstration dataset;
            \item Once the pen is built, use the estimated odometry map to navigate to the closest animal location. If no animals were seen before, use navigation policy to traverse the environment and search for animals;
            \item At the closest animal location, equip food to attract attention of the animals and lure them;
            \item Using the estimated odometry map, move back to where the pen was built while animals are following the agent;
            \item Once inside the pen together with the animals, move away from pen, turn around and throw a snowball to signal that the task was completed (end of episode).
        \end{enumerate}
    \item \textit{BuildVillageHouse}:
        \begin{enumerate}
            \item Use navigation policy to traverse the environment and search for a place to build a house;
            \item If the state classifier detects an open-space, build the house. The subtask to build the house directly repeats the actions taken by a human while building the house, as observed in provided demonstration dataset;
            \item Once the house is built, move away from it, turn around and throw a snowball to signal that the task was completed (end of episode).
        \end{enumerate}
\end{itemize}

\section{Human Evaluation Interface}\label{appendix:evaluation_interface}

\begin{figure}[!ht]
  \centering
  \includegraphics[width=0.95\linewidth]{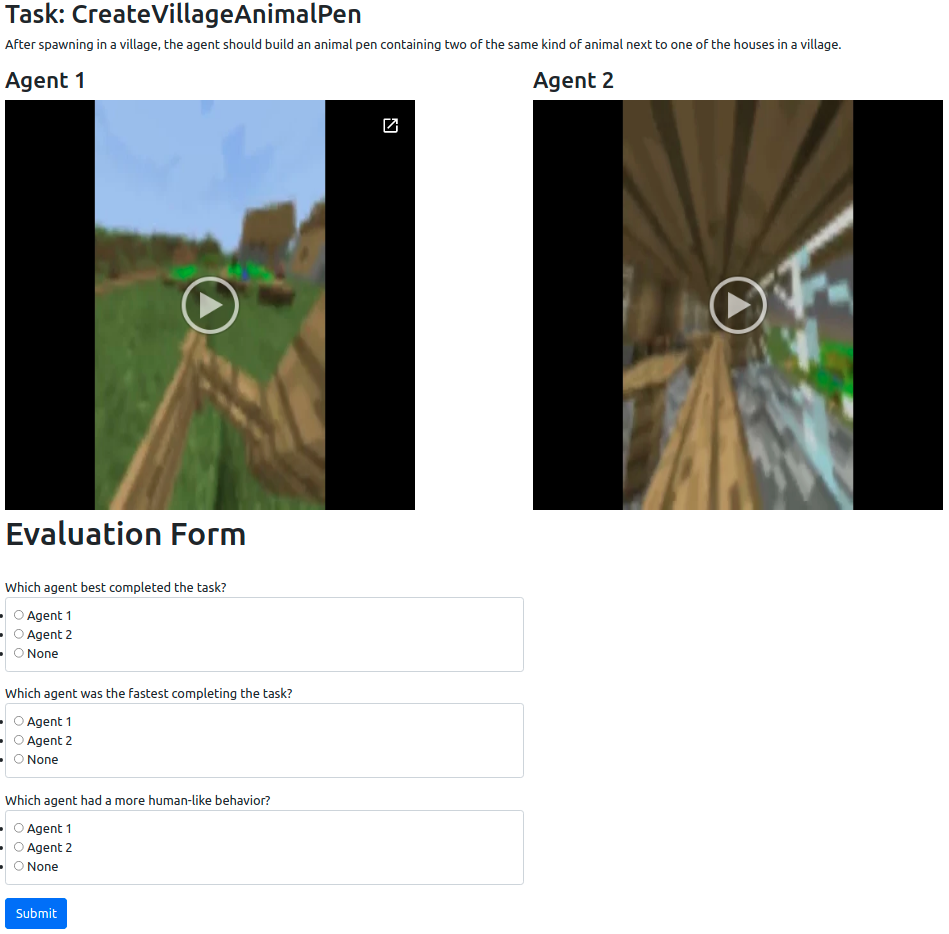}
  \caption{Web evaluation form used to collect additional human evaluation data to evaluate the multiple agent conditions presented in paper.}
  \label{fig:evaluation_interface}
\end{figure}

Figure \ref{fig:evaluation_interface} shows a sample of the web evaluation form available at \url{https://kairosminerl.herokuapp.com/} that was used to collect human evaluations for each of the four baselines in a head-to-head comparison, similar to how the teams were evaluated during the official MineRL BASALT competition.
Each participant was asked to see two videos of different agents performing the same task then answer three questions with respect to the agent's performance.

\section{\textit{TrueSkill}\textsuperscript{TM} Score per Match}\label{appendix:trueskill}

\begin{figure}[!ht]
  \centering
    \begin{tabular}{ccc}
        \subfloat[Best Performer]{\includegraphics[width=0.31\linewidth]{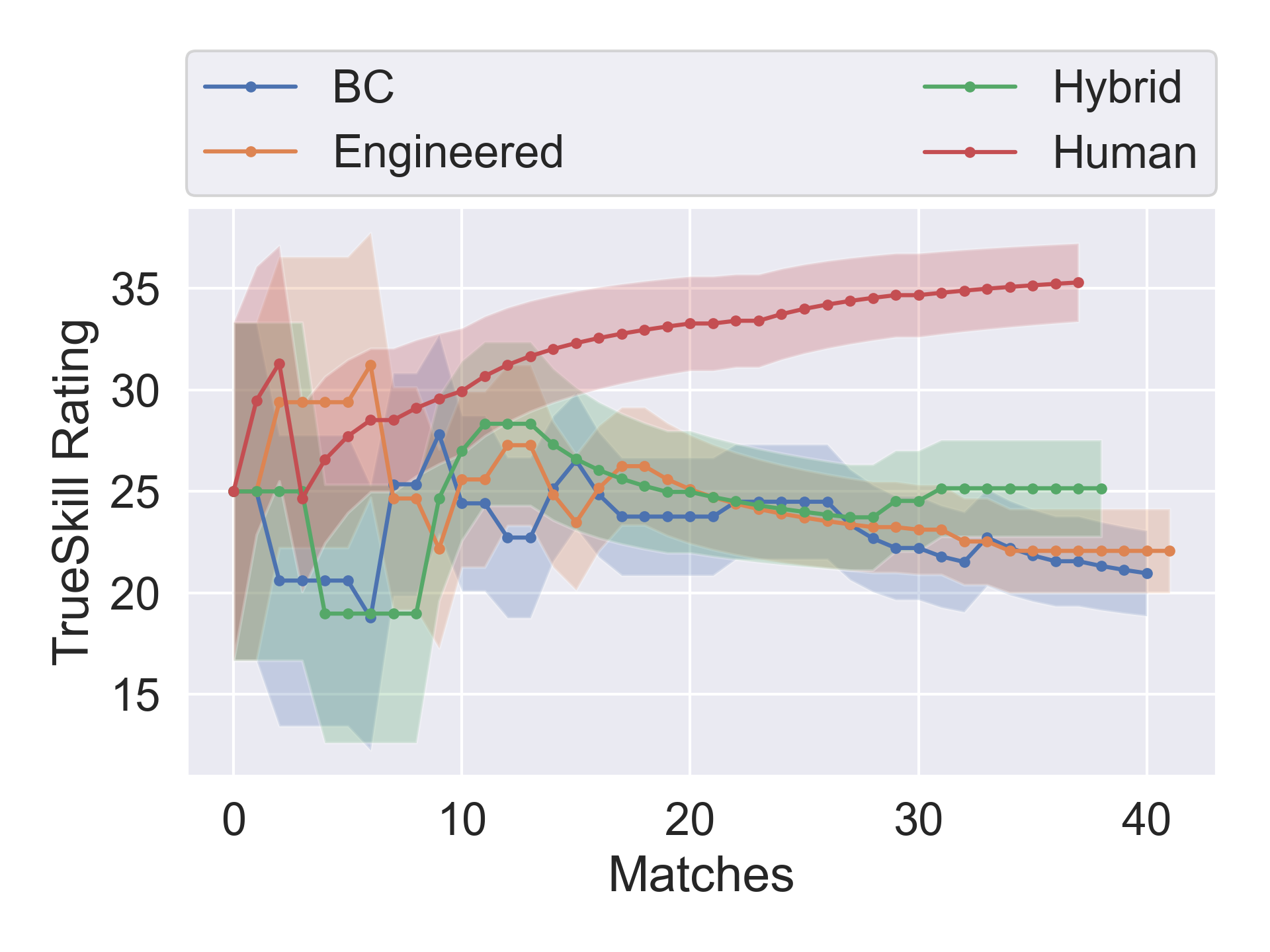}} &
        \subfloat[Fastest Performer]{\includegraphics[width=0.31\linewidth]{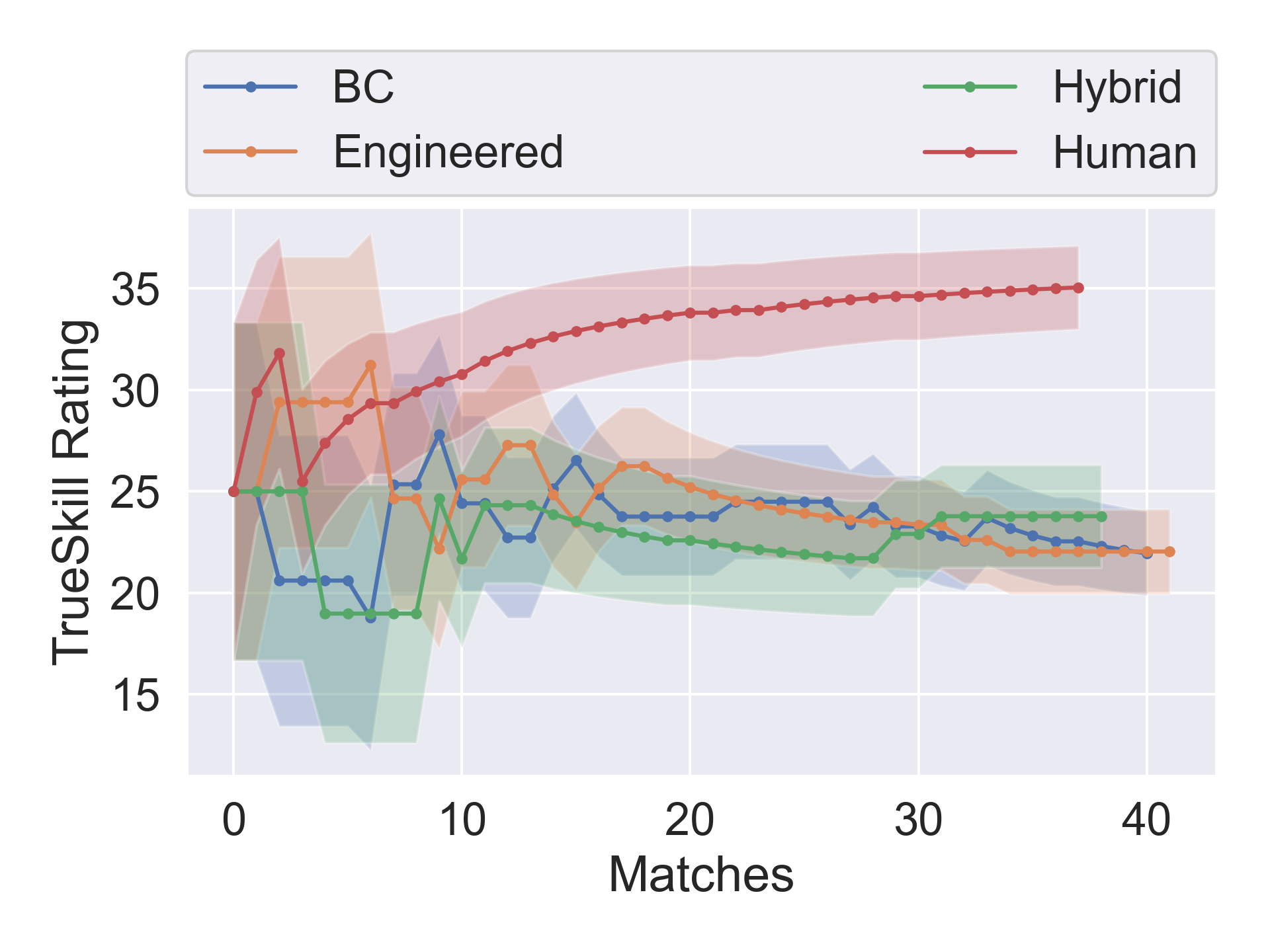}} &
        \subfloat[More Human-like Behavior]{\includegraphics[width=0.31\linewidth]{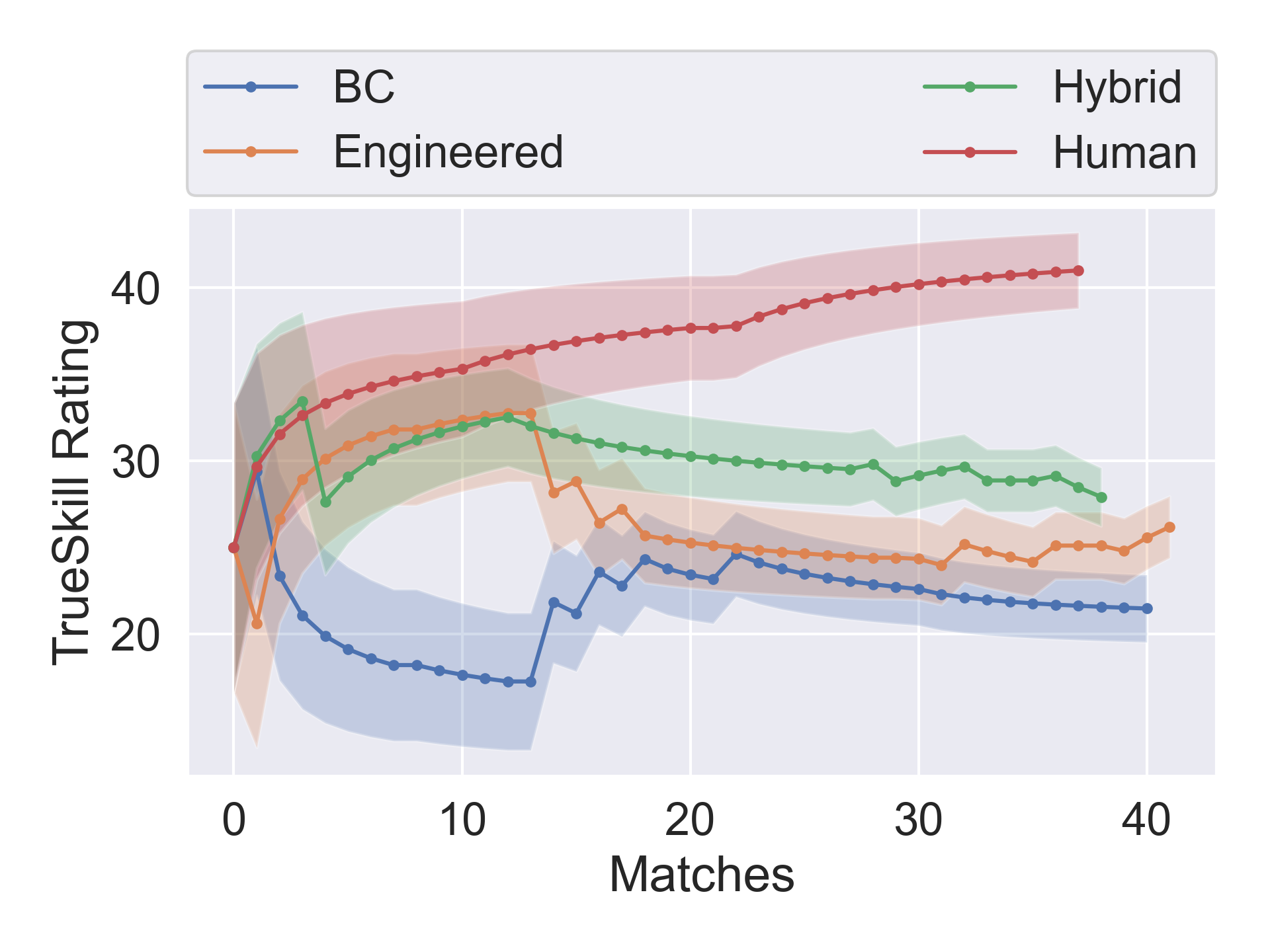}}
    \end{tabular}
  \caption{\textit{TrueSkill}\textsuperscript{TM}\cite{herbrich2006trueskill} scores computed from human evaluations separately for each performance metric and for each agent type performing the \textit{FindCave} task.}
  \label{fig:cave_trueskill}
\end{figure}

\begin{figure}[!ht]
  \centering
    \begin{tabular}{ccc}
        \subfloat[Best Performer]{\includegraphics[width=0.31\linewidth]{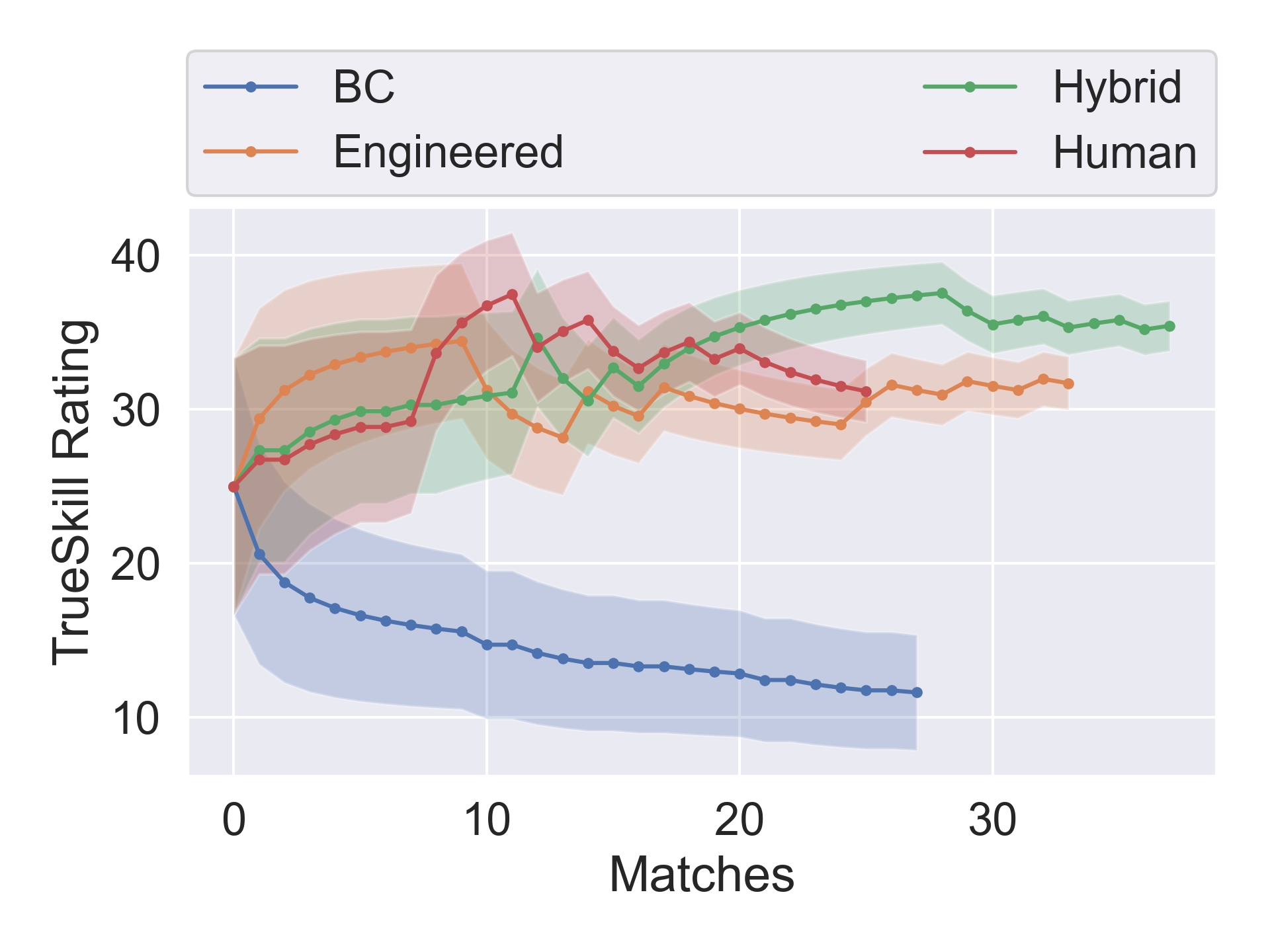}} &
        \subfloat[Fastest Performer]{\includegraphics[width=0.31\linewidth]{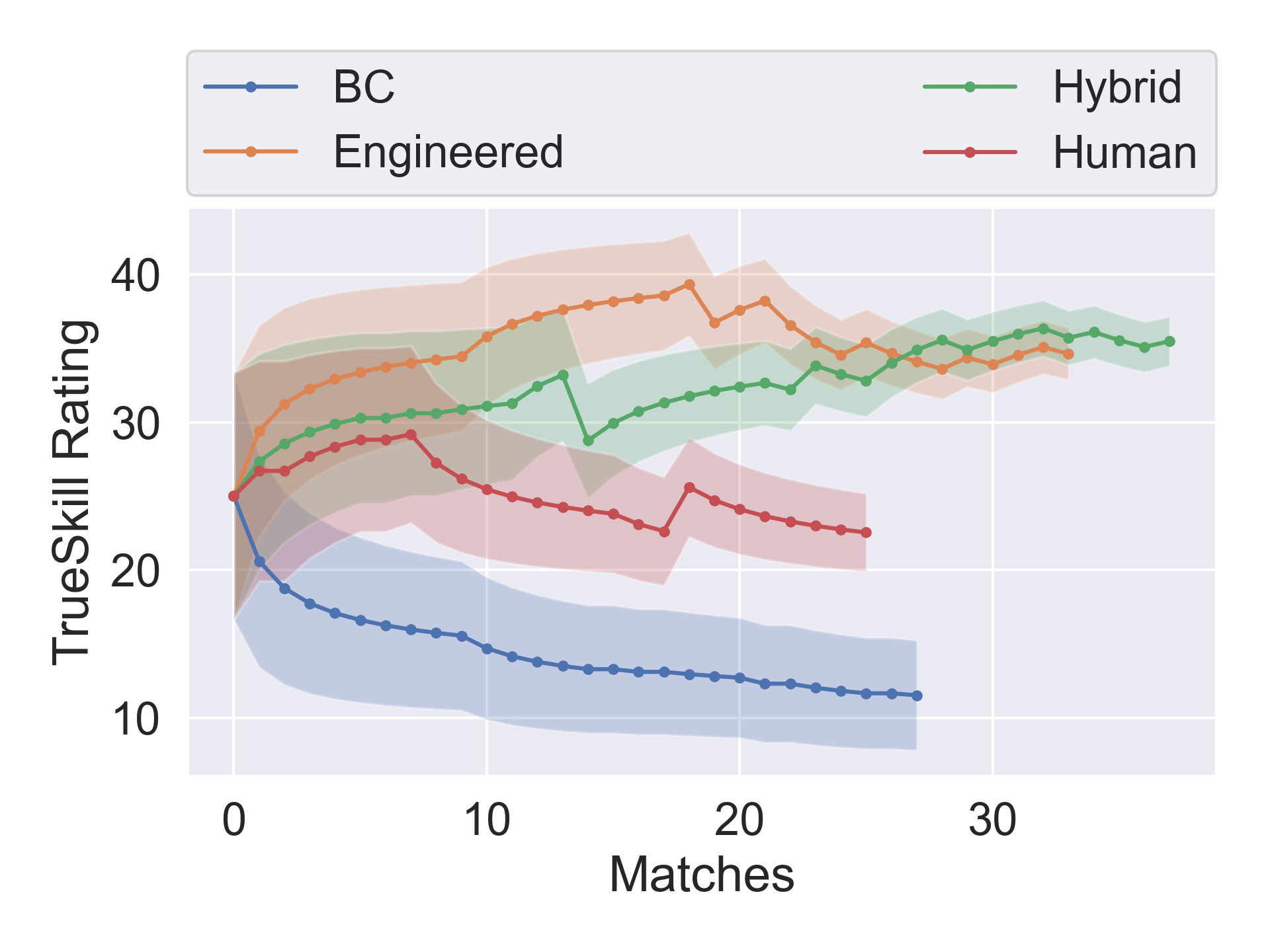}} &
        \subfloat[More Human-like Behavior]{\includegraphics[width=0.31\linewidth]{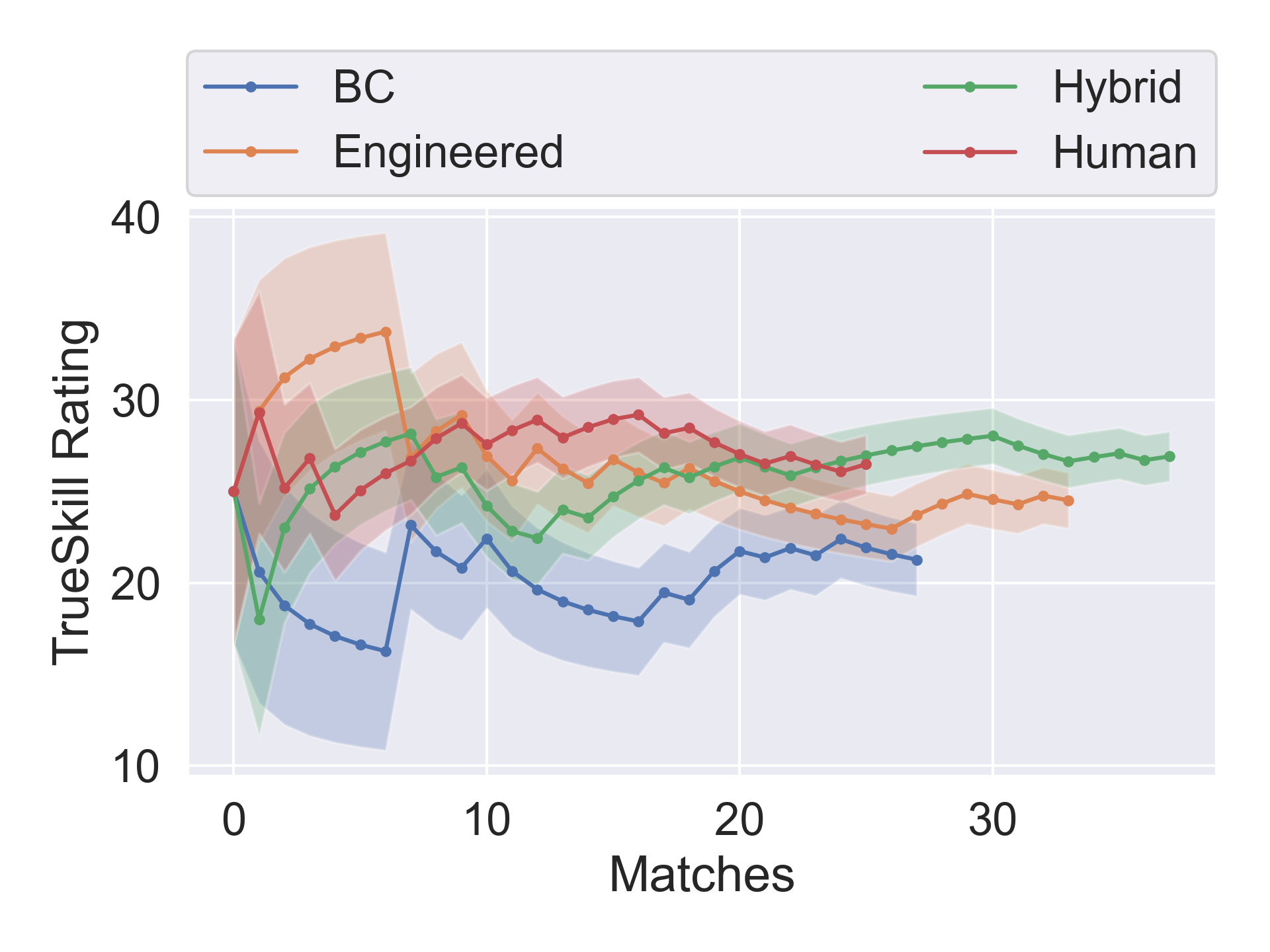}}
    \end{tabular}
  \caption{\textit{TrueSkill}\textsuperscript{TM}\cite{herbrich2006trueskill} scores computed from human evaluations separately for each performance metric and for each agent type performing the \textit{MakeWaterfall} task.}
  \label{fig:waterfall_trueskill}
\end{figure}

\begin{figure}[!ht]
  \centering
    \begin{tabular}{ccc}
        \subfloat[Best Performer]{\includegraphics[width=0.31\linewidth]{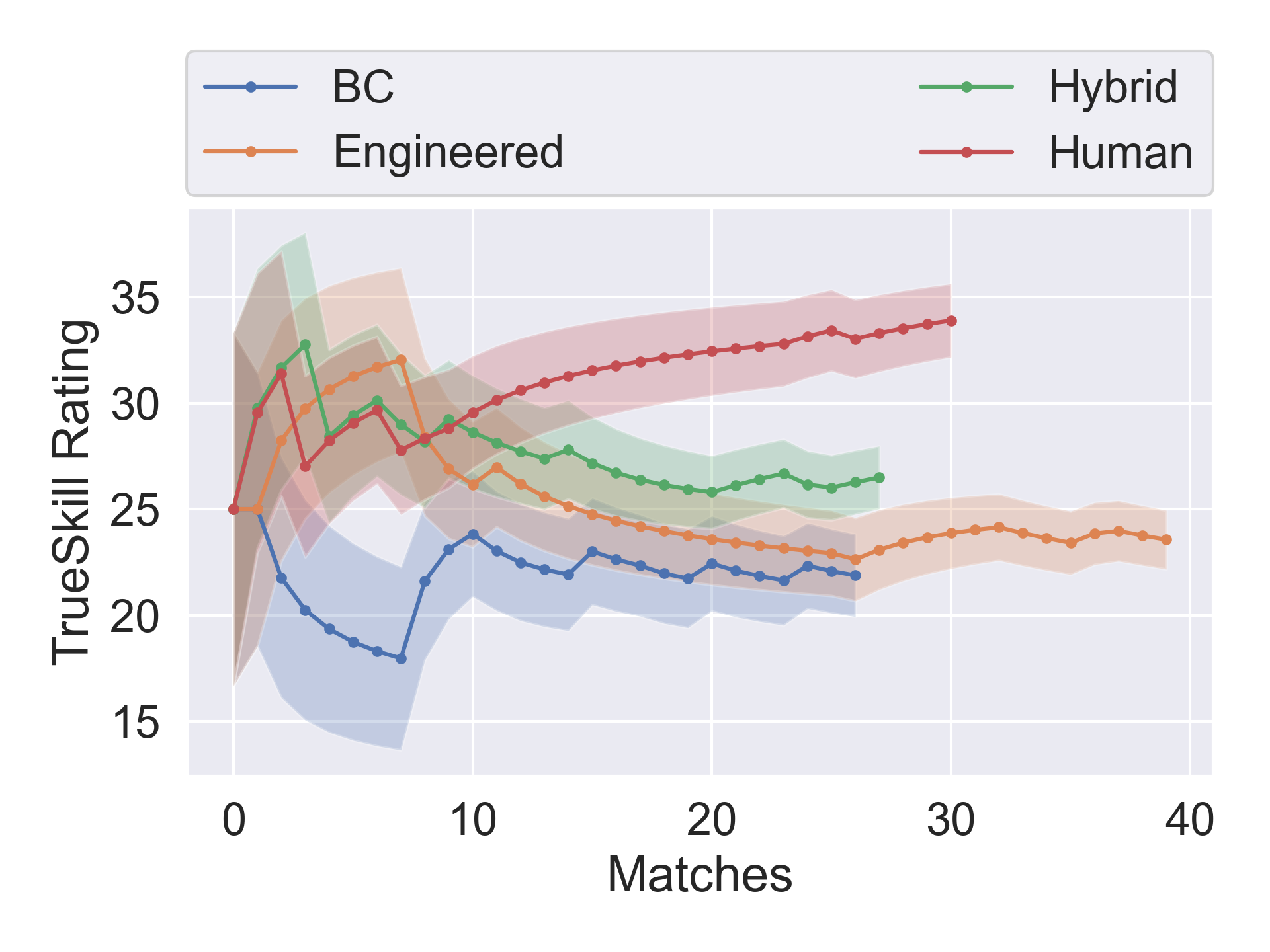}} &
        \subfloat[Fastest Performer]{\includegraphics[width=0.31\linewidth]{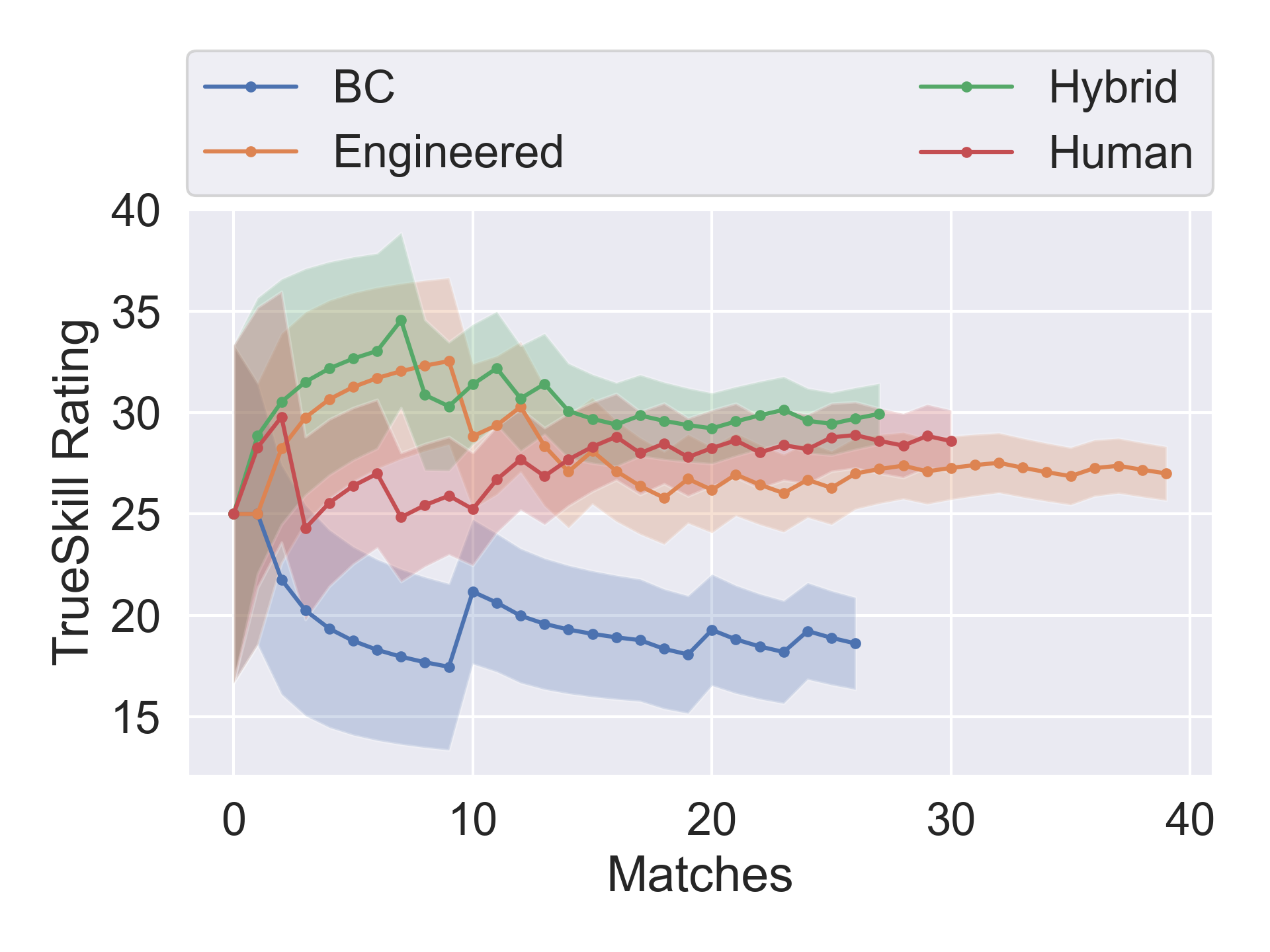}} &
        \subfloat[More Human-like Behavior]{\includegraphics[width=0.31\linewidth]{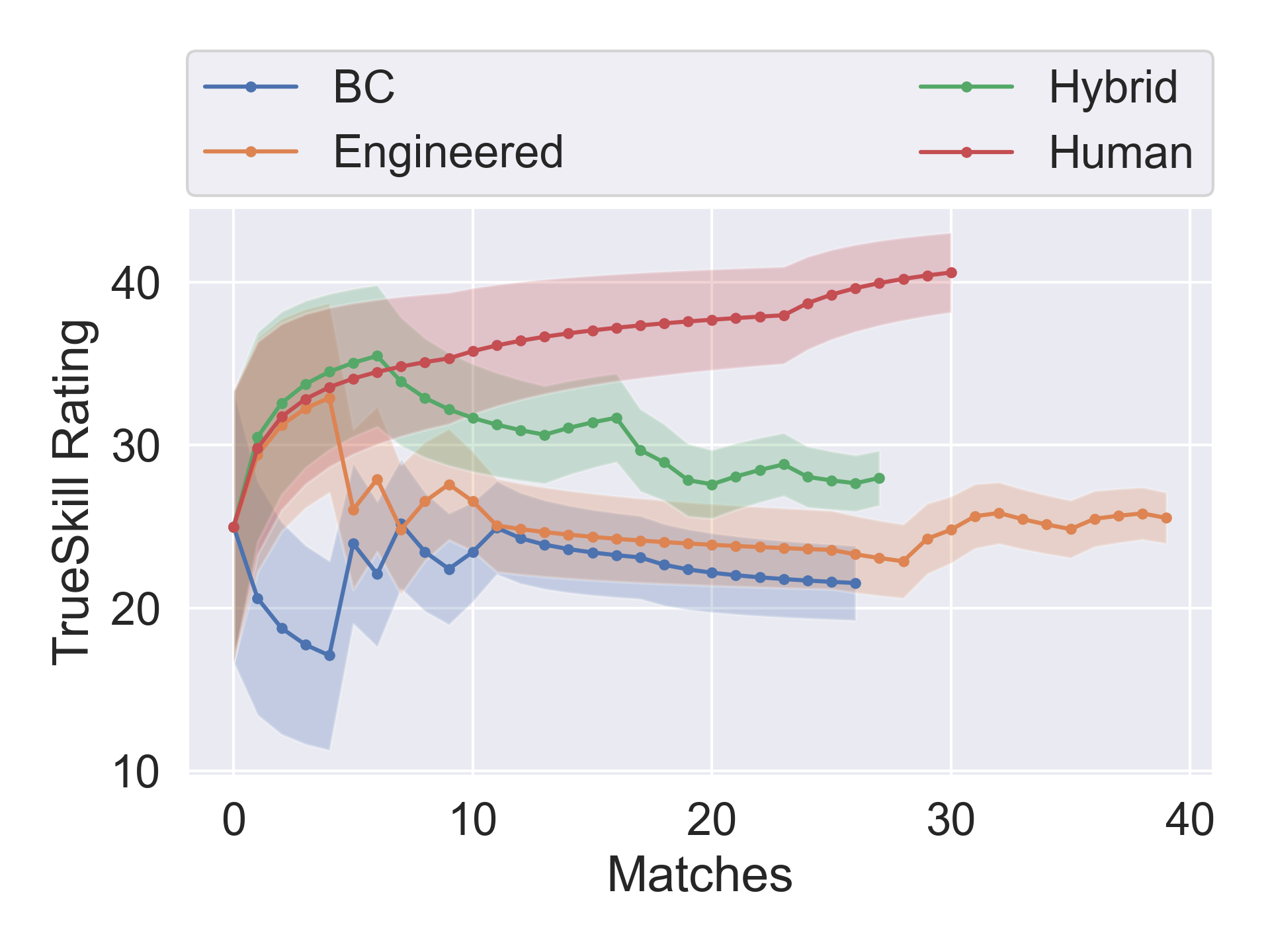}}
    \end{tabular}
  \caption{\textit{TrueSkill}\textsuperscript{TM}\cite{herbrich2006trueskill} scores computed from human evaluations separately for each performance metric and for each agent type performing the \textit{CreateVillageAnimalPen} task.}
  \label{fig:pen_trueskill}
\end{figure}

\begin{figure}[!ht]
  \centering
    \begin{tabular}{ccc}
        \subfloat[Best Performer]{\includegraphics[width=0.31\linewidth]{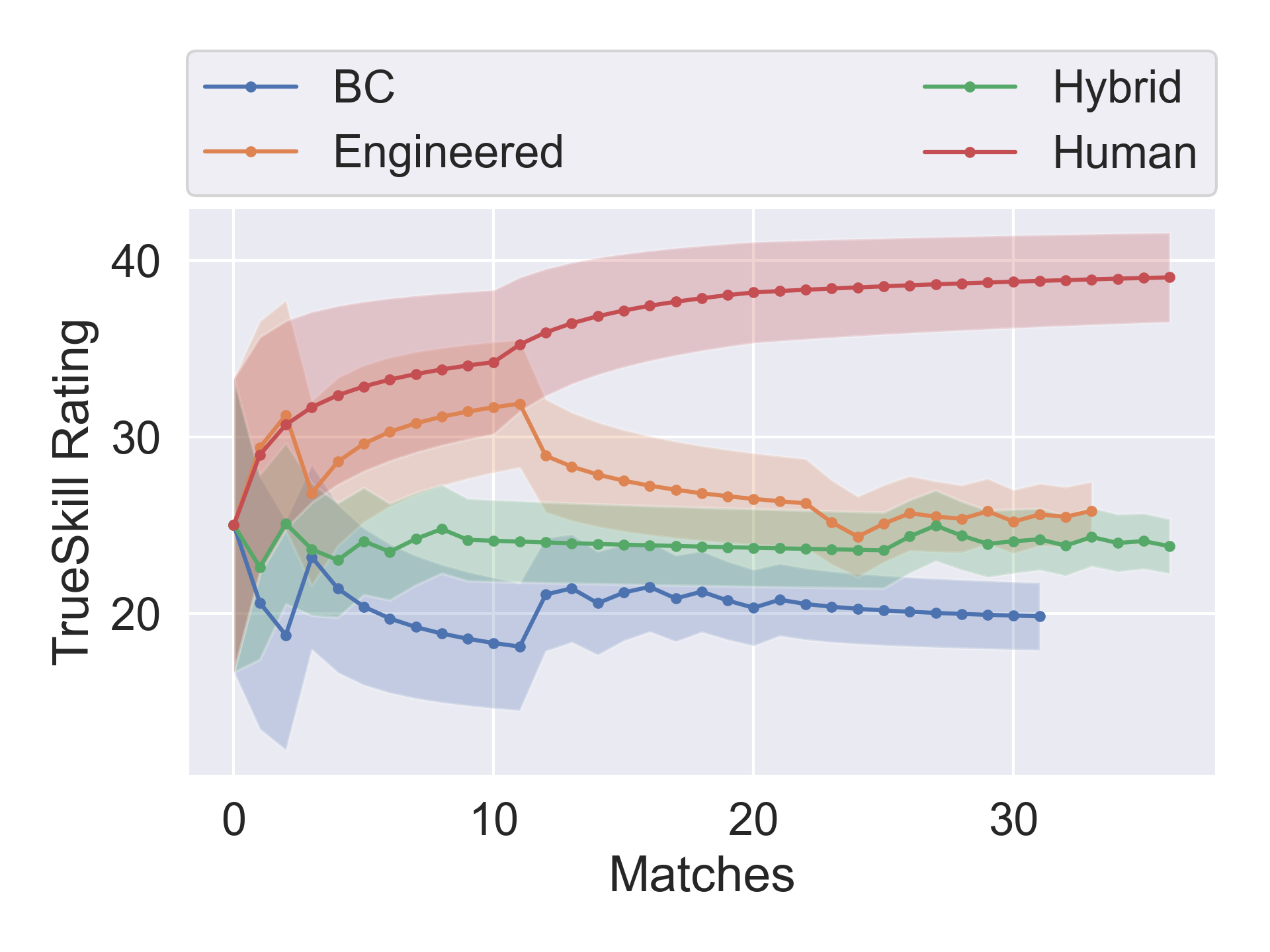}} &
        \subfloat[Fastest Performer]{\includegraphics[width=0.31\linewidth]{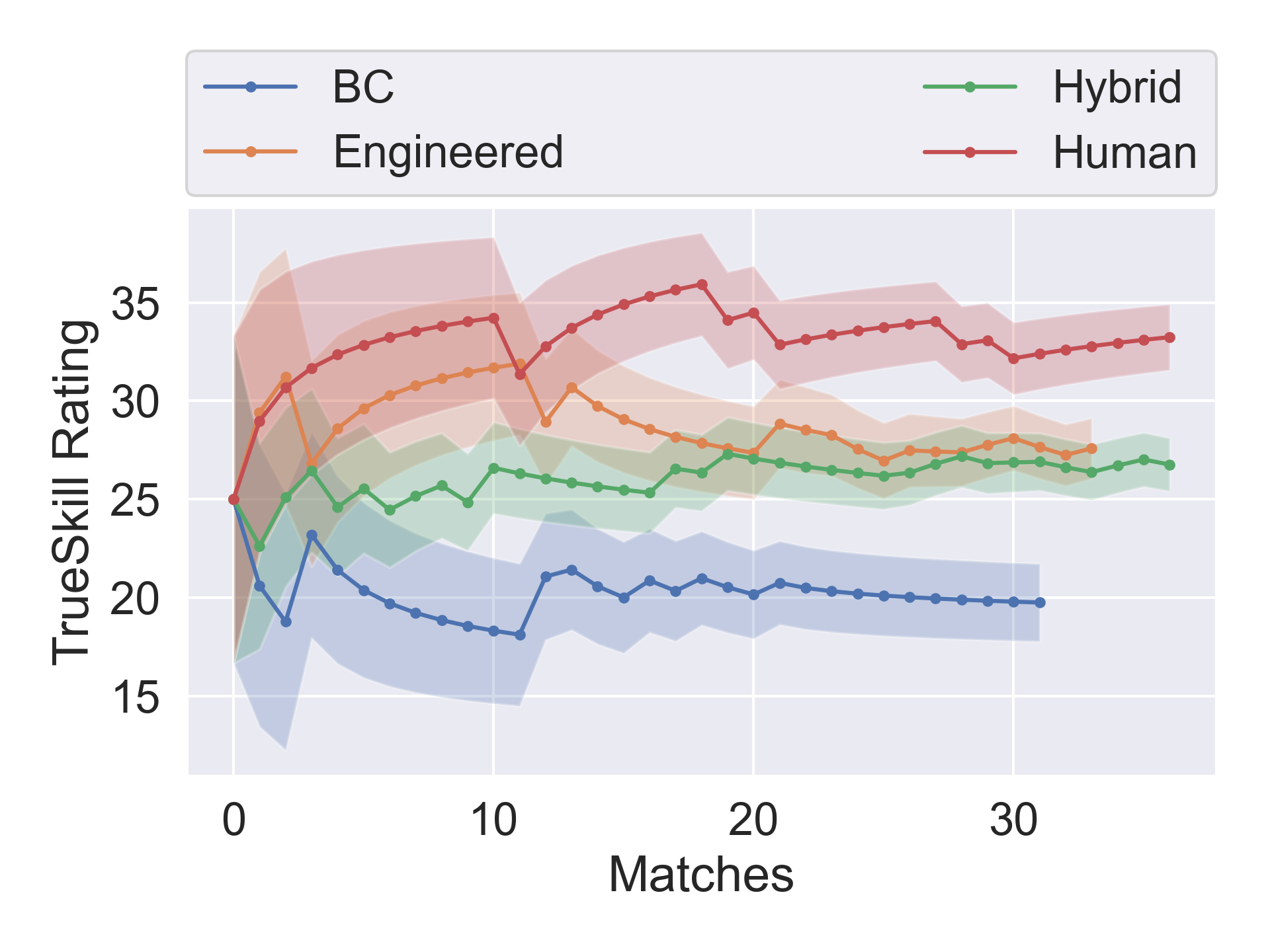}} &
        \subfloat[More Human-like Behavior]{\includegraphics[width=0.31\linewidth]{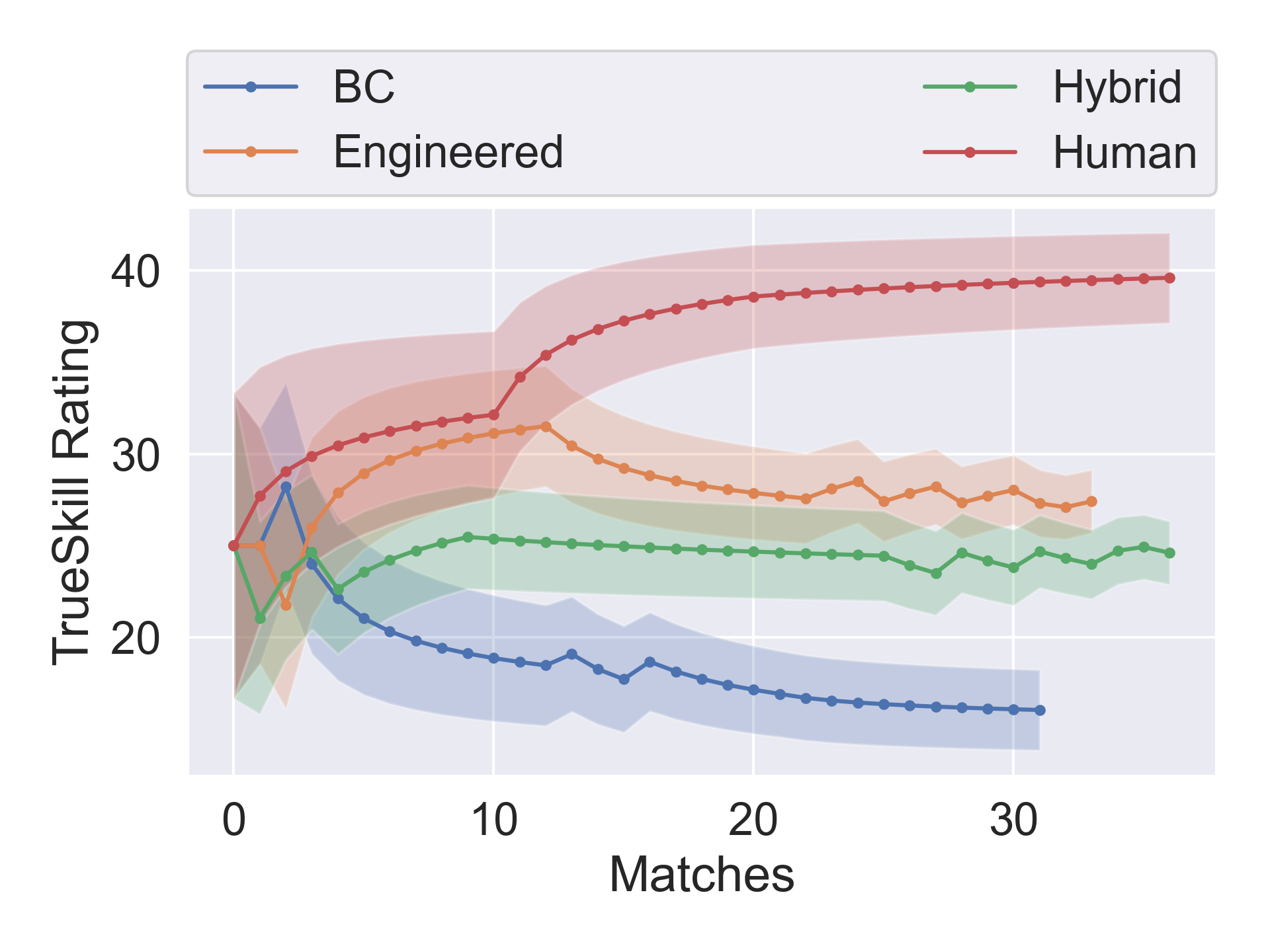}}
    \end{tabular}
  \caption{\textit{TrueSkill}\textsuperscript{TM}\cite{herbrich2006trueskill} scores computed from human evaluations separately for each performance metric and for each agent type performing the \textit{BuildVillageHouse} task.}
  \label{fig:house_trueskill}
\end{figure}

Figures \ref{fig:cave_trueskill}, \ref{fig:waterfall_trueskill}, \ref{fig:pen_trueskill}, and \ref{fig:house_trueskill} show the evolution of the \textit{TrueSkill}\textsuperscript{TM} scores after each match (one-to-one comparison between different agent types) for each performance metric when the agents are solving the \textit{FindCave}, \textit{MakeWaterfall}, \textit{CreateVillageAnimalPen}, and \textit{BuildVillageHouse} tasks, respectively.
The bold line represents the mean estimated skill rating and shaded area the standard deviation of the estimation.

\section{Pairwise Comparison per Performance Metric and Task}\label{appendix:barplot}

\begin{figure}[!ht]
  \centering
    \begin{tabular}{ccc}
        \subfloat[BC vs Engineered]{\includegraphics[width=0.3\linewidth]{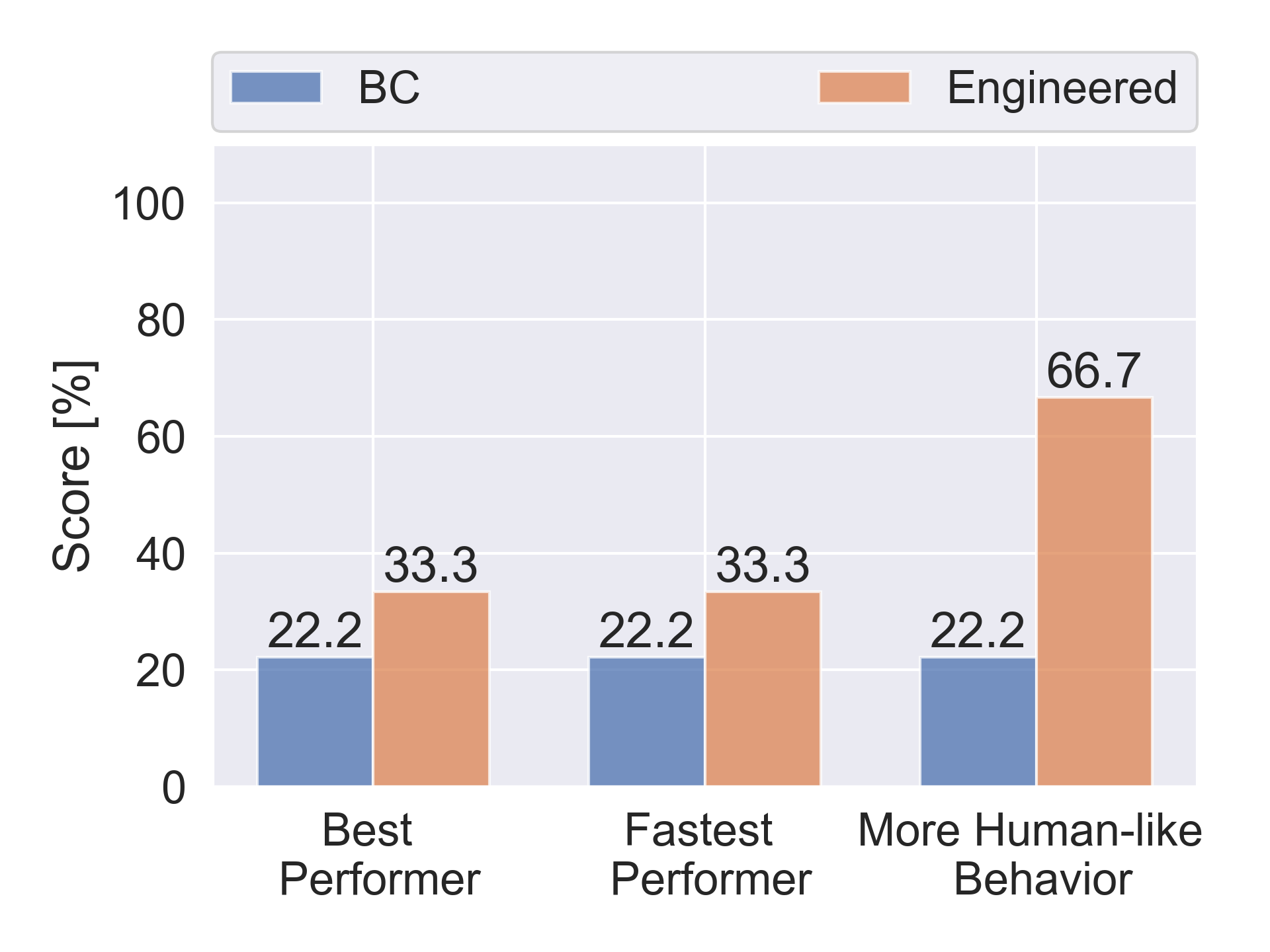}} &
        \subfloat[BC vs Human]{\includegraphics[width=0.3\linewidth]{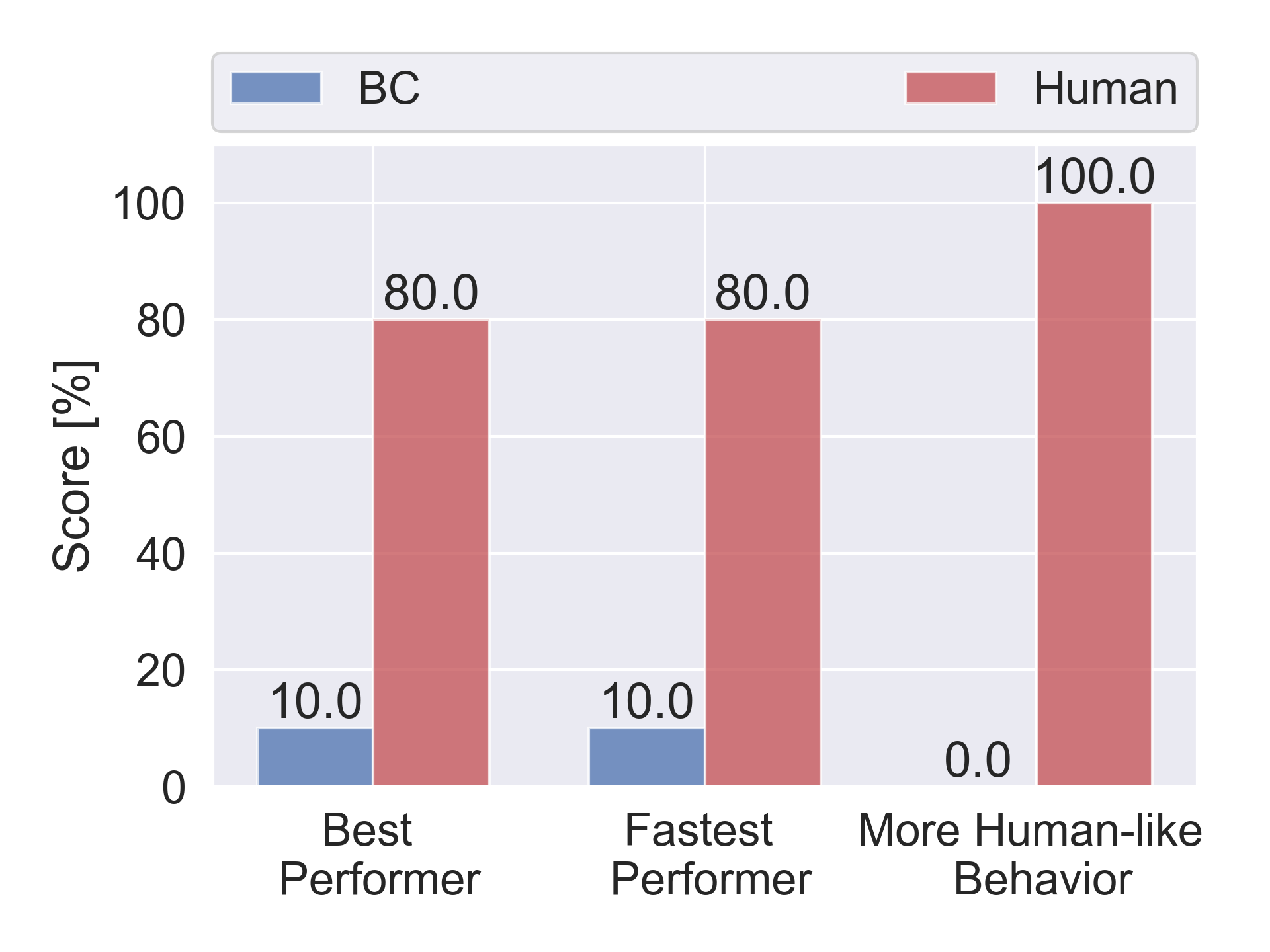}} &
        \subfloat[BC vs Hybrid]{\includegraphics[width=0.3\linewidth]{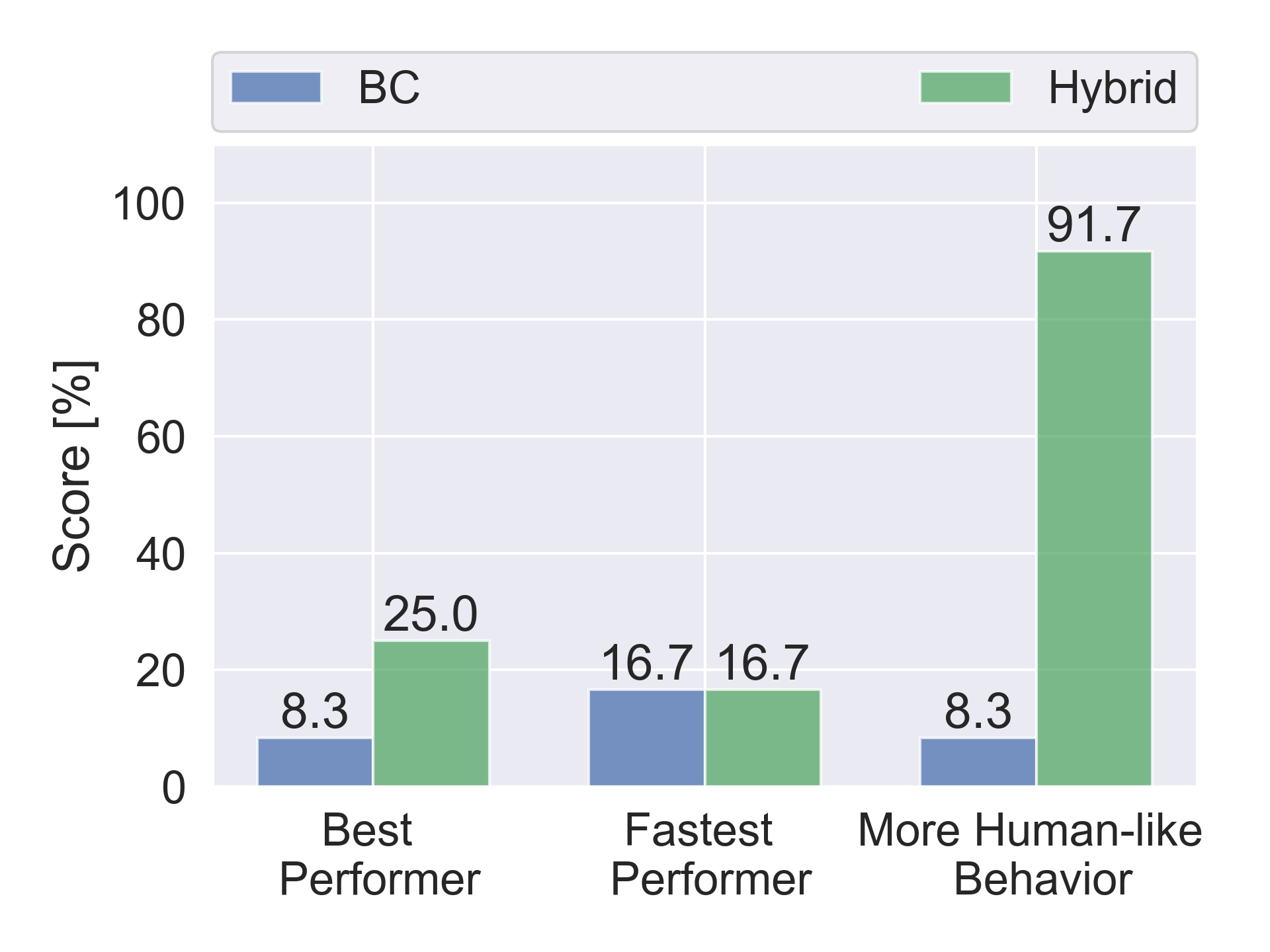}}\\
        \subfloat[Engineered vs Hybrid]{\includegraphics[width=0.3\linewidth]{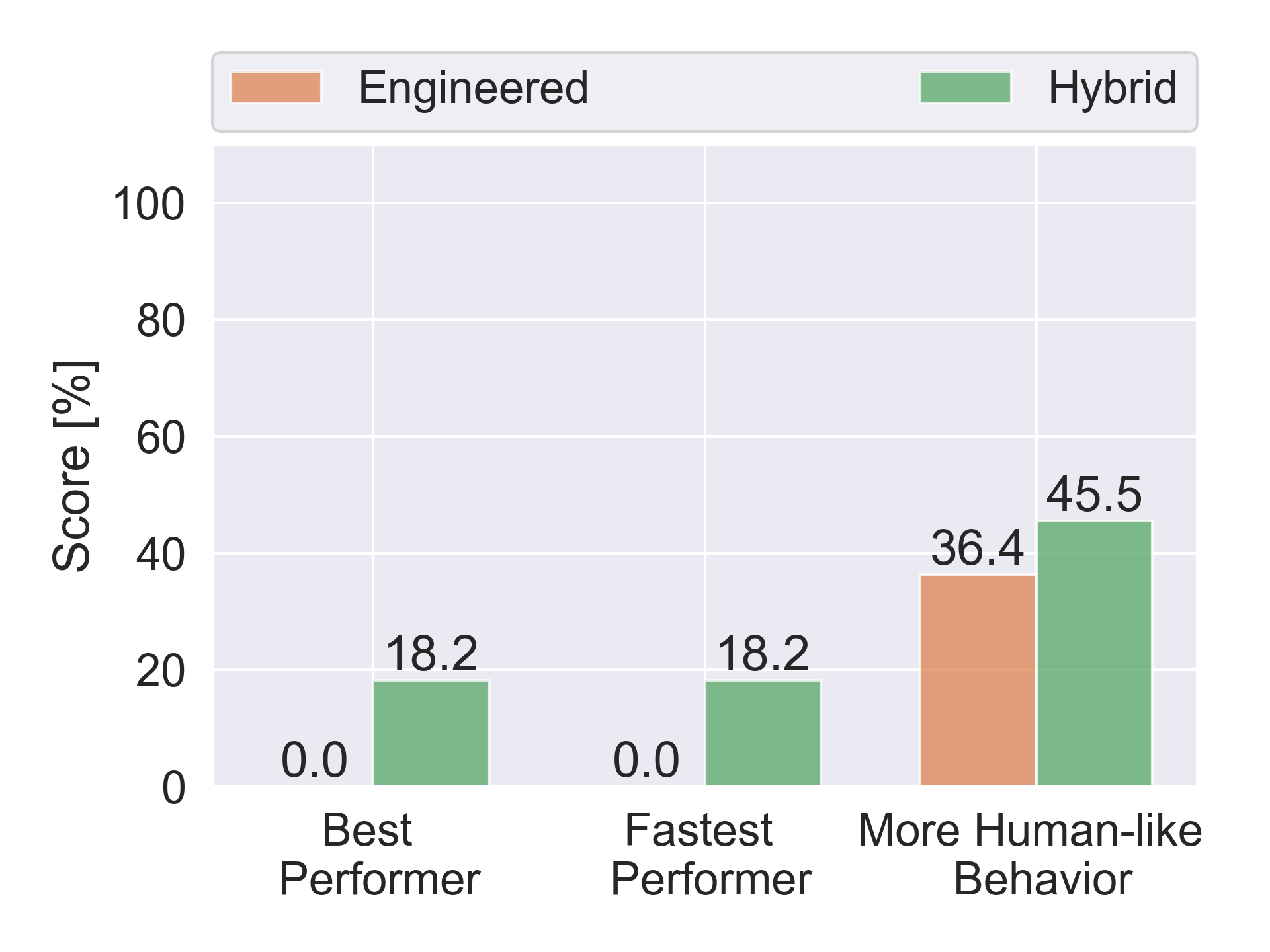}} &
        \subfloat[Human vs Engineered]{\includegraphics[width=0.3\linewidth]{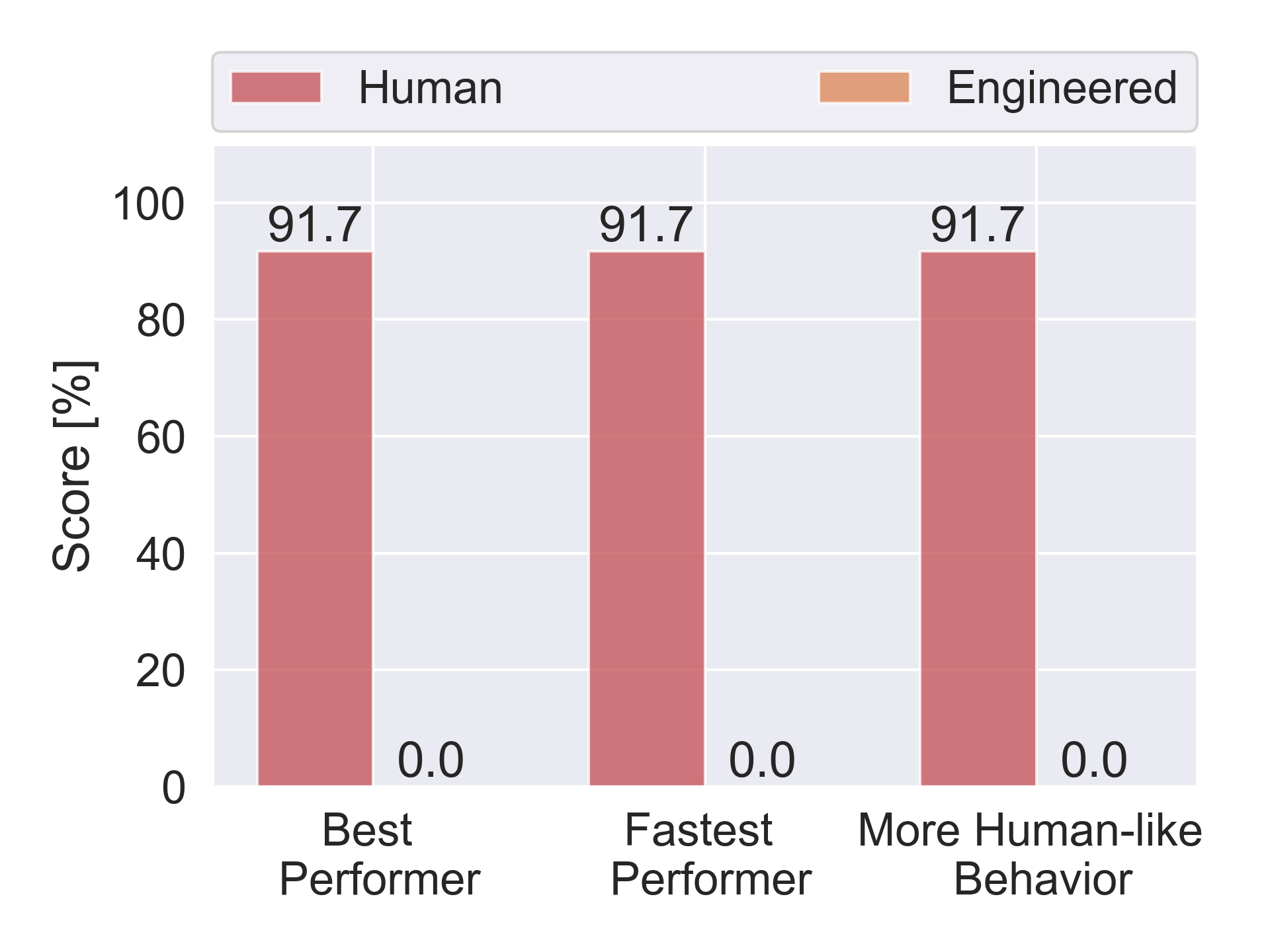}} &
        \subfloat[Human vs Hybrid]{\includegraphics[width=0.3\linewidth]{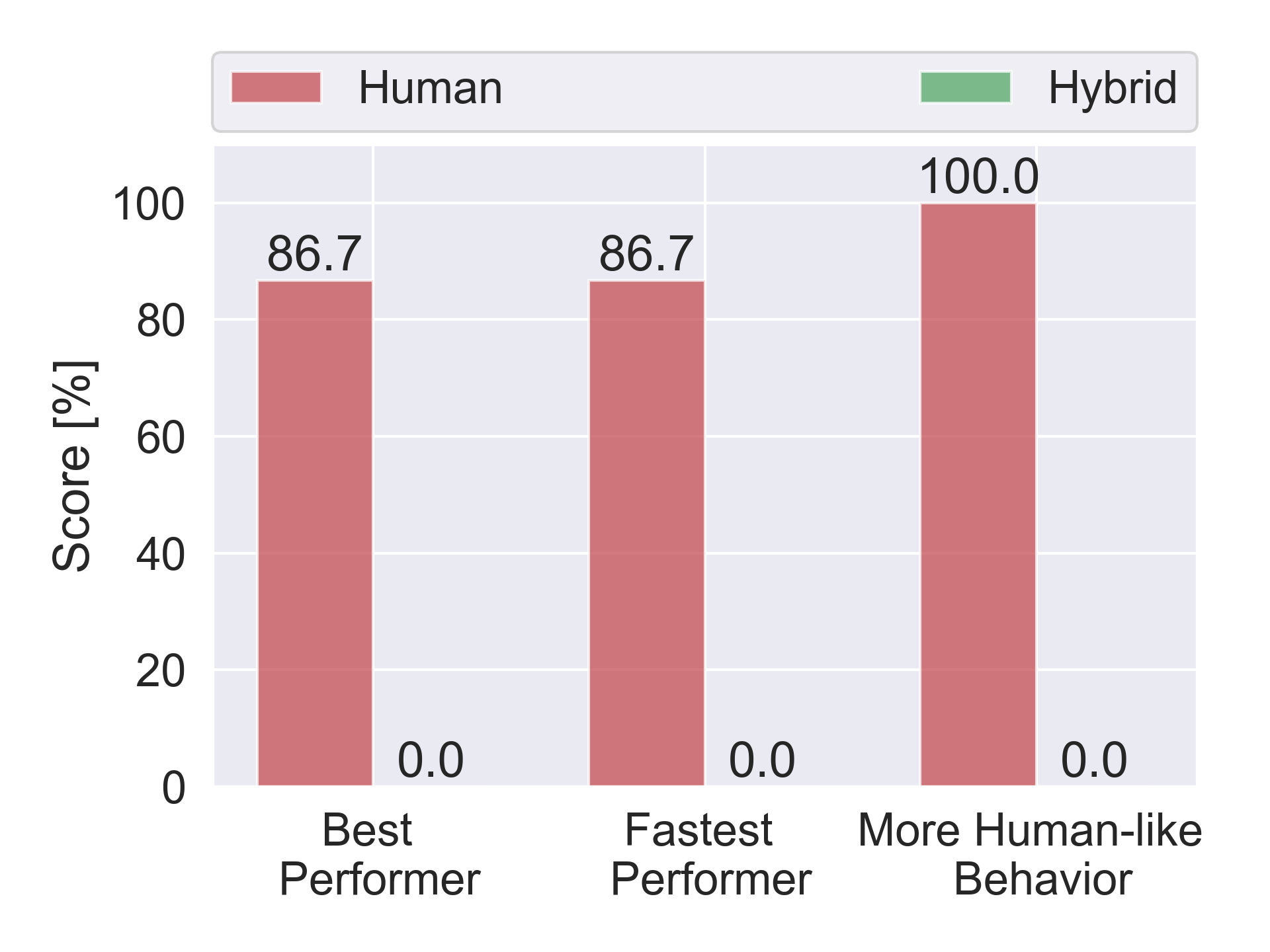}}
    \end{tabular}
  \caption{Pairwise comparison displaying the normalized scores computed from human evaluations separately for each performance metric on all possible head-to-head comparisons for all agent type performing the \textit{FindCave} task.}
  \label{fig:cave_barplot}
\end{figure}

\begin{figure}[!ht]
  \centering
    \begin{tabular}{ccc}
        \subfloat[BC vs Engineered]{\includegraphics[width=0.3\linewidth]{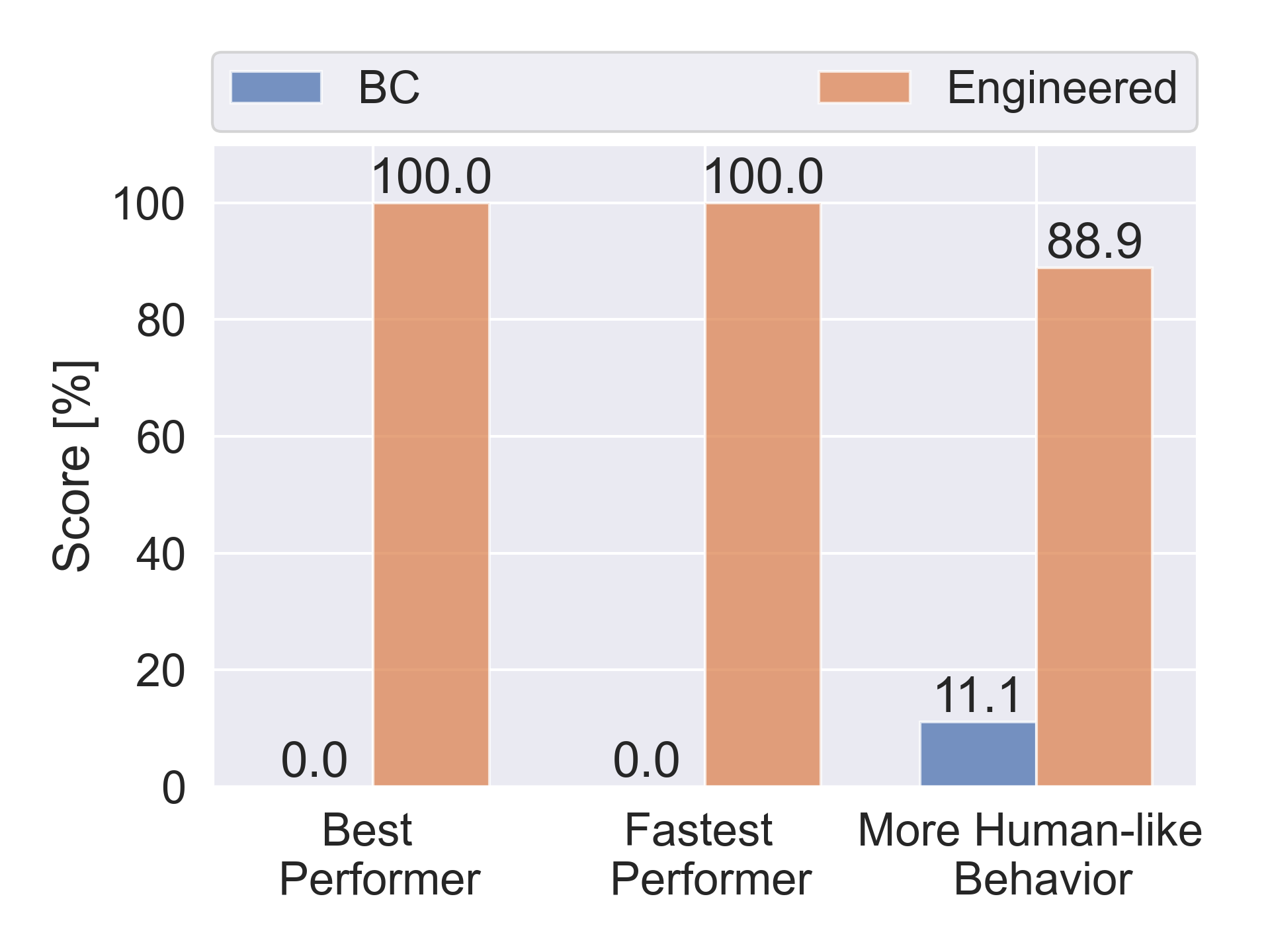}} &
        \subfloat[BC vs Human]{\includegraphics[width=0.3\linewidth]{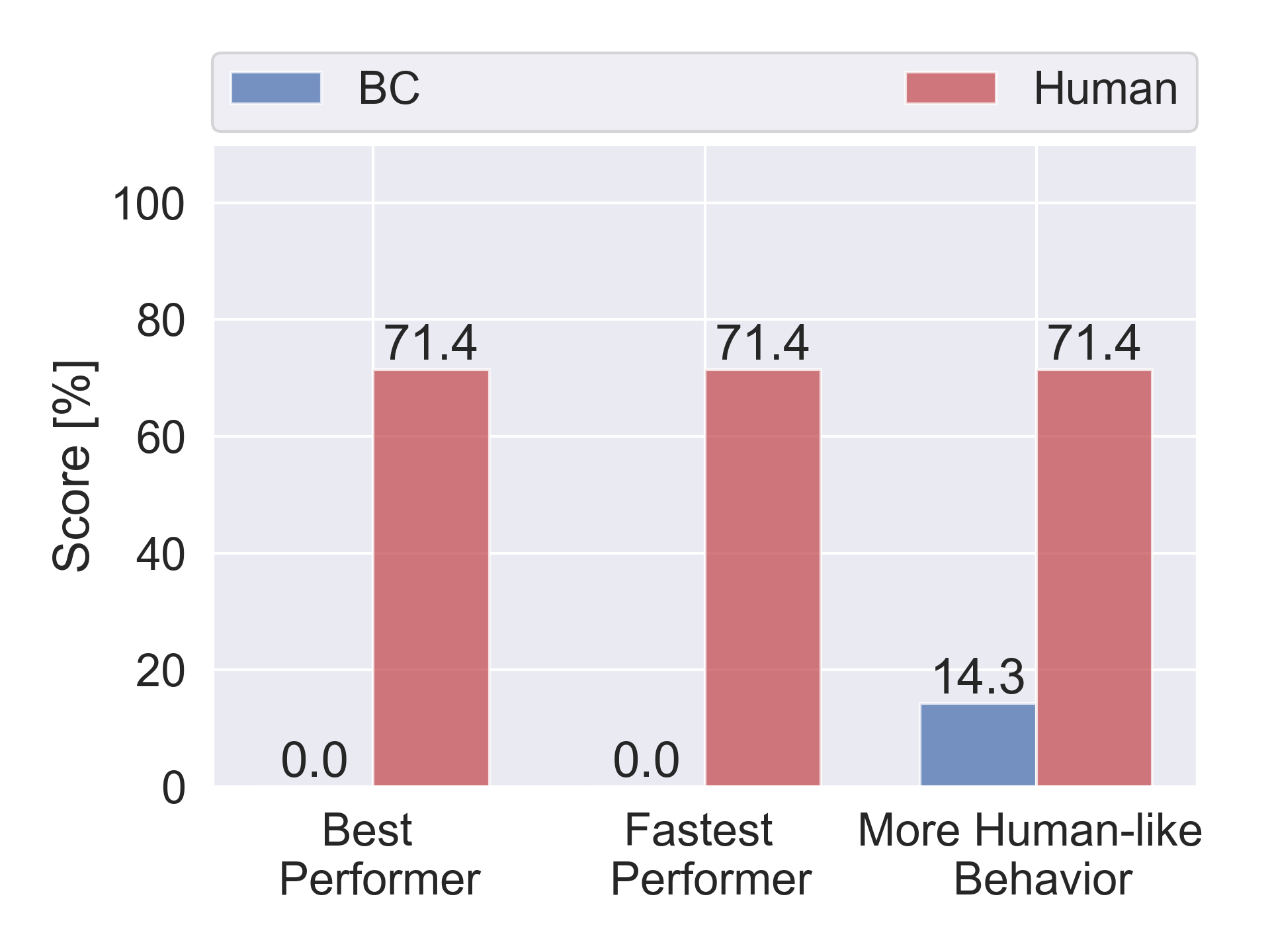}} &
        \subfloat[BC vs Hybrid]{\includegraphics[width=0.3\linewidth]{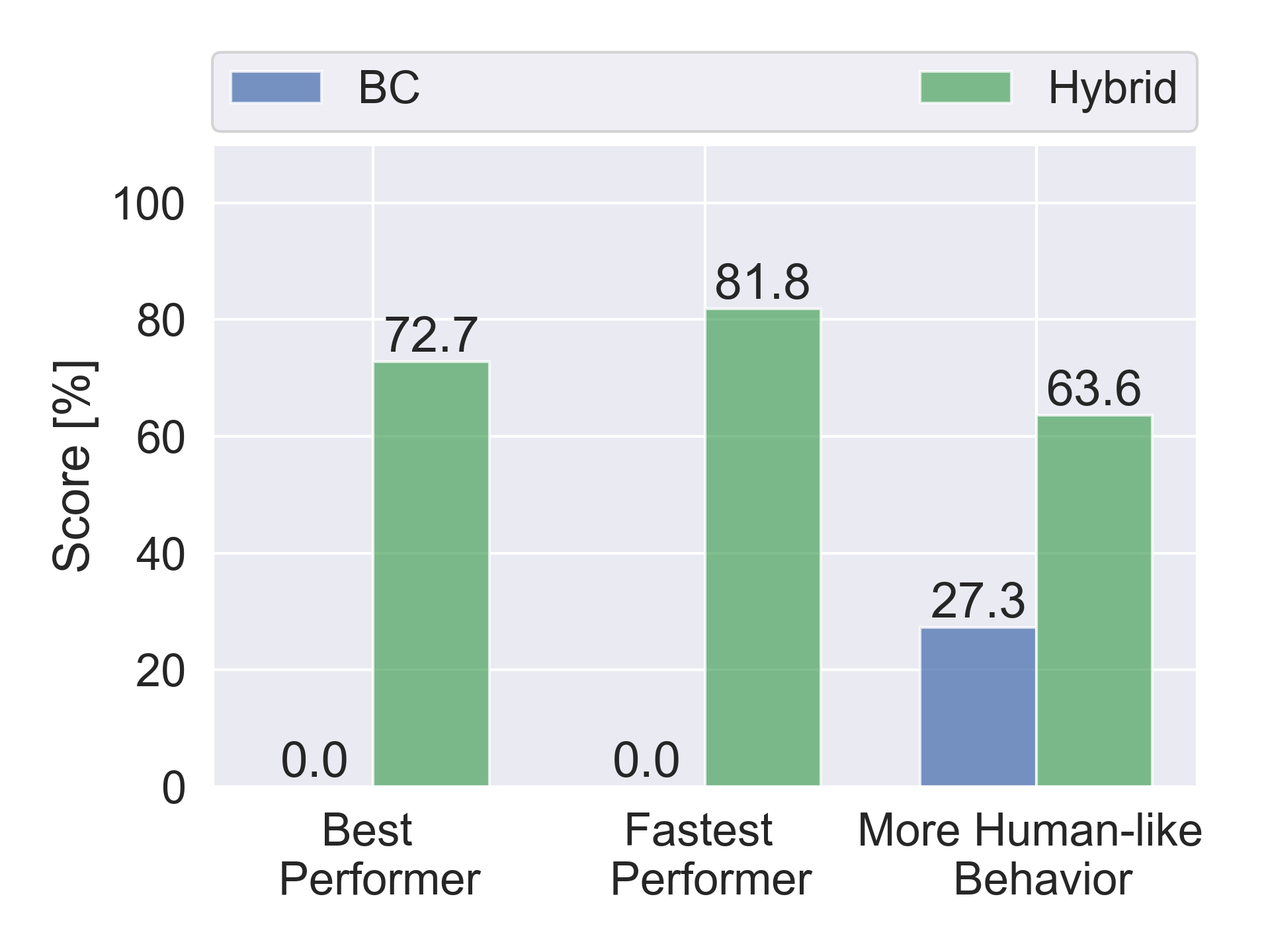}}\\
        \subfloat[Engineered vs Hybrid]{\includegraphics[width=0.3\linewidth]{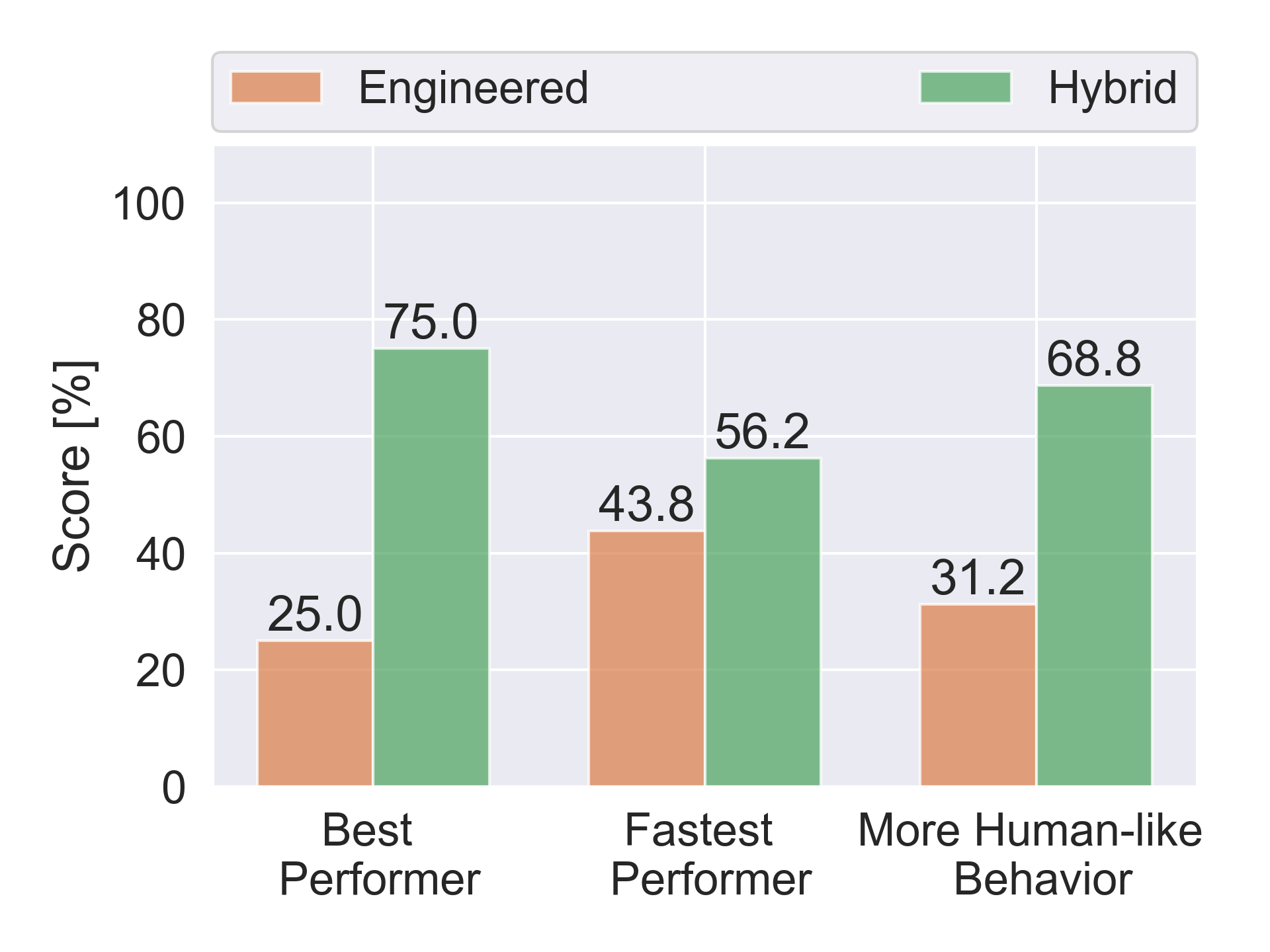}} &
        \subfloat[Human vs Engineered]{\includegraphics[width=0.3\linewidth]{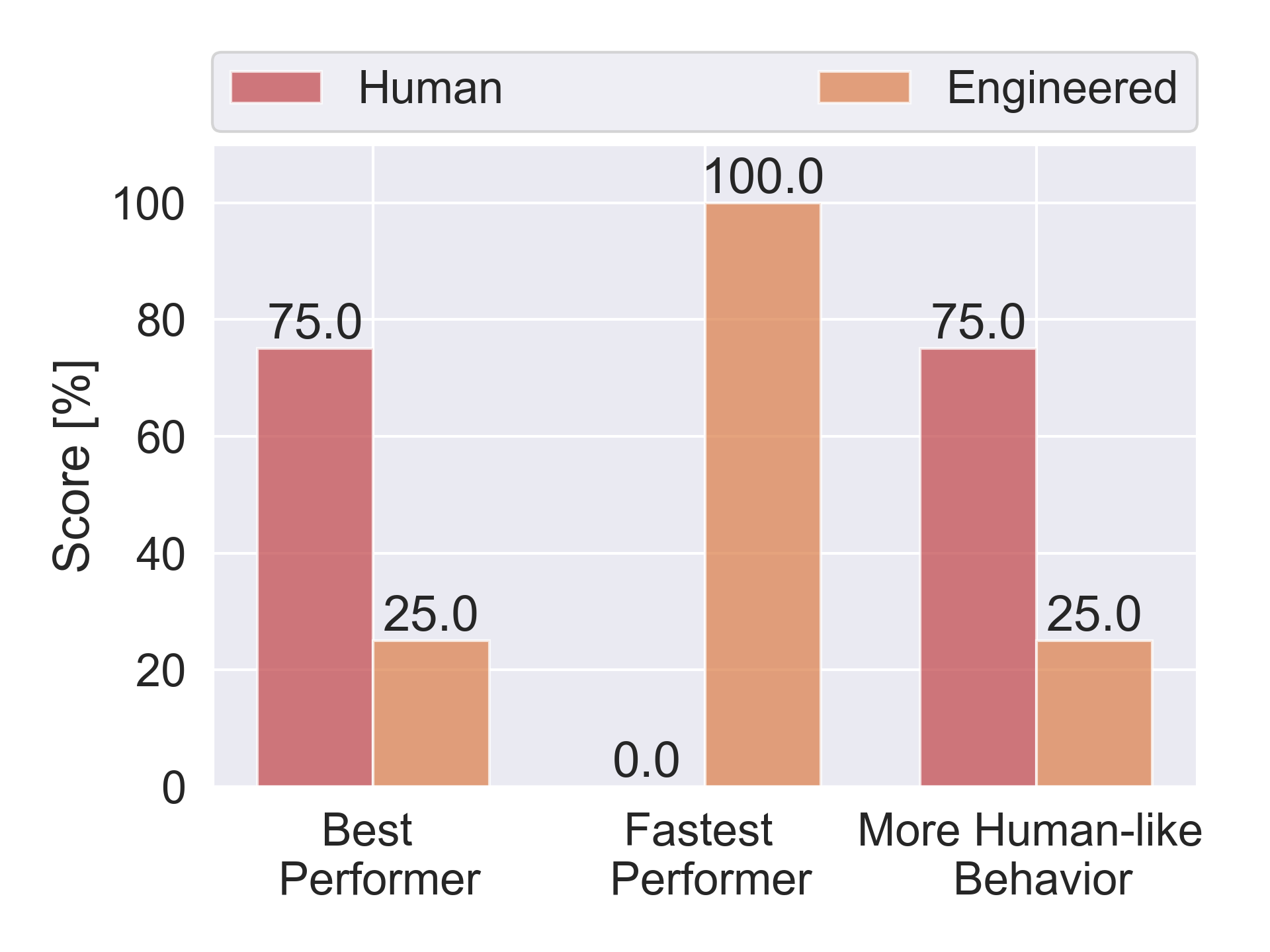}} &
        \subfloat[Human vs Hybrid]{\includegraphics[width=0.3\linewidth]{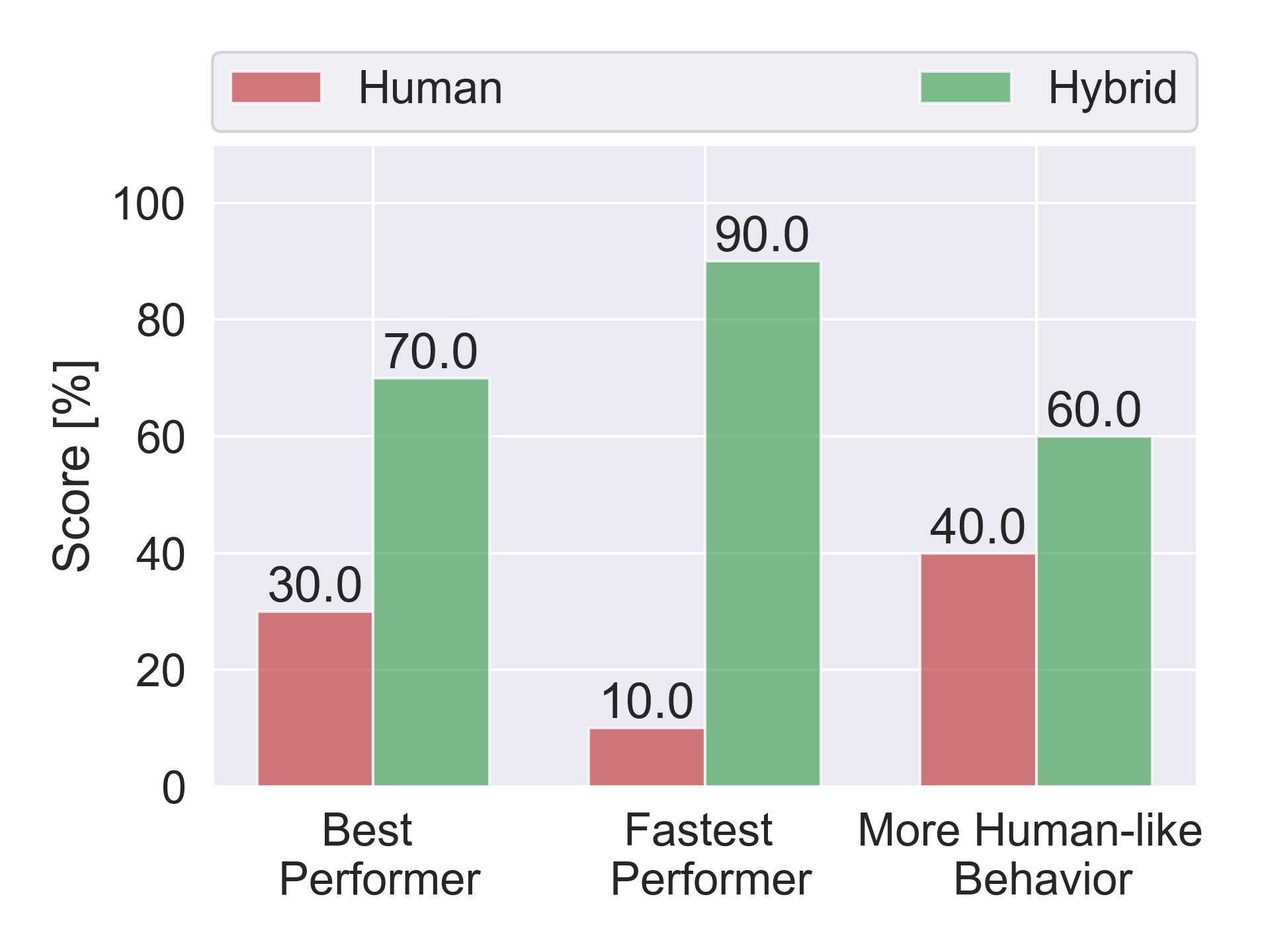}}
    \end{tabular}
  \caption{Pairwise comparison displaying the normalized scores computed from human evaluations separately for each performance metric on all possible head-to-head comparisons for all agent type performing the \textit{MakeWaterfall} task.}
  \label{fig:waterfall_barplot}
\end{figure}

\begin{figure}[!ht]
  \centering
    \begin{tabular}{ccc}
        \subfloat[BC vs Engineered]{\includegraphics[width=0.3\linewidth]{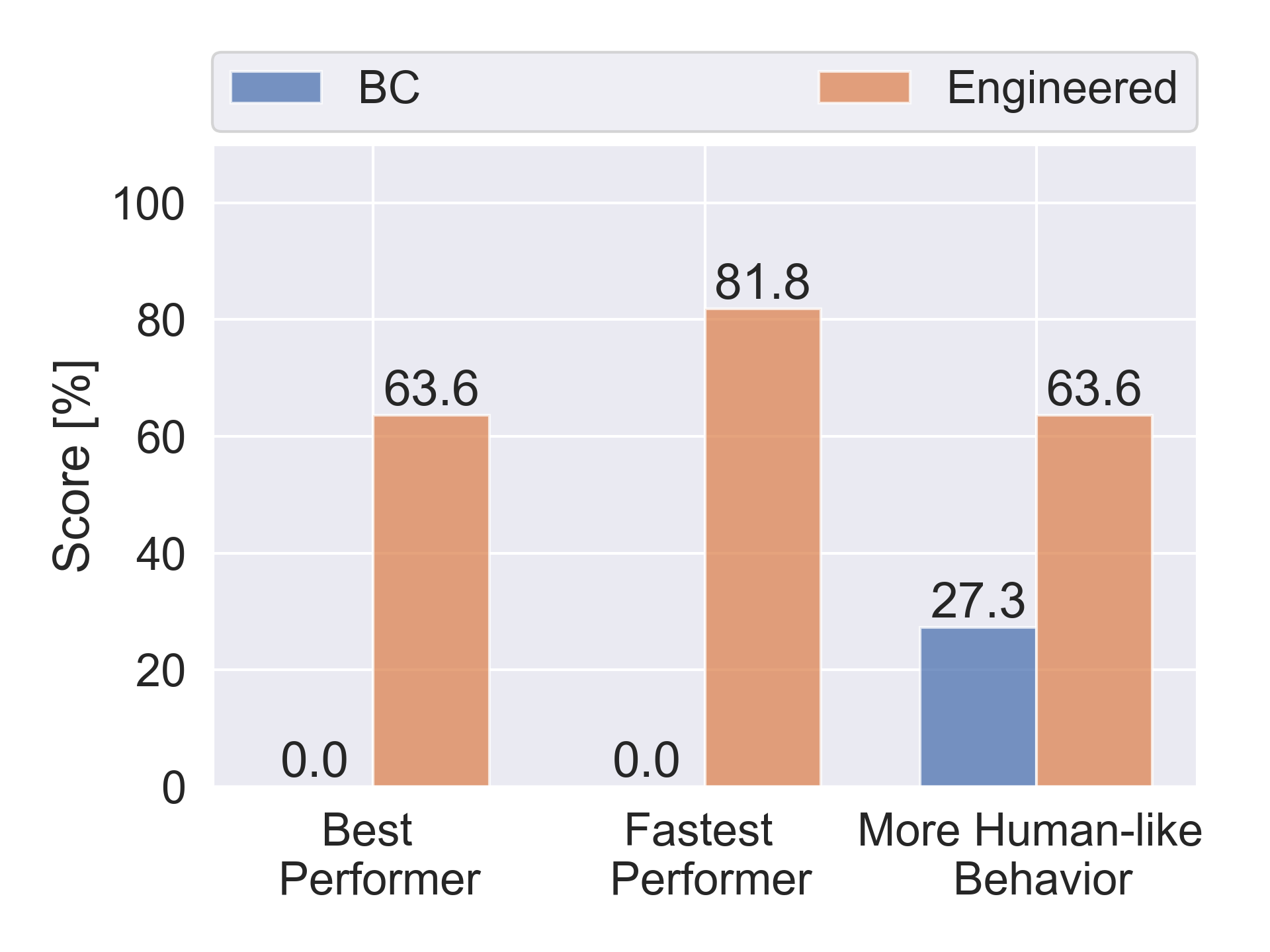}} &
        \subfloat[BC vs Human]{\includegraphics[width=0.3\linewidth]{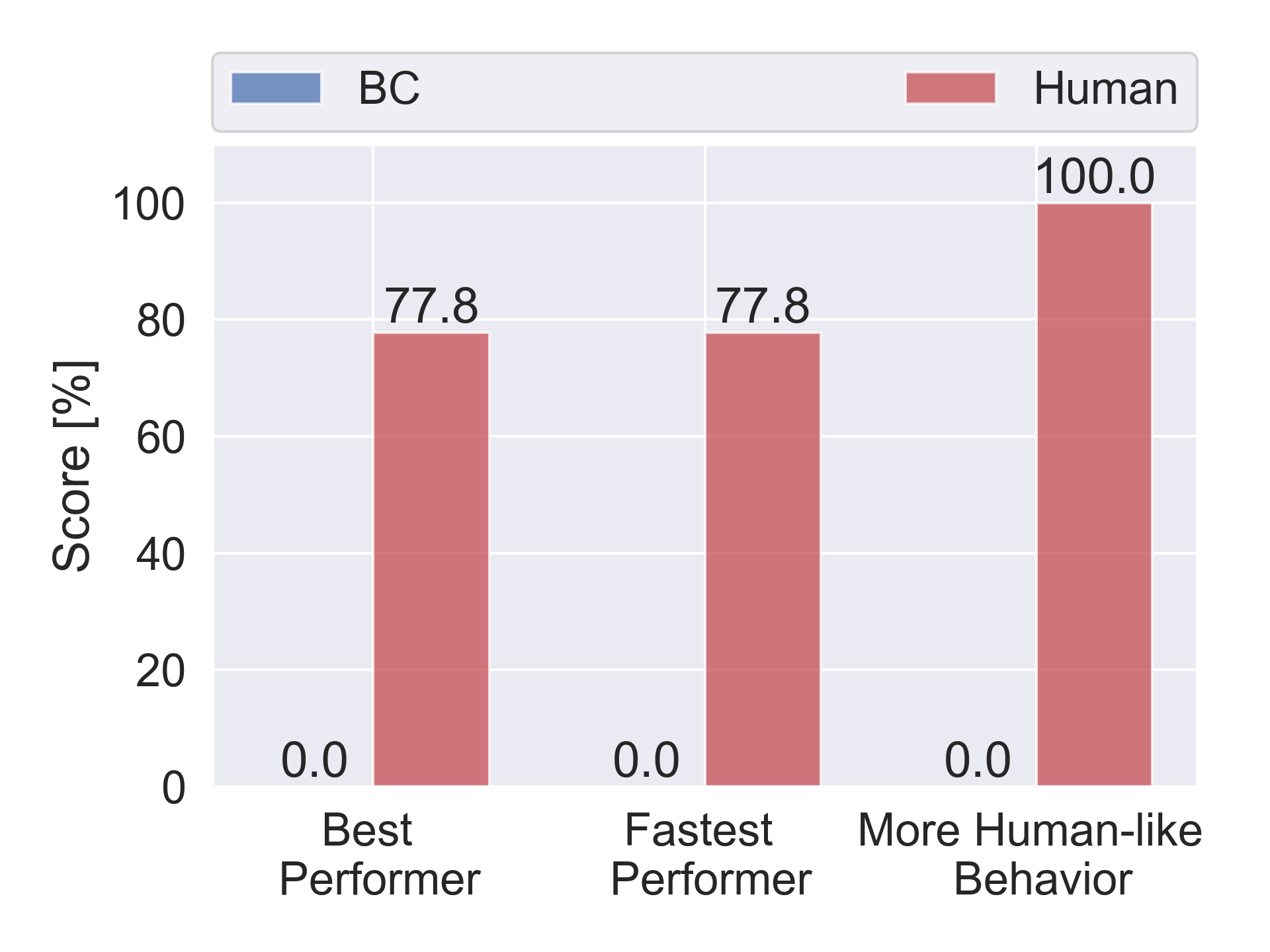}} &
        \subfloat[BC vs Hybrid]{\includegraphics[width=0.3\linewidth]{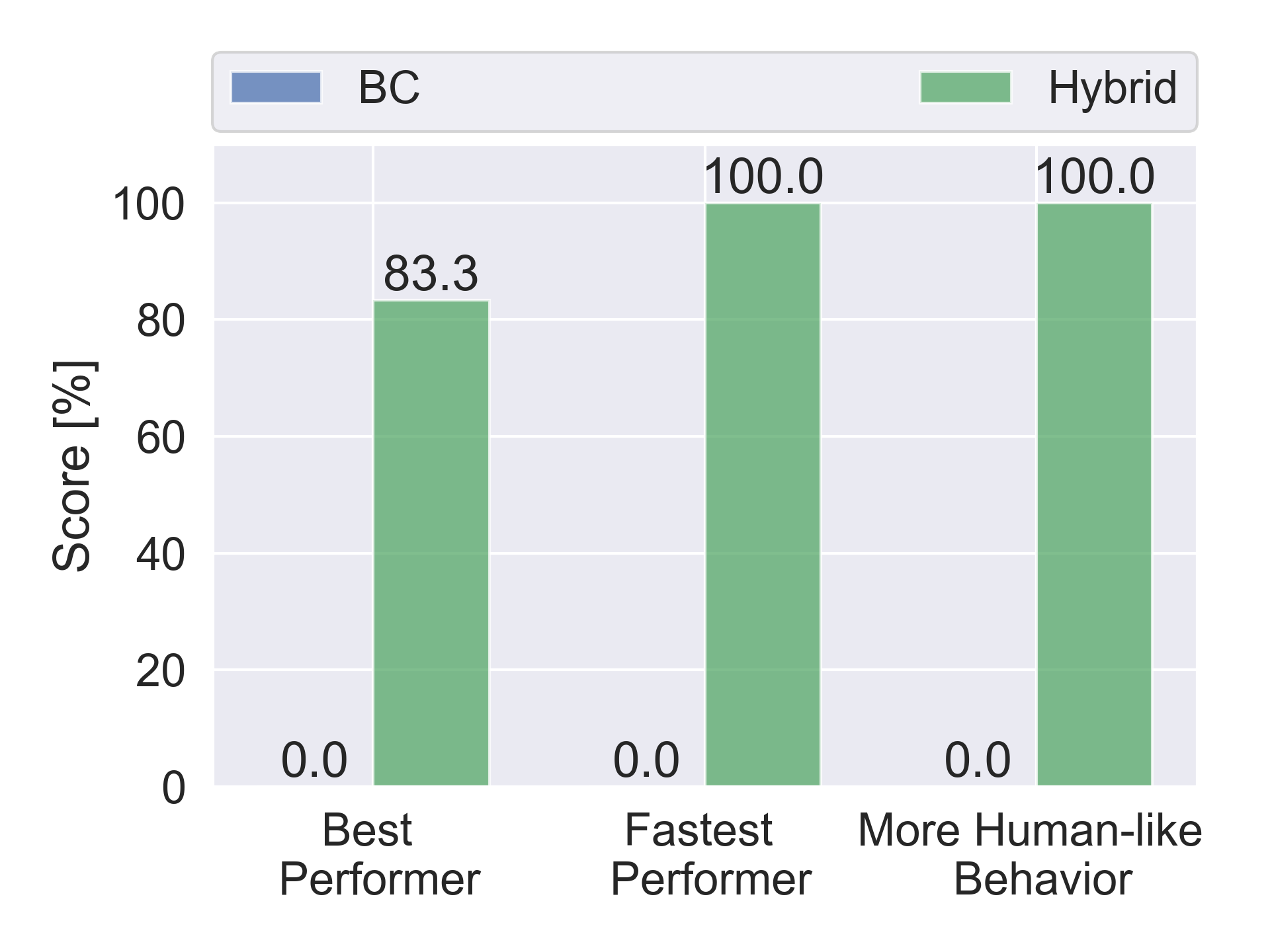}}\\
        \subfloat[Engineered vs Hybrid]{\includegraphics[width=0.3\linewidth]{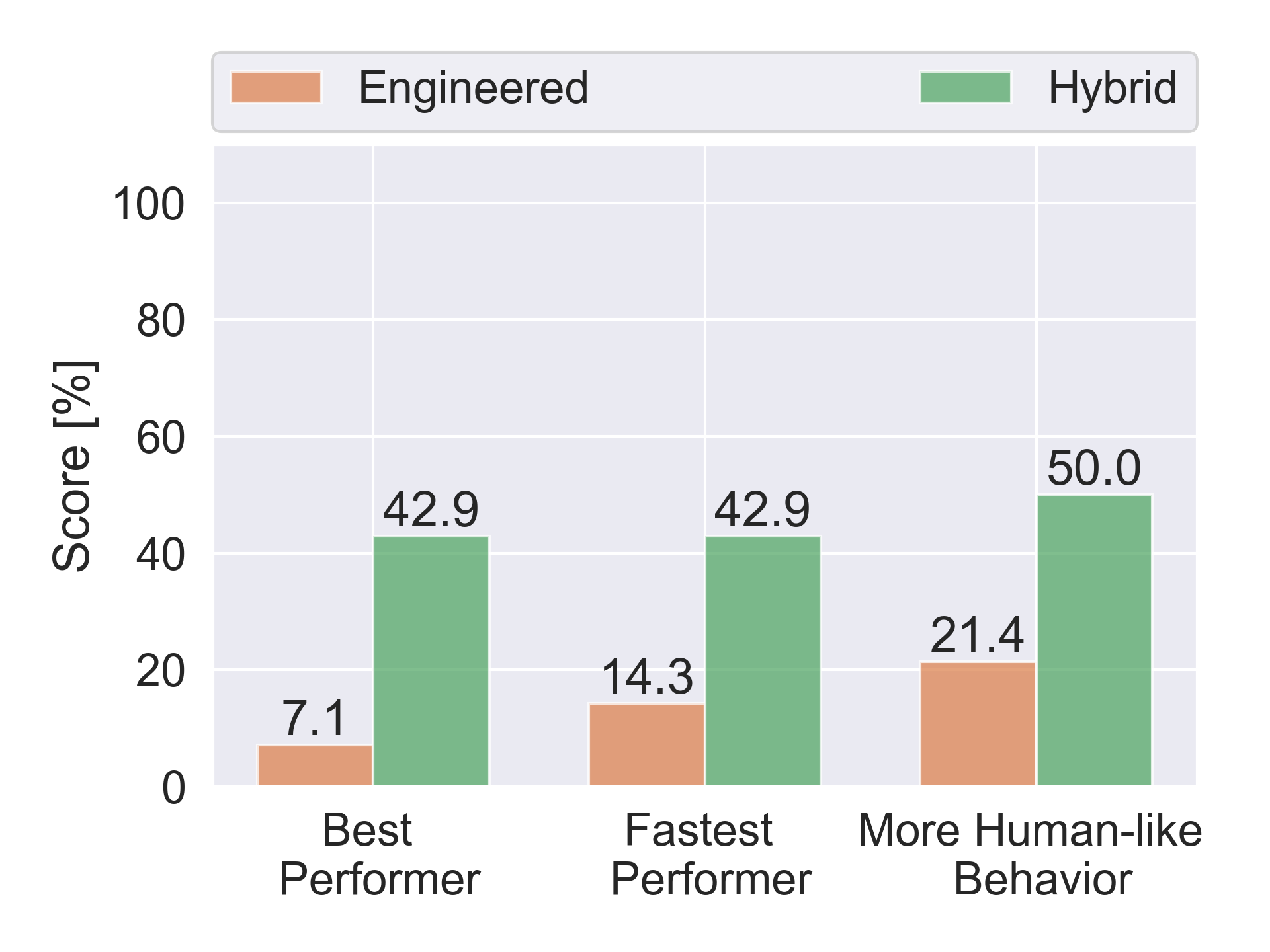}} &
        \subfloat[Human vs Engineered]{\includegraphics[width=0.3\linewidth]{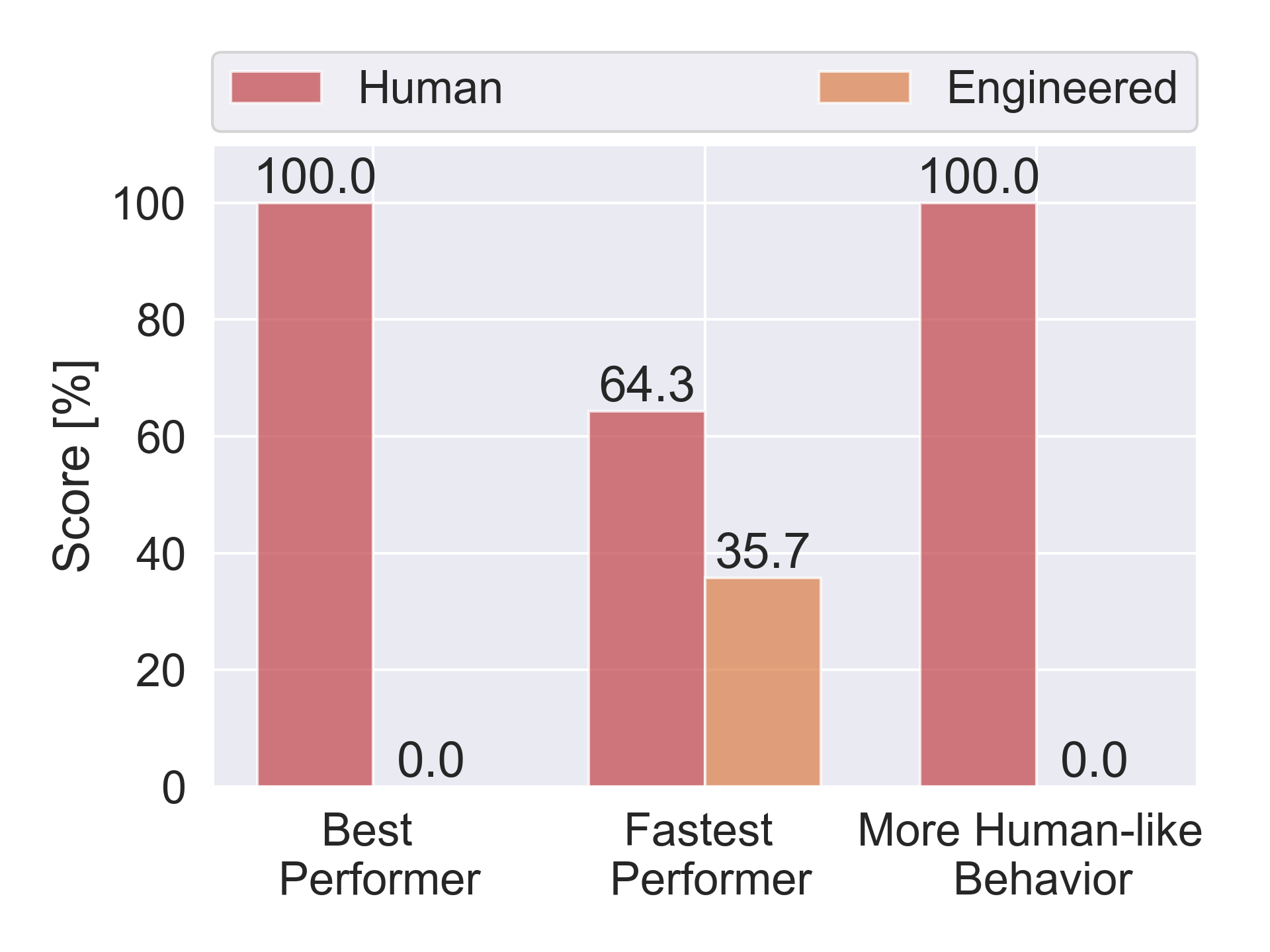}} &
        \subfloat[Human vs Hybrid]{\includegraphics[width=0.3\linewidth]{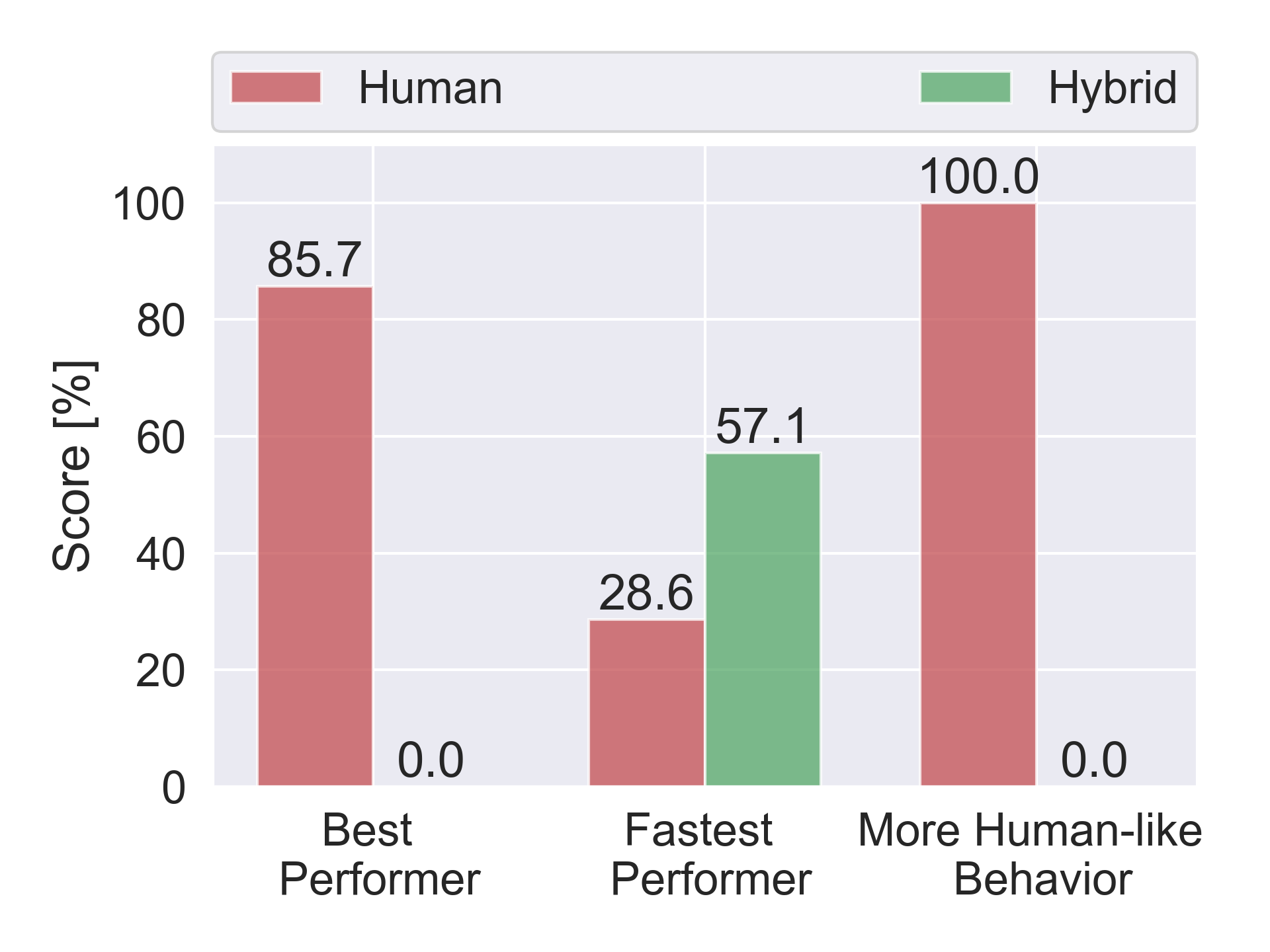}}
    \end{tabular}
  \caption{Pairwise comparison displaying the normalized scores computed from human evaluations separately for each performance metric on all possible head-to-head comparisons for all agent type performing the \textit{CreateVillageAnimalPen} task.}
  \label{fig:pen_barplot}
\end{figure}

\begin{figure}[!ht]
  \centering
    \begin{tabular}{ccc}
        \subfloat[BC vs Engineered]{\includegraphics[width=0.3\linewidth]{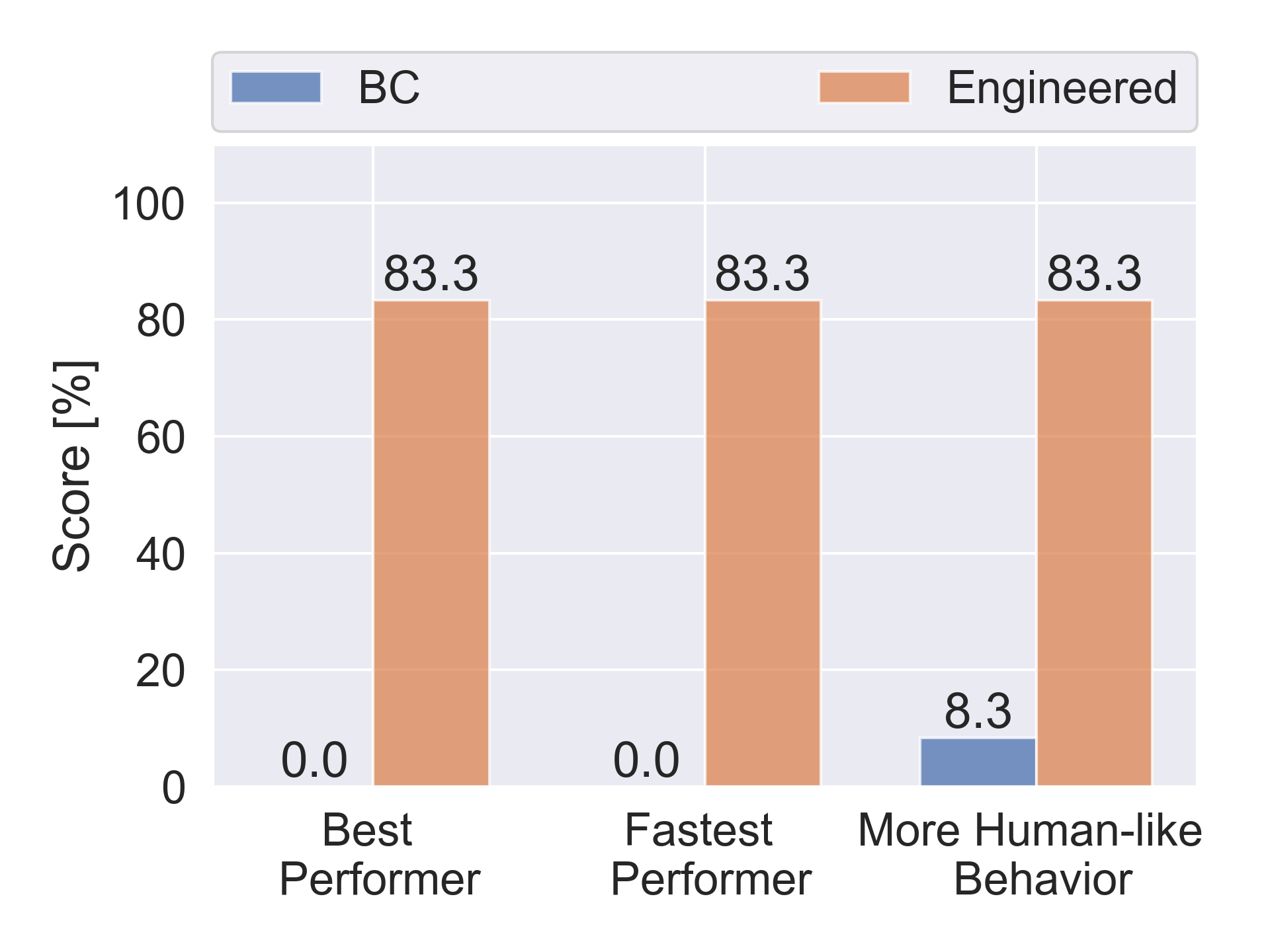}} &
        \subfloat[BC vs Human]{\includegraphics[width=0.3\linewidth]{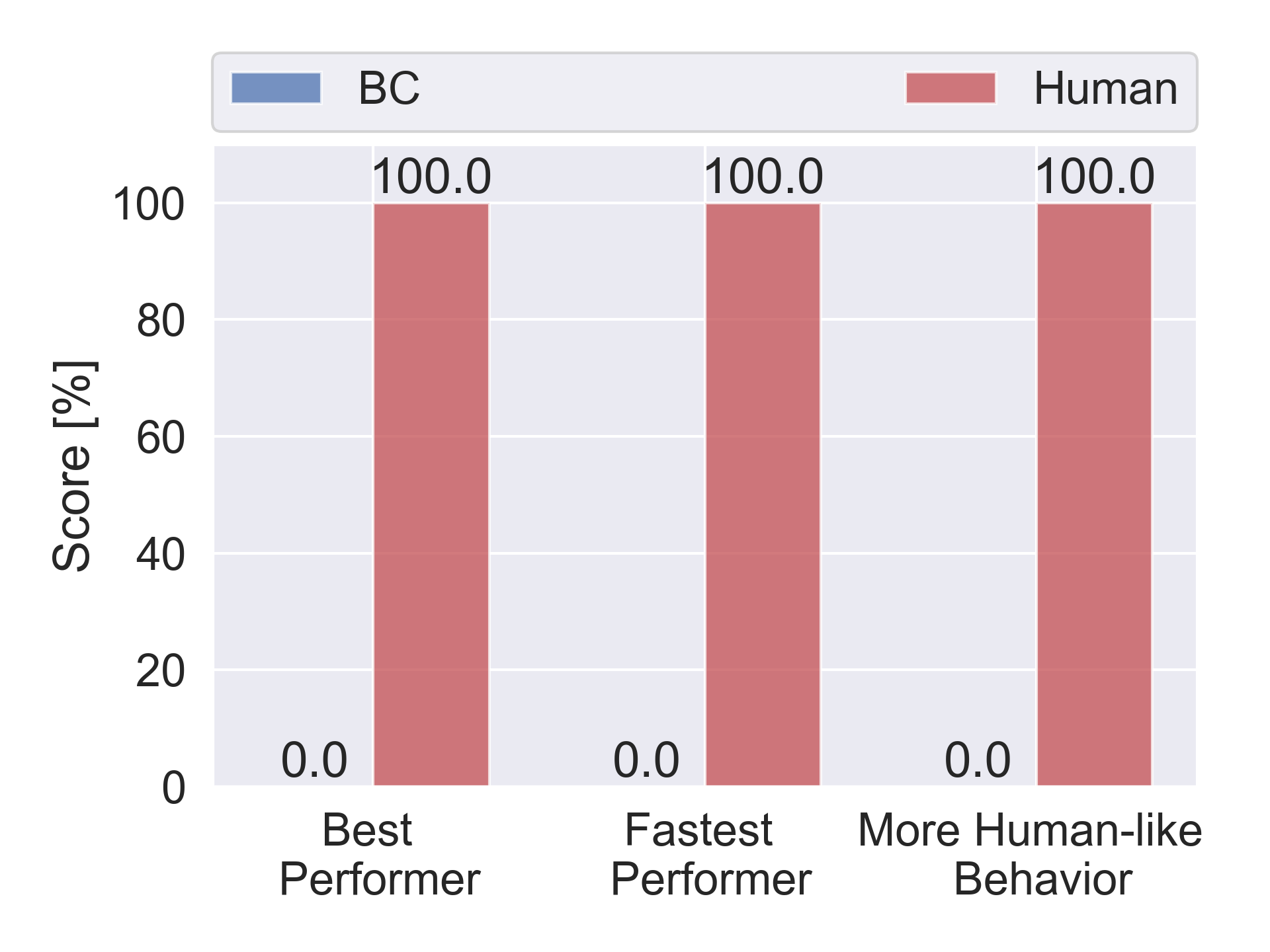}} &
        \subfloat[BC vs Hybrid]{\includegraphics[width=0.3\linewidth]{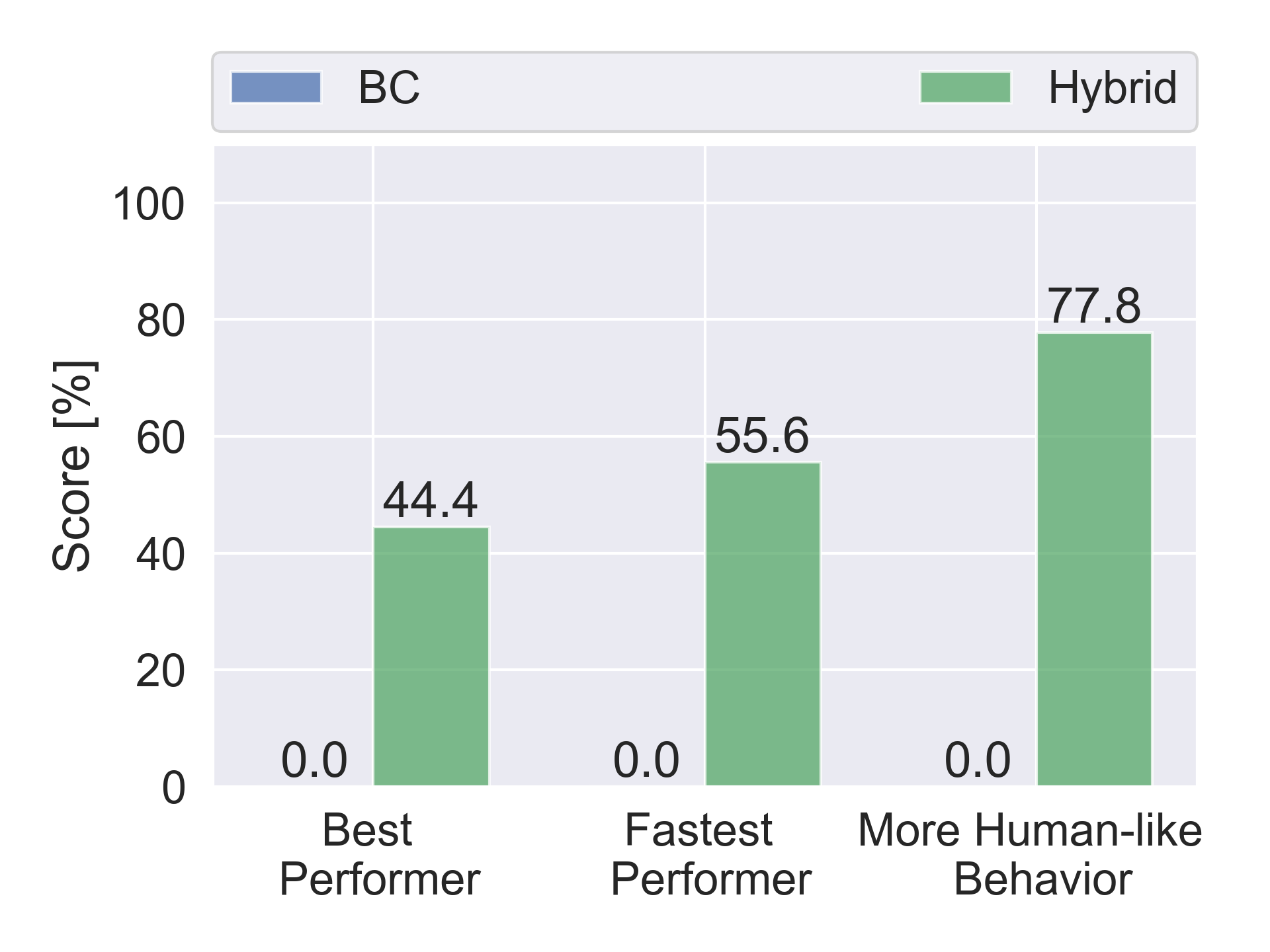}}\\
        \subfloat[Engineered vs Hybrid]{\includegraphics[width=0.3\linewidth]{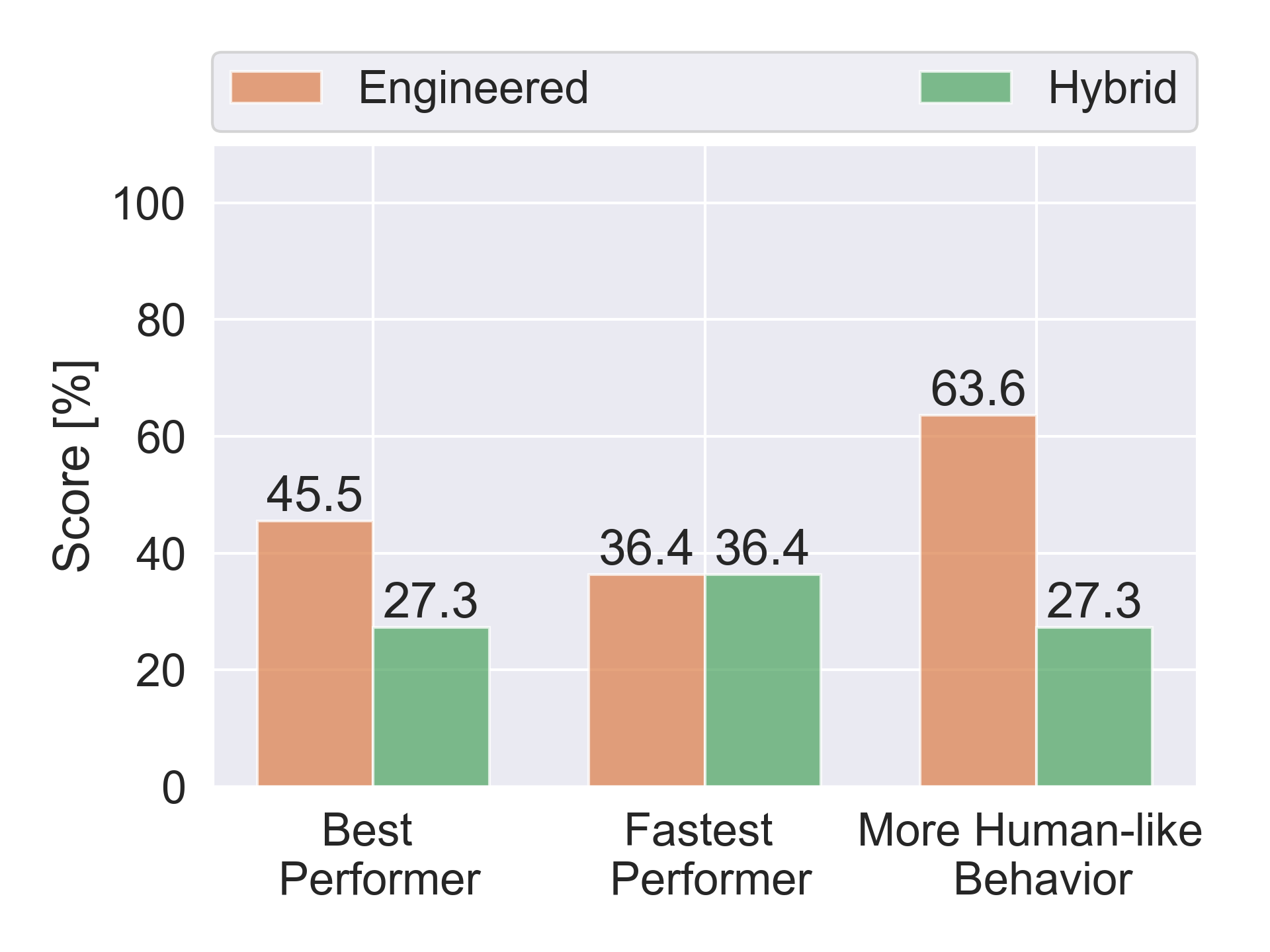}} &
        \subfloat[Human vs Engineered]{\includegraphics[width=0.3\linewidth]{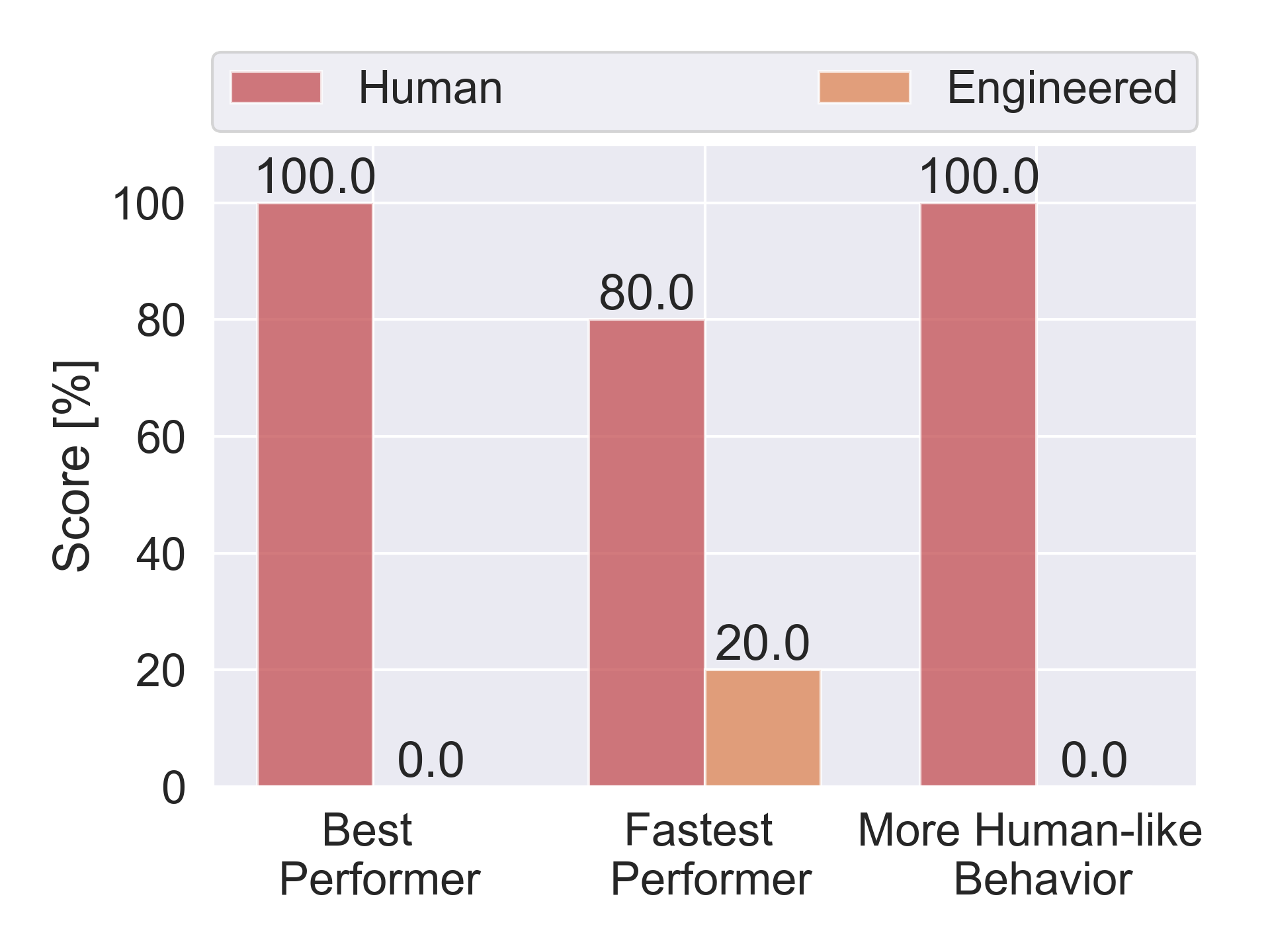}} &
        \subfloat[Human vs Hybrid]{\includegraphics[width=0.3\linewidth]{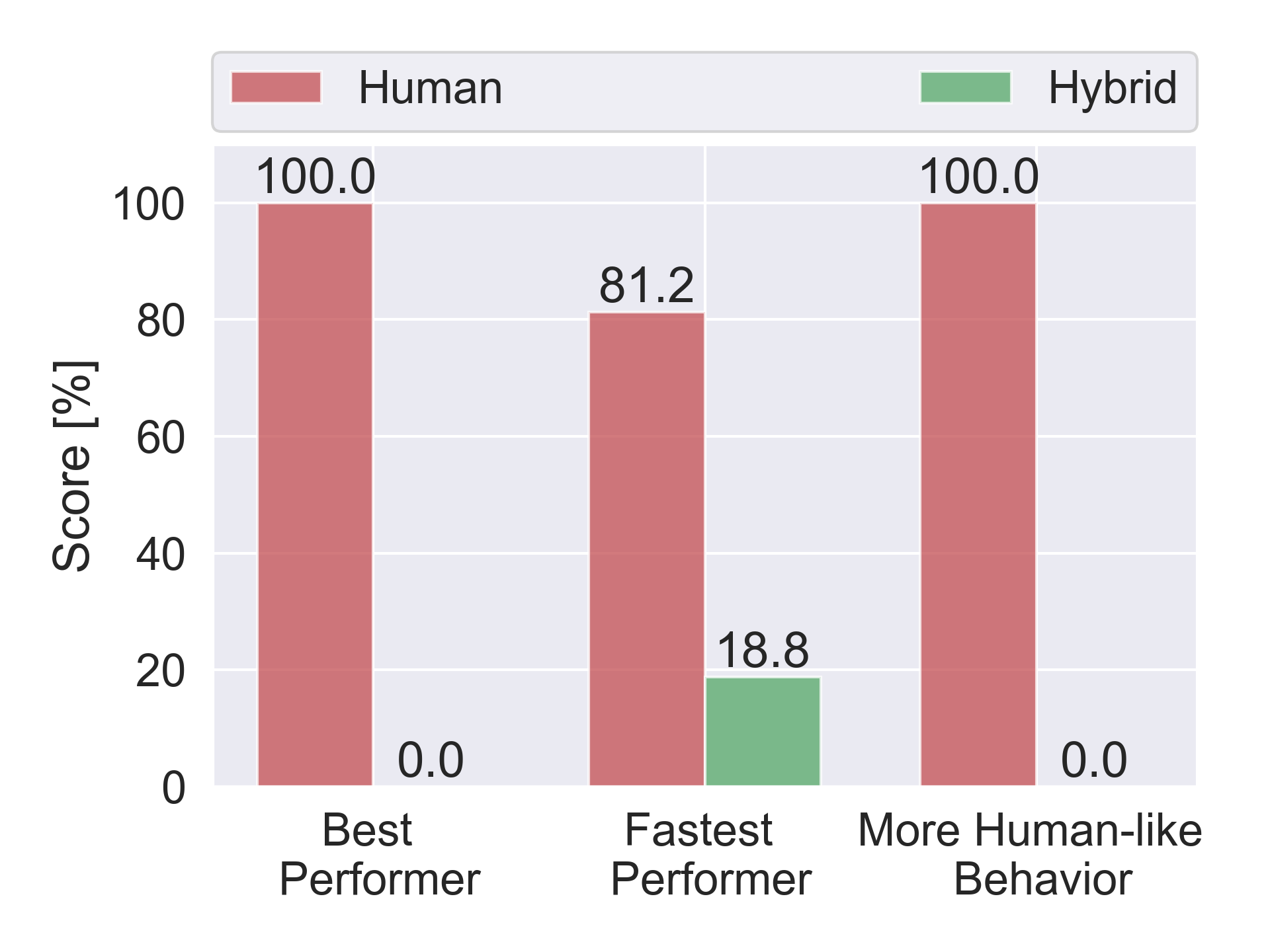}}
    \end{tabular}
  \caption{Pairwise comparison displaying the normalized scores computed from human evaluations separately for each performance metric on all possible head-to-head comparisons for all agent type performing the \textit{BuildVillageHouse} task.}
  \label{fig:house_barplot}
\end{figure}

Figures \ref{fig:cave_barplot}, \ref{fig:waterfall_barplot}, \ref{fig:pen_barplot}, and \ref{fig:house_barplot} show bar plots with the individual pairwise comparisons compiled from the human evaluations for the \textit{FindCave}, \textit{MakeWaterfall}, \textit{CreateVillageAnimalPen}, and \textit{BuildVillageHouse} tasks, respectively.
Each bar represents the percentage of the time a given condition was selected as a winner for each performance metric by the human evaluator when they were presented with a video of the agent performance solving the task for each analysed condition.
For example, when analyzing Figure \ref{fig:cave_barplot}(a), we can see that the human evaluator was presented with a video of the ``Behavior Cloning'' agent and another from the ``Engineered'' agent, they selected the ``Engineered'' agent as the best performer $33.3 \%$ and the ``Behavior Cloning'' agent $22.2 \%$ of the time. The remaining accounts for the ``None'' answer to the questionnaire selected when none of the agents were judged to have solved the task.

When directly comparing the ``Behavior Cloning'' baseline to the main proposed ``Hybrid'' method for all tasks, as shown in Figures \ref{fig:cave_barplot}, \ref{fig:waterfall_barplot}, \ref{fig:pen_barplot}, and \ref{fig:house_barplot} (c) plots, we observe that the proposed hybrid intelligence agent always match or outperforms the pure learned baseline.
This is similar to the case we compare the ``Engineered'' agent to the ``Hybrid'' agent, where the proposed hybrid method outperforms the fully engineered approach in all tasks except the \textit{BuildVillageHouse} task, as seen in Figure \ref{fig:house_barplot}.
The human players always outperform the hybrid agent with exception to the \textit{MakeWaterfall} task, where the ``Hybrid'' agent is judged to better solve the task $70 \%$ of the time, to solve it faster $90 \%$ of the time, and even present a more human-like behavior $60 \%$ of the time.
The ``Hybrid'' agent performing better  can be attributed to the fact that the human players were not always able or willing to solve the task as described in the prompt.

\section{Samples of Hybrid Agent Solving the Tasks}\label{appendix:frames}

\begin{figure}[!ht]
    \centering
    \begin{tabular}{cc}
        \subfloat[]{\includegraphics[width=0.4\linewidth]{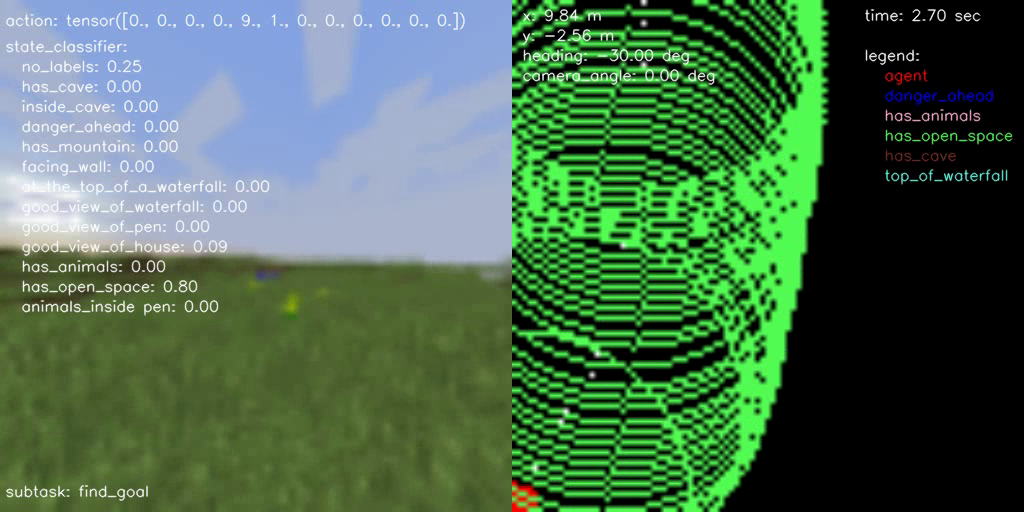}} &
        \subfloat[]{\includegraphics[width=0.4\linewidth]{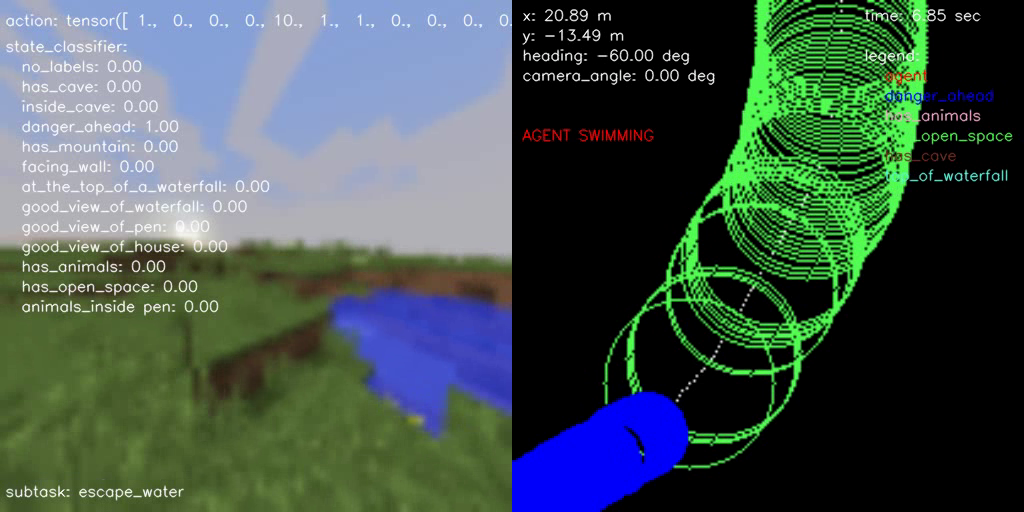}}\\
        \subfloat[]{\includegraphics[width=0.4\linewidth]{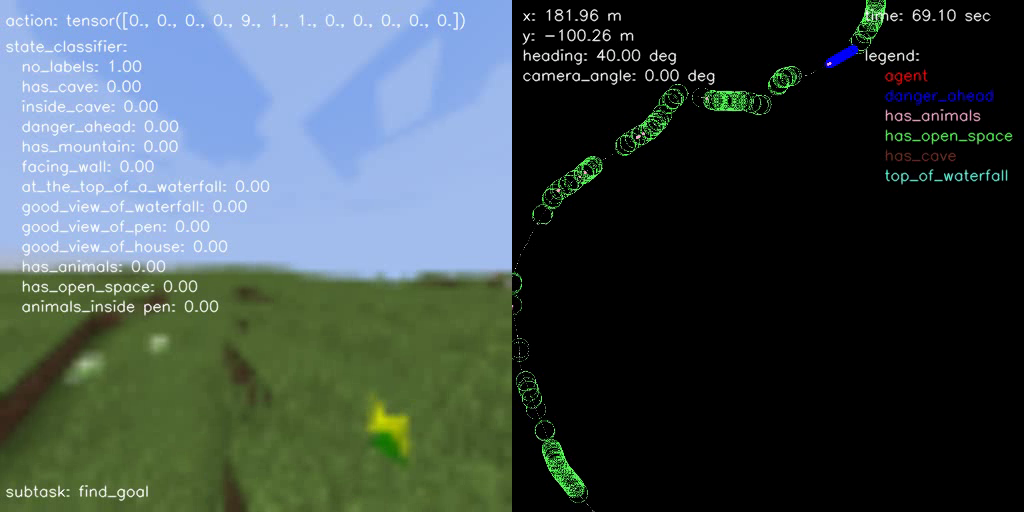}} &
        \subfloat[]{\includegraphics[width=0.4\linewidth]{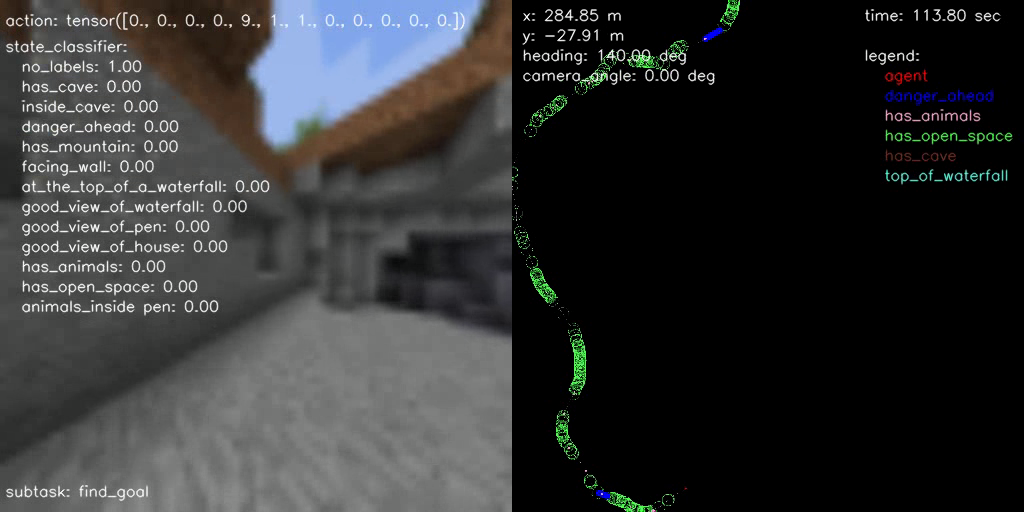}}
    \end{tabular}
    \caption{Sequence of frames of our hybrid agent solving the \textit{FindCave} task (complete video available at \url{https://youtu.be/MR8q3Xre_XY}).}
    \label{fig:bestcave_frames}
\end{figure}

\begin{figure}[!ht]
  \centering
    \begin{tabular}{cc}
        \subfloat[]{\includegraphics[width=0.4\linewidth]{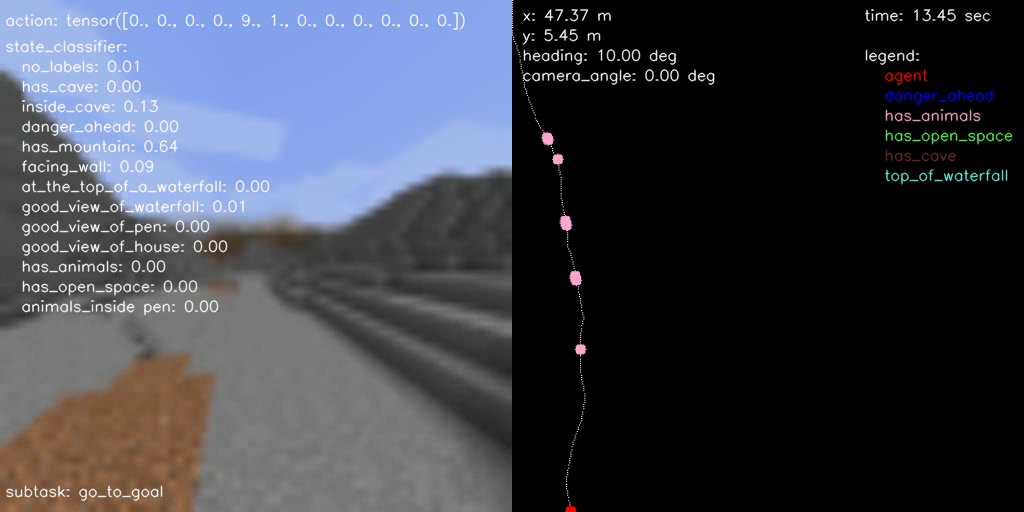}} &
        \subfloat[]{\includegraphics[width=0.4\linewidth]{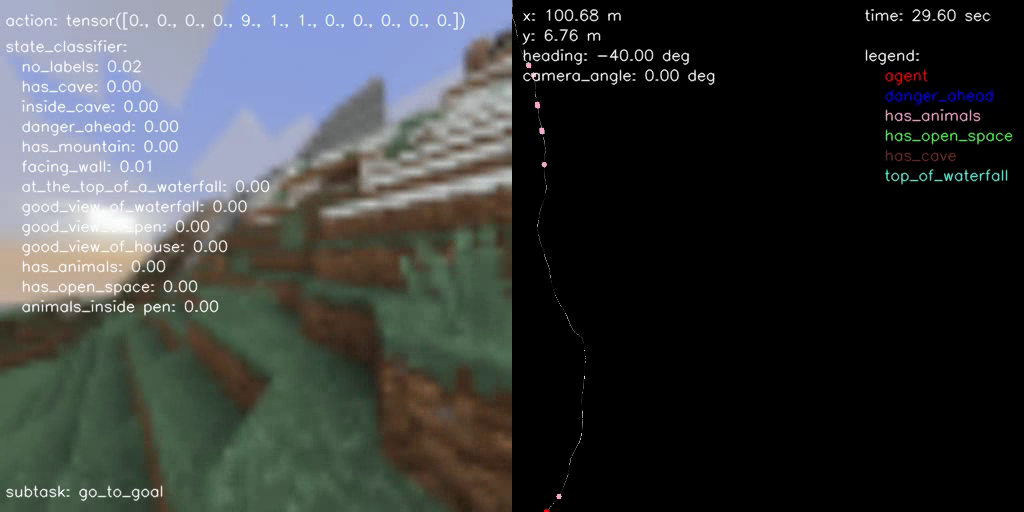}}\\
        \subfloat[]{\includegraphics[width=0.4\linewidth]{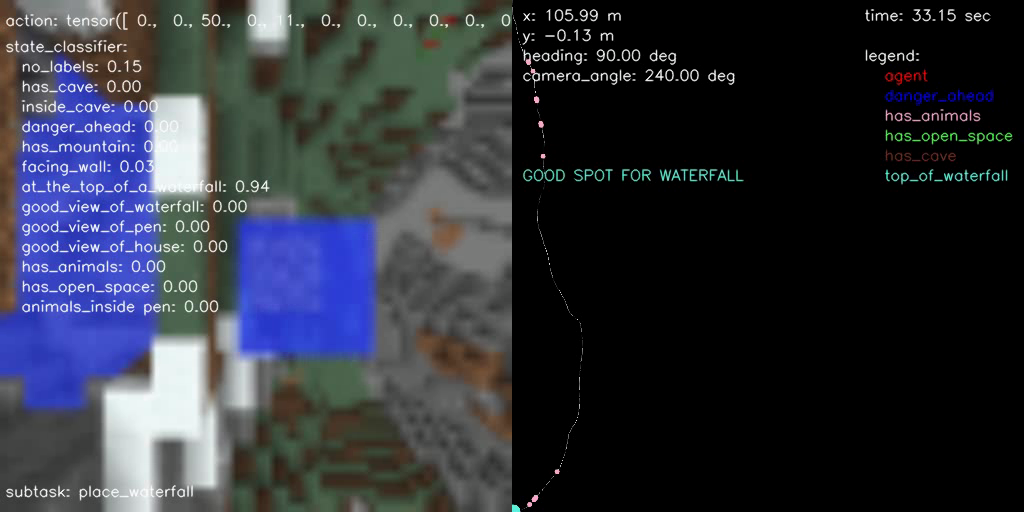}} &
        \subfloat[]{\includegraphics[width=0.4\linewidth]{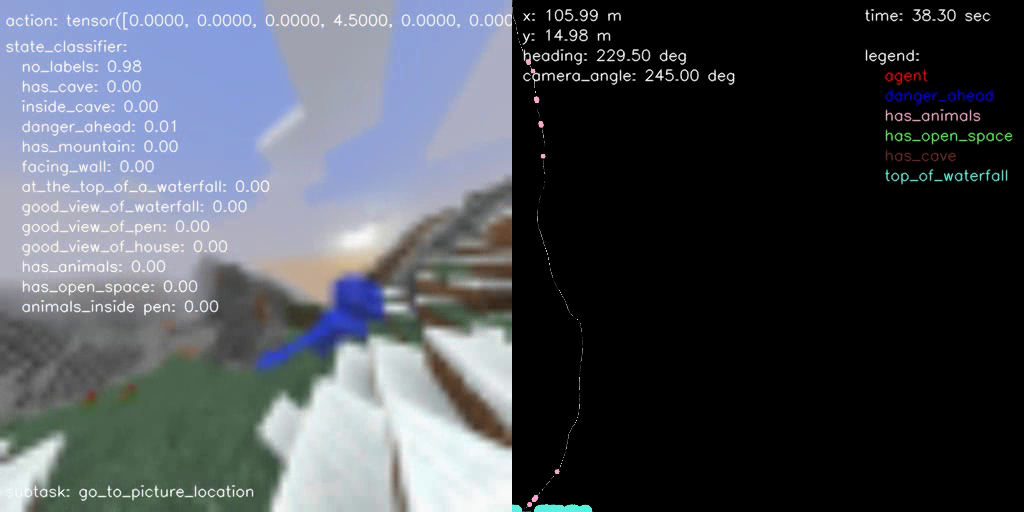}}
    \end{tabular}
  \caption{Sequence of frames of our hybrid agent solving the \textit{MakeWaterfall} task (complete video available at \url{https://youtu.be/eXp1urKXIPQ}).}
  \label{fig:bestwaterfall_frames}
\end{figure}

\begin{figure}[!ht]
  \centering
    \begin{tabular}{cc}
        \subfloat[]{\includegraphics[width=0.4\linewidth]{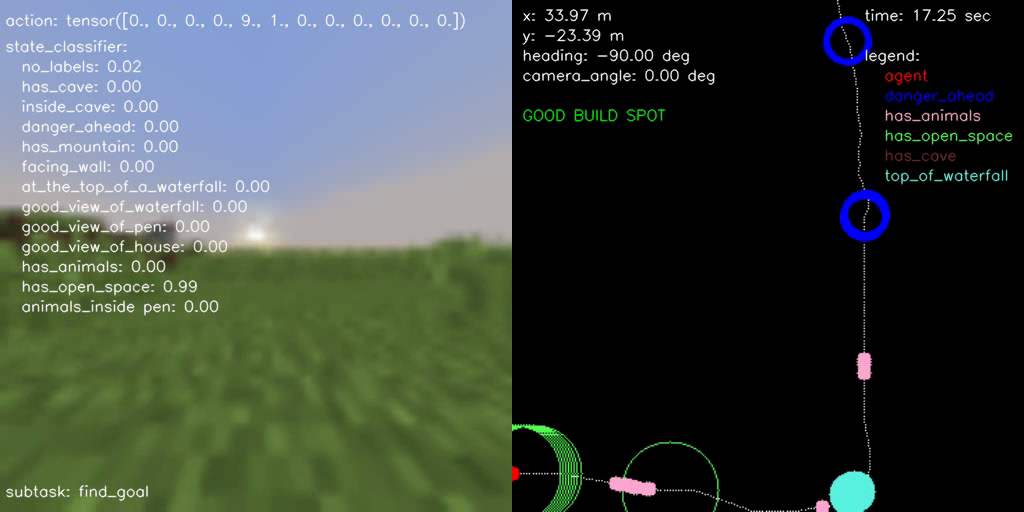}} &
        \subfloat[]{\includegraphics[width=0.4\linewidth]{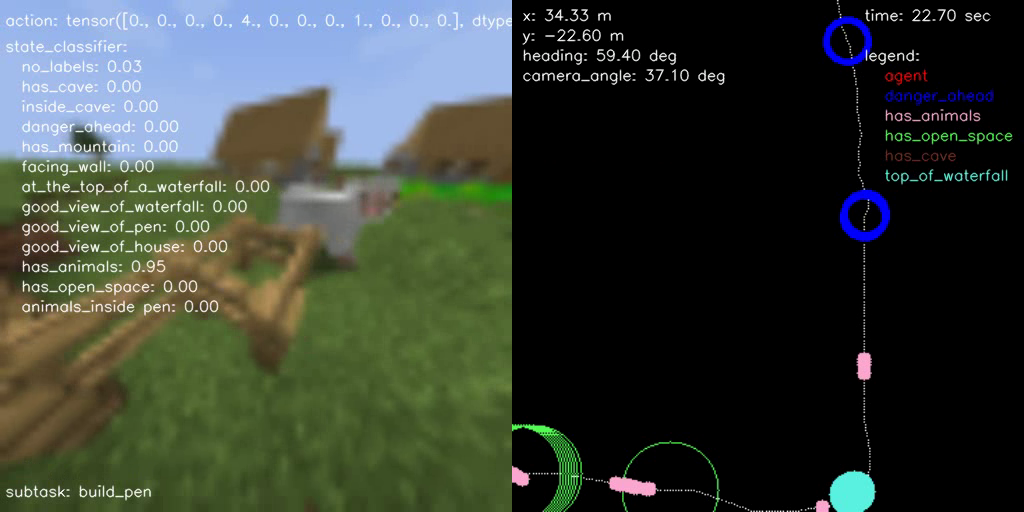}}\\
        \subfloat[]{\includegraphics[width=0.4\linewidth]{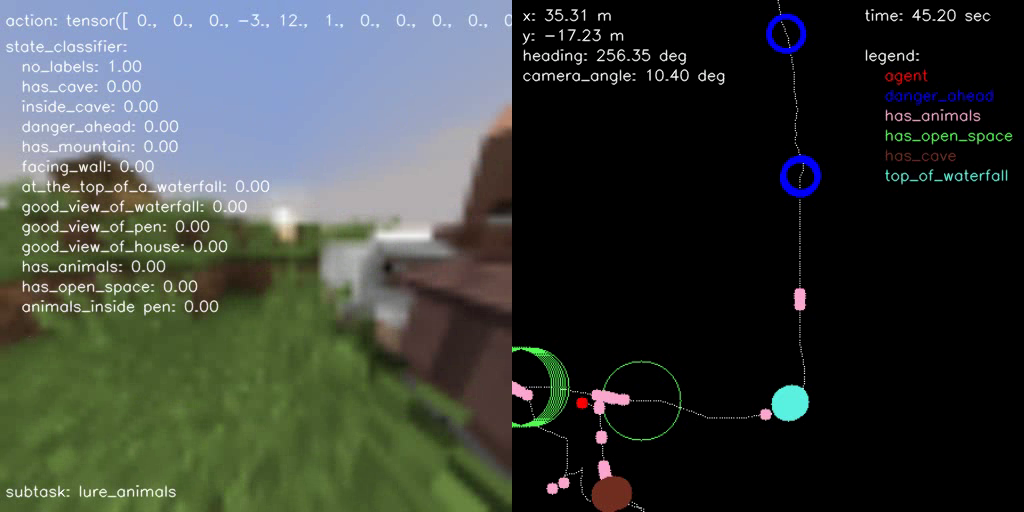}} &
        \subfloat[]{\includegraphics[width=0.4\linewidth]{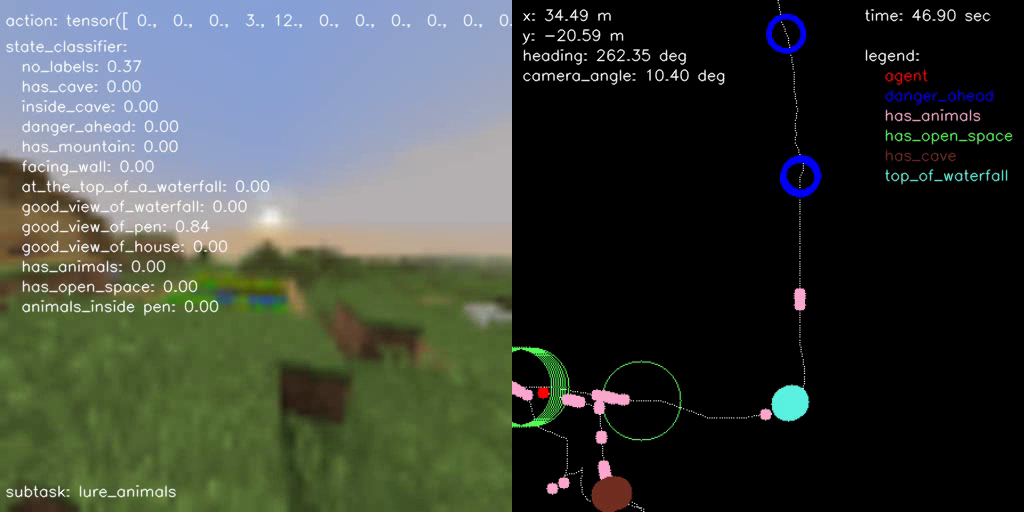}}
    \end{tabular}
  \caption{Sequence of frames of our hybrid agent solving the \textit{CreateVillageAnimalPen} task (complete video available at \url{https://youtu.be/b8xDMxEZmAE}).}
  \label{fig:bestcreatepen_frames}
\end{figure}

\begin{figure}[!ht]
  \centering
    \begin{tabular}{cc}
        \subfloat[]{\includegraphics[width=0.4\linewidth]{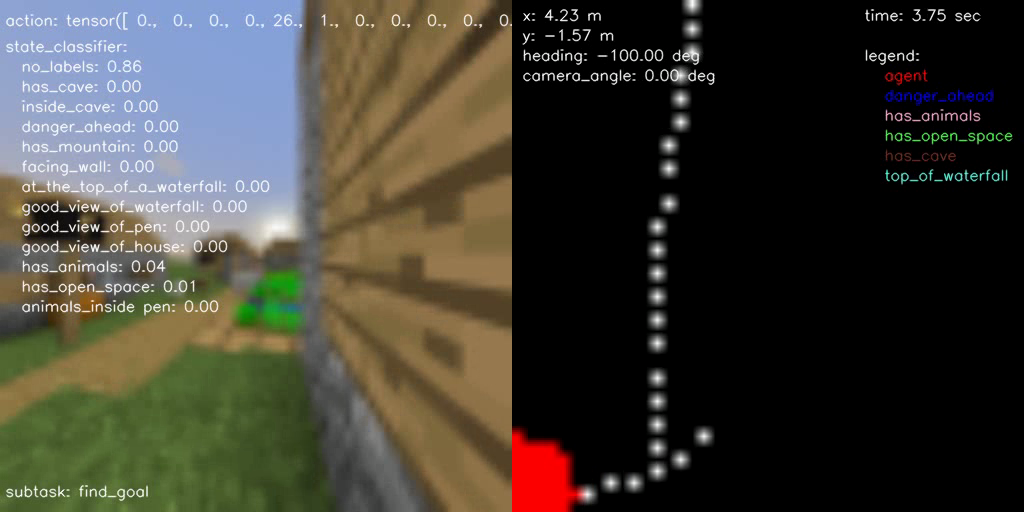}} &
        \subfloat[]{\includegraphics[width=0.4\linewidth]{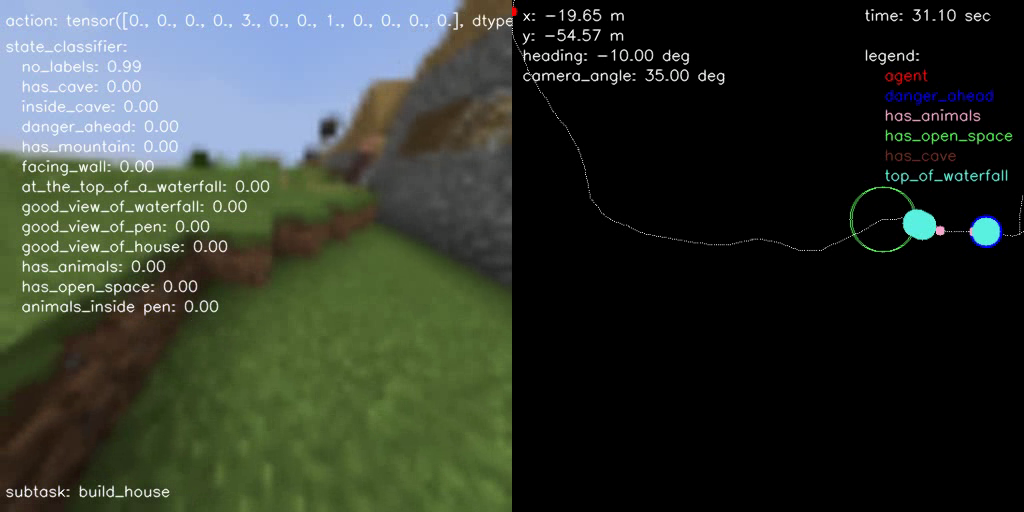}}\\
        \subfloat[]{\includegraphics[width=0.4\linewidth]{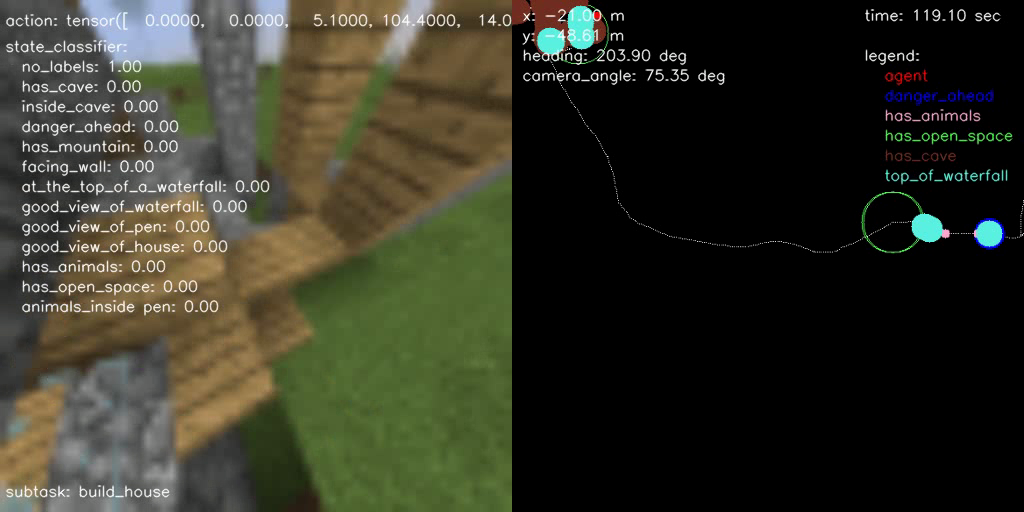}} &
        \subfloat[]{\includegraphics[width=0.4\linewidth]{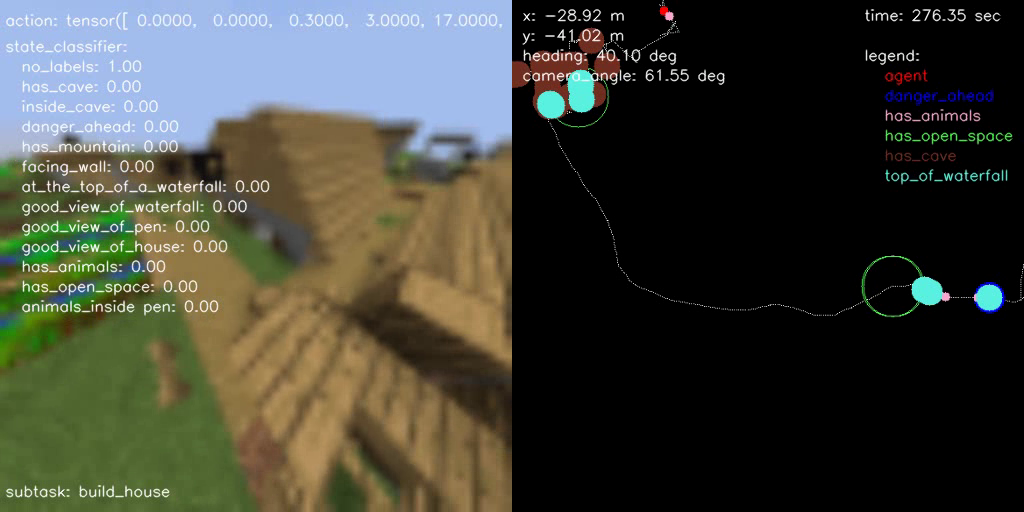}}
    \end{tabular}
  \caption{Sequence of frames of our hybrid agent solving the \textit{BuildVillageHouse} task (complete video available at \url{https://youtu.be/_uKO-ZqBMWQ}).}
  \label{fig:bestbuildhouse_frames}
\end{figure}

In terms of qualitative results, Figures \ref{fig:bestcave_frames}, \ref{fig:bestwaterfall_frames}, \ref{fig:bestcreatepen_frames}, and \ref{fig:bestbuildhouse_frames} show a sample episode illustrated by a sequence of frames of our hybrid agent solving the \textit{FindCave}, \textit{MakeWaterfall}, \textit{CreateVillageAnimalPen}, and \textit{BuildVillageHouse} tasks, respectively. Each figure shows the image frames received by the agent (left panel) overlaid with the actions taken (top), output of the state classifier (center), and the subtask currently being followed (bottom). The right panel shows the estimated odometry map overlaid with the location of the relevant states identified by the state classifier.
Link to the videos are provided in the figure captions.

\end{document}